Moscow Power Engineering Institute

(Technical University)

Training of spiking neural networks based on information theoretic costs

by

Sinyavskiy Oleg Y.

A thesis for the candidate degree of technical sciences

Specialty: 05.13.17 "Theoretical Foundations of Informatics"

Supervisor:

Prof. Kobrin A.I.

Moscow, 2011

# Abstract

Spiking neural network is a type of artificial neural network in which neurons communicate between each other with spikes. Spikes are identical Boolean events characterized by the time of their arrival. A spiking neuron has internal dynamics and responds to the history of inputs as opposed to the current inputs only. Because of such properties a spiking neural network has rich intrinsic capabilities to process spatiotemporal data. However, because the spikes are discontinuous "yes or no" events, it is not trivial to apply traditional training procedures such as gradient descend to the spiking neurons. In this thesis we propose to use stochastic spiking neuron models in which probability of a spiking output is a continuous function of parameters. We formulate several learning tasks as minimization of certain information-theoretic cost functions that use spiking output probability distributions.

We develop a generalized description of the stochastic spiking neuron and a new spiking neuron model that allows to flexibly process rich spatiotemporal data. We formulate and derive learning rules for the following tasks:

- a supervised learning task of detecting a spatiotemporal pattern as a minimization of the negative log-likelihood (the surprisal) of the neuron's output
- an unsupervised learning task of increasing the stability of neurons output as a minimization of the entropy
- a reinforcement learning task of controlling an agent as a modulated optimization of filtered surprisal of the neuron's output

We test the derived learning rules in several experiments such as spatiotemporal pattern detection, spatiotemporal data storing and recall with autoassociative memory, combination of supervised and unsupervised learning to speed up the learning process, adaptive control of simple virtual agents in changing environments.

# Acknowledgements

The thesis was done at the subdepartment of theoretical mechanics and mechatronics in Moscow Power Engineering Institue. I would like to express the deepest gratitude to my supervisor professor Alexander I. Kobrin for the invaluable help and advices during the preparation of the thesis. I would like to thank all my colleagues from the subdepartment for the warm support and help with the thesis.

Especially, I would like to deeply thank professor Witali L. Dunin-Barkowski for numerous valuable advices and discussions.

# Table of Contents

# Introduction

A real biological neuron is a complex biochemical system [1] that processes a continuous stream of multidimensional signals. The signals are called "spikes" - sharp voltage pulses propagating through neuron's body and appendages. Input spikes arrive at neuron's input extensions – dendrites. A neuron generates output spike trains that are propagated to other neurons via its single output extension – an axon. A connection site between the output of one neuron and the input of another (between the axon and dendrite) is called a "synapse". The duration of spike is usually about 1-2ms. It is common to ignore the variability in spike amplitude and duration so that a spike train can be considered as a stream of identical events. The single feature of a spike is the time of its occurrence in the communication channel. Typical profiles of the membrane potential of the real neuron are shown in Fig. 1 where one can see spike events as sharp voltage impulses.

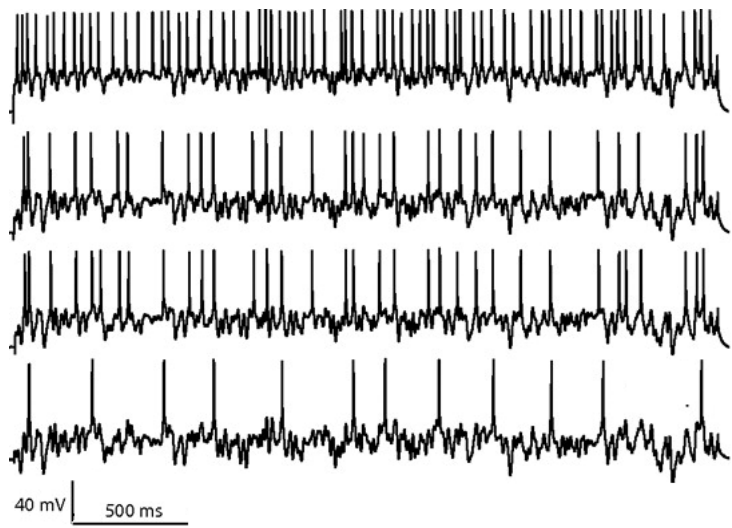

**Fig**. 1. Membrane potential traces of real neurons. Uniform events of membrane potential spikes can be clearly distinguished. The figure is taken from [52]. Copyright © 2008 National Academy of Sciences.

If one wants to study various properties of neurons using modelling, the models should take into account many details of neuron's inner workings such as the dynamic excitation properties of the neuron's membrane, the spatial arrangement of the neuron's components etc. However, in machine learning and computer science it is an open question which features of biological neurons are necessary and which can be neglected in order to build fast, powerful and efficient artificial neural networks. For example, there are discussions about the so-called "neural code" [2]: what characteristics of spikes patterns are needed for information processing and what characteristics are only artifacts of a particular biological implementation of such processing. Different points of view on this topic led to the creation of various types of neural network models.

Historically, the first simplified models of neurons are McCulloch – Pitts binary neurons [3]. Modelling of the binary neural network is performed in discrete time. At each time step a binary neuron produces a signal "1" (an output spike) or "0" (no spike). Input binary signals are summed up by the neuron with certain weights. If the weighted sum is larger than the threshold, the neuron generates an output spike. One of the first implementations of the learning algorithm for a binary neural network was proposed by F. Rosenblatt [4] ("perceptron").

Further development of neural learning algorithms led to the creation of rate-based neuron models. There is an experimental evidence showing that neurons use only a spike firing rate to perform certain computations (for example, during the primary associative processing of sensory signals [5,6,2]). Popular firing rate neuron models produce a firing rate output which is a weighted sum of input firing rates compressed by a nonlinear activation function (e.g. by the sigmoid function). Input and output signals in such networks are represented with float numbers. Initially, in order to train the rate-based networks researches used a variant of Hebb's postulate [7]: "When an axon of cell A is near enough to excite a cell B and repeatedly and persistently takes part in firing it, some growth or metabolic change takes place in one or both cells such that A's efficiency, as one of the cells firing B, is increased". This idea is usually implemented by increasing the weight between two simultaneously active neurons. For example, based on this postulate J. Hopfield developed autoassociative neural memory networks [8]. However, a big step forward in learning theory for the float-output networks happened after the development of the rigorous mathematical methods based on the error backpropagation [9, 10]. Such methods use gradient of the network's output with respect to the weights of neurons in order to minimize certain cost function. This allowed engineers to efficiently train float-output neural networks and use them in various practical applications [11-13].

Modelling of binary and firing rate neurons is usually done in discrete time. The output of such models does not depend on the input history but depends only on the current input and model parameters (for example, synaptic weights). This implies that the neurons have to have all necessary information at every time step for the successful learning and subsequent task execution. Usually such models are used with spatial data without temporal structure. However, in some practical applications it is necessary to process data that has rich temporal structure such as pattern prediction tasks, recognition of moving objects and adaptive control. These requirements led to the development of the firing rate models that use various methods of transforming temporal features into spatial features. This allows to use well-developed cost function minimization methods such as backpropagation. Examples include networks with delays [14] and recurrent neural networks that use the backpropagation-in-time algorithm [15]. There are also modified firing rate models which state is described by a system of differential equations that allows them to explicitly process a temporal component of the input data [16].

In parallel with the development of the firing rate models, neuroscience researchers started to gather evidence that at least some real neural systems use precise spike timing for signal encoding [17, 18]. It was shown in [19,20] that the speed of visual object recognition in multilayer neural networks of animals can not be achieved with firing rate-based signalling. Real neurons simply do not have time to accumulate a necessary number of spikes from the previous layers to get a robust firing rate estimate. Additionally, it was shown in [21] that even a single spike can influence the behavior of the whole network. In [22] it was shown that some brain structures perform firing rate to spike timing encoding. An important evidence that precise spike timing is important came from the discovery of Spike Timing Dependent Plasticity [23] (STDP) in late 90-s. According to STDP the changes in synaptic weights depend on the precise timing of input and output spikes. STDP is a generalization of the Hebb's rule: if an event A occurred before an event B the link between them should be increased. If the event A happened before the event B the link between A and B should be weakened. This creates an associative connection that respects the order and causality between the events and that can be weakened if the causality does not hold. These observations make us believe that the spike sequence processing is one of the basic computations performed by real biological neurons.

It is common to ignore the variability in duration and spike amplitude [1] and consider a spike train as a sequence of identical point events that are characterized only by the time of their appearance in the communication channel. A spiking neuron is a neuron model that processes spikes as a continuous multidimensional stream of point events. One of the simplest spiking neuron models is integrate-and-fire model [25]. The state of the neuron model consists of the membrane potential $u(t)$. If the membrane potential value crosses a certain threshold $u(t_k) = Th$, the neuron generates a spike at $t_k$. After that the membrane potential is reset to a certain value called "refractory potential": $u \leftarrow u_{refr}$. The membrane potential evolution is described by the following differential equation:

$$C\frac{du}{dt} = -u + \sum_i \sum_j w_i \delta(t - t_j^i) + \sum_k (u_{refr} - u)\delta(t - t_k^{out}), \qquad (1)$$

where $u$ is the membrane potential, $C$ is the time constant, $w_i$ is the input synapse weights, $t_j^i$ is the input spike time arrived at the $i$-th input, $t_k^{out}$ is the output spike time, $Th, u_{refr}$ is the threshold and the refractory potential. The first term at the right hand side attracts the potential to zero. The second term represents an impulse response of the potential to input spikes. Every input spike instantaneously changes the membrane potential by the value of $w_i$. After an output spike the potential instantaneously steps by $u_{refr} - u$ and becomes equal to $u_{refr}$. The value of $u_{refr}$ is chosen to avoid excessive spike generation (so called "refractoriness").

There are multiple modifications of the Integrate-and-fire model. One example is the quadratic integration model (QIF) [26] that is described by the following equation:

$$C\frac{du}{dt} = (u - u_{rest})(u - u_C) + \sum_i \sum_j w_i \delta(t - t_j^i) + \sum_k (u_{refr} - u)\delta(t - t_k^{out}),$$

where $u_{rest}$, $u_C$ are certain constants.

In general, every input spike generates an impulse response of the neuron, where the neuron is viewed as a dynamical system. Here we call such responses "postsynaptic potentials". Let us denote $q(t)$ a set of the neuron's state variables. The generation of an output spike in the general case happens when a certain condition on the state is satisfied $S(q(t)) = true$. The neuron's state changes after an output spike that can be viewed as another impulse response of the dynamical system. E. Izhikevich neuron model is an example of a more complex spiking neuron [27]. This model is capable of realistic simulation of real biological neurons behavior. The state evolution is described by two differential equations:

$$\frac{dv}{dt} = 0.04v^2 + 5v + 140 - u + I(\{t_j^i\}) + \sum_k (c - v)\delta(t - t_k^{out})$$

$$\frac{du}{dt} = a(bv - u) + d\sum_k \delta(t - t_k^{out})$$

where $u, v$ are the state variables, $I(\{t_j^i\})$ is the input current caused by the input spikes at $\{t_j^i\}$, $t_k^{out}$ - the output spike time, $a, b, c, d$ are the model constants. A spike generation condition is $v > -30mv$. The constants $a, b, c, d$ are usually chosen in order to simulate the dynamics of various types of real neurons.

A spiking neuron model integrates input signals with certain weights and generates an output like other neuron models. However, as opposed to the firing rate or binary neurons, spiking neurons use not only spatial but also a temporal component of the input data. Input spike at time $t_i$ can make a neuron to generate an output spike at $t_i + \Delta t$, where $\Delta t$ can be quite long. Therefore, the spiking neuron model itself takes into account temporal relations between inputs signals and does not require additional structures to transform temporal components into the spatial data. The explicit temporal processing makes us believe that spiking neural networks are natural candidates for solving practical spatiotemporal data processing problems.

Let us notice that precise spike timing coding hypothesis does not prohibit the usage of the firing rate codes along with other types of coding (e.g. oscillation phase-based coding [29, 30]). The research of spiking neurons augments the firing rate neurons research. Moreover, [31] shows that certain averaging operations on the spike timing code lead to the firing rate or binary neuron models, so there might be a continuum between spiking and non-spiking models.

## The advantages of spiking neurons.

1. **A single neuron can react to temporal aspects of input data**

The usage of time-distributed point events as a fundamental type of signalling allows spiking networks to process temporal data streams in a natural way in such applications as temporal prediction, recognition of fast moving objects and control [32, 33]. In "classical" neural networks (e.g. firing rate or binary neurons) the state of the unit depends only on the current values of inputs (only on the spatial components of the data) (Fig 2, left).

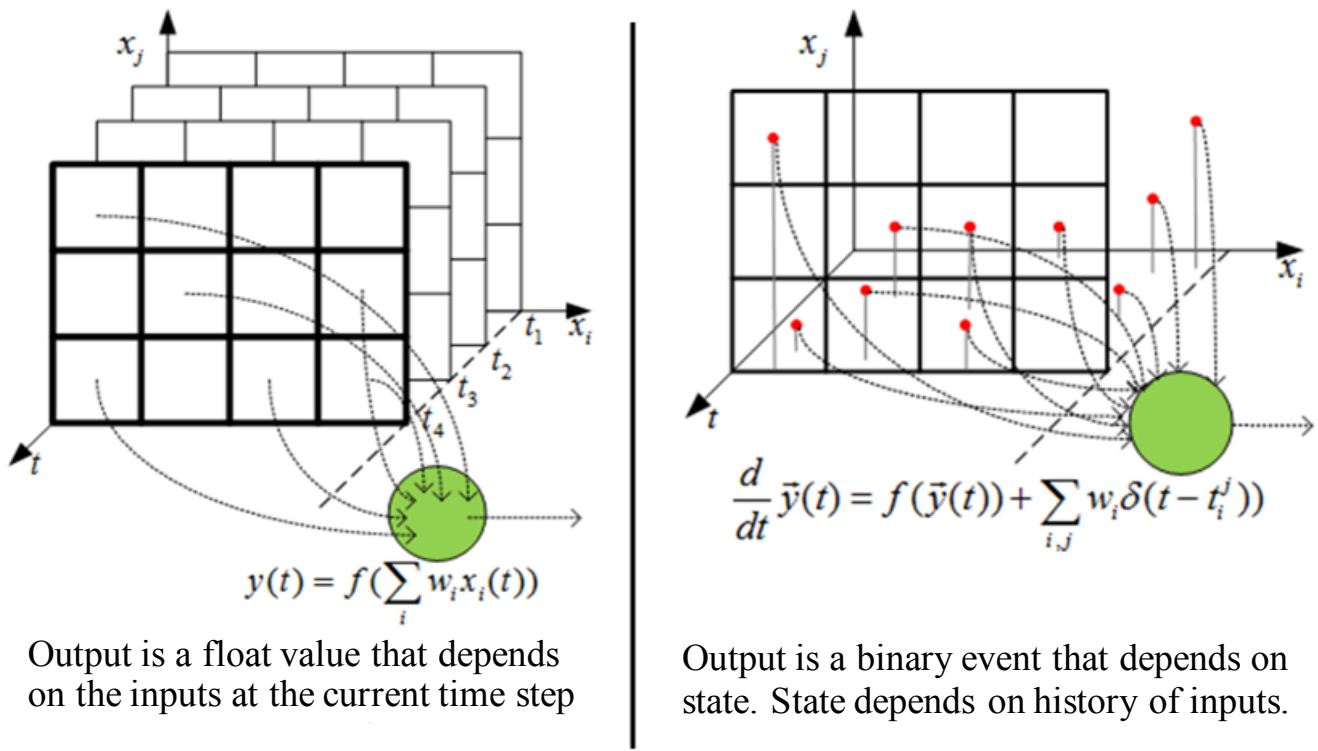

Output is a float value that depends on the inputs at the current time step | Output is a binary event that depends on state. State depends on history of inputs.

**Fig. 2.** Spatial and temporal processing of input data by a "traditional" firing rate neuron (left) and a spiking neuron (right). The firing rate neuron uses the input data from the current time step to produce a float value output. The spiking neuron uses the history of the input data to change its state variables. It generates binary events if its state satisfies certain conditions (e.g. the membrane potential crosses the threshold).

For processing data with temporal components additional structural mechanisms are used (recurrent connections, delays). In contrast, a spiking neuron responds to the input history because it's state is described by differential equations (Fig 1, right). This allows using the neuron's short term memory to deal with temporal aspects of data without introducing additional mechanisms. This suggests that certain tasks can be solved by simpler spiking networks with a smaller number of units and connections.

2. **The advantages of implementation of spiking neurons in highly parallel hardware**

New developments in computational hardware are often targeted to increase their parallel computation abilities. In particular, the development of neurocomputers (highly parallel biologically inspired computational systems) seems to be a particularly promising direction [34, 35]. However, with the increase of number of computational units, the number of connections between them grows exponentially. If neurons with float value outputs are used (e.g. firing rate units), a connection capacity has to be large enough to transmit float values with a necessary precision. Moreover, the "classic" neural networks are designed to transmit signals on every step of the computation. It is necessary to correctly synchronize the multilayered float output neural networks execution in order to correctly propagate the signals through the network.

For spiking neurons one needs to transmit only 1 bit of information between computational nodes and only during a spike occurrence. This does not put constraints on the capacity of the connections (Fig. 3). Spikes occur relatively infrequently which makes it unnecessary to transmit a signal on every time step. Also, in general spiking neurons do not have to be synchronized with each other.

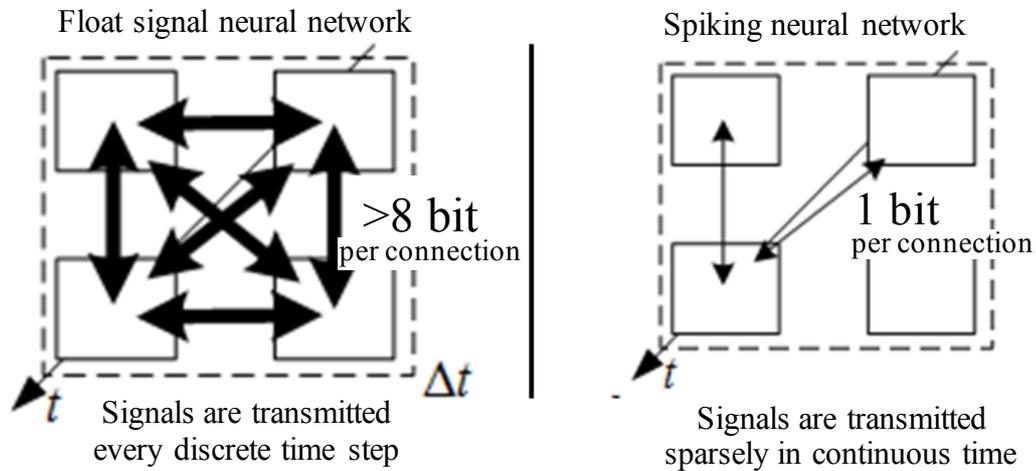

**Fig. 3**. A qualitative comparison of parallel hardware signal transmission with float output units (left) and spiking units (right). To transmit signals in the "traditional" firing rate network one needs to transmit several bits for every connection at every time step. To transmit signals in the spiking neural network one needs to transmit 1 bit per event only when needed. Additionally, the spiking network does not require a discretization of time.

Such features considerably lower the complexity of the hardware architecture, the size of the neurocomputer and its energy consumption comparing to the implementation of the same network using the float output units [36]. This should allow to use spiking neurocomputers in various small mobile devices.

### 3. Computational advantages of spiking neurons

Spiking neurons resemble McCulloch-Pitts binary neurons with their spike/no-spike output. However, spiking neurons also can process continuous values which can be encoded, for example, by the interspike interval duration. It was shown in [2, 24] that spiking neurons are capable of solving all the tasks that are solved by the float output and binary networks including continuous function approximation tasks.

In order to solve nonlinear classification task such as XOR, more than one-layer "classic" network is needed. However, it was shown in [37] that a single layer spiking neural network is capable of solving some tasks of non-linear classification. Such properties allow to decrease the number of spiking neurons and connections required to solve such tasks.

4. **Biological plausibility of spiking neuron models**

Spiking neuron models are closer to real biological neurons than firing rate or binary neurons. This allows researches to directly use biologically inspired methods to solve certain practical tasks. The results obtained during the spiking neuron modelling experiments can be directly compared to the data from the real neural networks. This allows to push forward our knowledge about the brain. Also it sometimes allows to find mistakes in the neural models by comparing its properties to the properties of the real network.

## The existing learning methods for spiking neurons

Currently the usage of spiking neurons in practical applications is quite limited. We speculate that this is caused by the lack of rigorous mathematical methods for spiking neurons training. This is in contrast to the gradient based cost function minimization methods which are extensively developed for the float output (firing rate) models. The differences between spiking and "classic" neuron models make it hard to apply existing well developed learning methods such as the error backpropagation. The main complication is the non-differentiability of the spiking model due to the threshold effects and the presences of internal state that evolves in time. The majority of learning rules for spiking models are based on the experimental neurophysiological observations (such as STDP) and do not have rigorous mathematical basis (however, sometimes they might still work well in practice [38-40]). In some applications spiking neurons are only used as a control or recognition system without using any learning [41, 42]. It also popular to use genetic algorithms [43-45] that proved to be effective with other neuron models [46, 47].

Some researchers expressed the opinion that the development of robust and effective learning rules for spiking neurons will lead to their broad adoption in practical applications [37, 24]. The main method for deriving the learning rules in float output networks is based on the optimization of certain cost function with respect to the parameters of the network. Unfortunately, the discontinuity that is present during a spike makes it hard to create a differentiable cost function that uses the network activity. Some learning methods do use the minimization of the difference between the actual and desired spike of Integrate-and-Fire neuron such as SpikeProp [48]: $\sum_{k}(t_k^{out} - t_k^d)^2 \to \min$. However, such cost function is discontinuous with respect to the neuron's weights. For example, if the weights of the non-active neuron are slowly increased, the neuron might start to generate output spikes because the membrane potential is going to reach the threshold. This generates a step discontinuity in the cost function value. Few heuristics are proposed in order to deal with this problem [48].

Adding the noise into the spike generation process allows one to study the probability of spike generation as a continuous function of neuron's weights [31]. This allows to construct well-behaved cost functions. There are several ways to add noise into the model (e.g. noise in weights, noise in spike timing). One way to add noise is to use a stochastic threshold mechanism. Deterministic spike generation is described as a certain Boolean condition on the neuron state (e.g. "the potential crosses the threshold"). With a stochastic threshold there is a non-zero probability of generating a spike even if such condition is not satisfied. The probability of a spike in this case is a continuous function of neuron's state and parameters. For example, the probability of spike might increase if the potential comes closer to the threshold value. Adding the noise to the model is also a valid modeling assumption since a real neuron does suffer from a large number of noise sources such as a thermal noise, probabilistic synaptic spike transmission etc.

It was proposed by some researchers [31, 49] to use the logarithm of the output spike generation probability as a cost function to minimize during the supervised learning. It was shown that the derived learning rules for weights resemble the STDP function. Furthermore, it was proposed in [50] to minimize the entropy of the neuron's output to obtain the optimal shape of the STDP curves. However, the full algorithm was too computationally demanding. It was proposed in [51] to use the amount of the transmitted information as a value to maximize during the unsupervised learning. In all mentioned cases the simple gradient descent methods were used. This shows that the information theoretic cost functions can be successfully used to formulate different learning tasks for spiking neuron models.

We think that the development of spiking neurons learning methods and unification of various learning tasks in a single mathematical framework is very important. In this work we will investigate into the approach of obtaining learning rules for spiking neurons using the entropic and information theoretic cost functions. In particular, we are going to make an attempt to formulate three kinds of learning tasks (supervised, unsupervised and reinforcement) from the generic point of view as an optimization of entropic functions. It is still an open question which model of the spiking neuron has all the necessary features to be practically useful in real world tasks. Because of that we aim to formulate learning processes for a generic spiking neural model without specifying too many details of its internal implementation. In the future research this should allow us to obtain learning rules not only for the simple models but also for complex biologically plausible models. In the experiments in this thesis we use a new spiking model of intermediate complexity to showcase obtained learning rules.

**The main purpose of the thesis** is to develop learning methods for spiking neural networks applicable in adaptive data processing, storing and recall of multidimensional spatiotemporal data and adaptive control tasks.

**The original results obtained in the thesis:**

1. We developed a new spiking neuron model - SMRM as an extension of the well-known Spike Response Model (SRM) with a stochastic threshold. This model has enriched response properties to the spatial and temporal aspects of the input spike patterns. It uses a special set of spike response kernels ("alpha functions") that allows it to adjust a response delay and amplitude on every input channel of the neuron.

2. We formulated a supervised learning task for the generic and SMRM neuron as a task of minimization of surprisal of the desired spiking pattern. A practical supervised learning task of detection of spatiotemporal patterns is solved using only a single SMRM neuron.

3. We developed the original architecture of the spatiotemporal autoassociative memory network using interacting spiking neurons. We used the developed supervised learning rules to train this network. The network is capable of storing several spiking patterns and is able to restore them based on the initial clue (the beginning of a pattern). System preserves not only the order but also the timing of the events in the patterns.

4. We formulated a particular task of unsupervised learning for the generic and SMRM spiking neuron and derived corresponding learning rules. In this task the neuron increases the robustness of generating the most likely output pattern by minimizing the its entropy. We conducted experiments with the SMRM neuron and showed that after training the robustness of a particular output pattern is increased and other less likely patterns are not generated anymore.

5. We showed that reinforcement learning of spiking neurons based on the direct gradient maximization of the received rewards can be formulated in the information-theoretic framework. We showed that a two-layer spiking neural network is capable of solving a particular control task even if in the lack of the necessary spatial data by using the temporal structure of the input and internal activity. We conducted the experiments in a dynamically changing environment and showed that the spiking neural network is able to control the agent and adapt to failures in its sensors and actuators. Additionally, we conducted the original experiment where the controlled by network virtual agent was replaced by another one. After that the network relearned to control the new agent.

# Chapter 1. Generalized spiking neuron model

**Summary:**

In this chapter we describe a multidimensional space of spiking patterns. We define a special metric on this space that allows us to estimate how close different spiking patterns are.

We introduce a generalized stochastic model of a spiking neuron that processes multidimensional spiking patterns and generates a one dimensional pattern. We derive the conditional probability distribution of a spike pattern using the point process intensity function.

Several learning tasks for the generalized spiking neuron can be formulated using the probability of an output spike pattern conditioned on the state of the neuron and the input spike pattern. It is proposed to use certain information theoretic characteristics as the cost functions to optimize during learning. In particular, we propose to use negative log-likelihood ("surprisal") for supervised learning and entropy on the space of output spike patterns for unsupervised learning. We describe how a reinforcement learning task can be formulated for the generalized neuron model.

Also we develop a particular implementation of the generalized spiking neuron model: Spike Multi Response Model – "SMRM". SMRM is a model of a neuron from the class of Spike Response Models (SRM0). We augment the model with a set of impulse response kernels per synapse (postsynaptic "alpha functions"). In a basic SRM0 model only a single alpha function per synapse is used. We proposed to use a weighted sum of alpha functions in a single synapse that allows the neuron to adapt the strength and delay of the postsynaptic potentials caused by input spikes. Several alpha functions allow the neuron to have a short term memory about its input patterns. This enables the neuron to react adaptively on the spatial and temporal structure of input patterns.

## 1.1. The space of spiking patterns

Real neurons communicate between each other using spikes – identical events characterized by the time of their appearance in a communication channel. Usually real neurons exhibit refractoriness property: it is hard for the neuron to generate a spike just after the previous spike. In this thesis we consider only sequences of single separated spikes assuming that neuron can not generate more than one spike per a small time interval ~1ms (we ignore spike bursts). An absolute refractory period $\Delta t_{refr}$ is a time interval after an output spike when the generation of the next spike is impossible. In other words, the probability of generating more that one spike during the interval which is smaller than refractory period is zero: $P\{n(t + \Delta t_{refr}) - n(t) > 1\} = 0$, where $n(t)$ is the number of spikes before the time $t$. In the theory of random processes such signal stream can be described as a realization of a stochastic point process.

Let us define a particular sequence of point events on an interval $T$ in the $i$-th input channel with small roman letters: $s_T^i$. The spike pattern $s_T^i$ can be fully specified with a sequence of absolute spike arrival times: $\{t_1, t_2, \ldots, t_n\}$, where $t_k$ is the absolute time of the arrival of the $k$-th spike. Another way to describe the pattern is to specify a reference time point $t_0$ and a sequence of interspike intervals (ISI) $\{\Delta t_1, \Delta t_2, \ldots, \Delta t_n\}$, where $\Delta t_1 = t_1 - t_0$, and $\Delta t_k$ is the time difference between the $k$-th and the $(k-1)$-th spike times.

We are going to use the upper index to define the index of the input channel (e.g., $\Delta t_k^i$). A particular multidimensional spike pattern $s_T^1, s_T^2, \ldots, s_T^n$ on the interval $T$ in $n$ communication channels is going to be denoted as $\bar{s}_T$. Notice that $\bar{s}_T$ can also be described as a set of interspike intervals $\Delta t_{k,p}^{i,j}$ - the interval between the $k$-th spike in the $i$-th channel and the $p$-th spike in the $j$-th channel. The number of free parameters needed to describe $\bar{s}_T$ is equal to the number of spikes in the pattern. Various ways to specify a spiking pattern are presented in Fig. 4.

A particular spiking sequence $s_T^i$ in the $i$-th channel belongs to the space $S_T$ of all possible spiking sequences in a single input channel on a particular time interval $T$. A particular $n$-dimensional spiking pattern $\bar{s}_T$ belongs to the space $\bar{S}_T^n$ - an $n$-dimensional Cartesian product of the spaces $S_T$.

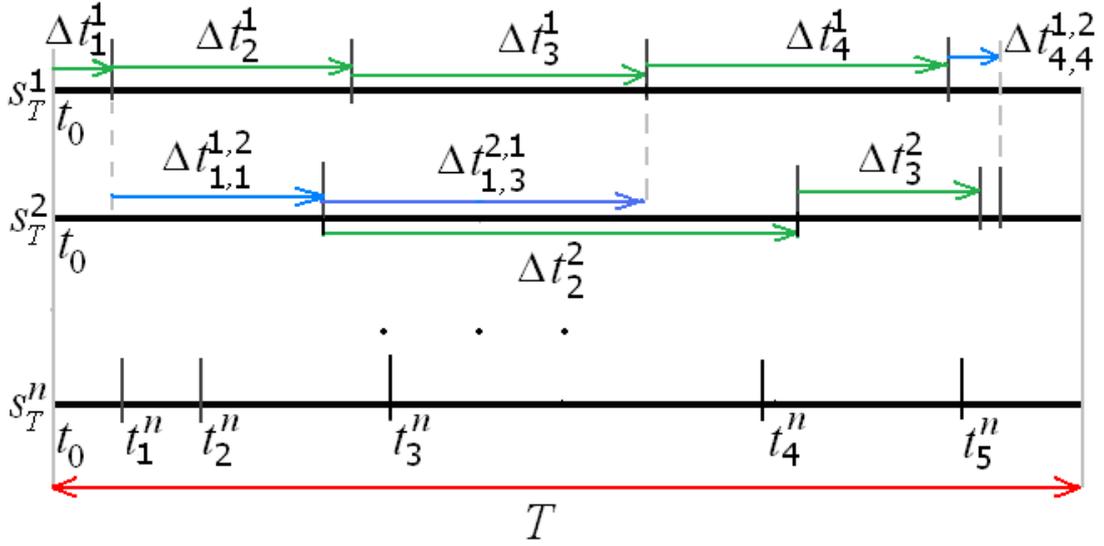

**Fig. 4.** Various ways to specify a spiking pattern. The spiking pattern is a set of identical point events in several communication channels on a given interval T. The spiking pattern can be specified using interspike intervals between the events in the channel (green arrows), using interspike intervals between events in different channels (blue arrows) or using the absolute times of the events (the bottom channel).

We can introduce a concatenation operation on the spiking pattern spaces. This operation transforms two spiking patterns $\bar{s}_{T_1}$ and $\bar{s}_{T_2}$ defined on the intervals $T_1$ and $T_2$ into the pattern $\bar{s}_T$ defined on the interval $T = T_1 + T_2$. The resulting spike pattern is going to consist of spikes first taken from pattern $\bar{s}_{T_1}$ and then from $\bar{s}_{T_2}$.

### 1.2. A distance function on the spiking patterns space

It is useful to have a method for a quantitative estimation of the similarities between the elements of the spike pattern space $\bar{S}_T^n$. We can construct a function $d:(\bar{S}_T^n \times \bar{S}_T^n) \to [0,\infty]$, that defines a distance that satisfies the standard metric axioms (non-negativity, identity, symmetry and the triangle inequality). Let us also define intuitive units of measurement of the spiking distance as the number of spikes in which two patterns $\bar{x}_T$ and $\bar{y}_T$ are different. In particular, if you take an empty pattern $\bar{0}_T$ without spikes on the interval T then the distance $d(\bar{x}_T, \bar{0}_T)$ should be equal to the number of spikes in the pattern $\bar{x}_T$. The time of a single spike in a pattern can vary continuously. We would like the distance measure to continuously vary as well. For example, let us assume that there are two identical spiking patterns $\bar{x}_T$ and $\bar{y}_T$ so that $d(\bar{x}_T, \bar{y}_T) = 0$. If we start to change the time of a particular spike $t_i$ in $\bar{x}_T$, the distance should slowly increase. If we move the spike far from other spikes, the distance value should tend to 2, which is a number of non-matching spikes in those patterns (Fig. 5).

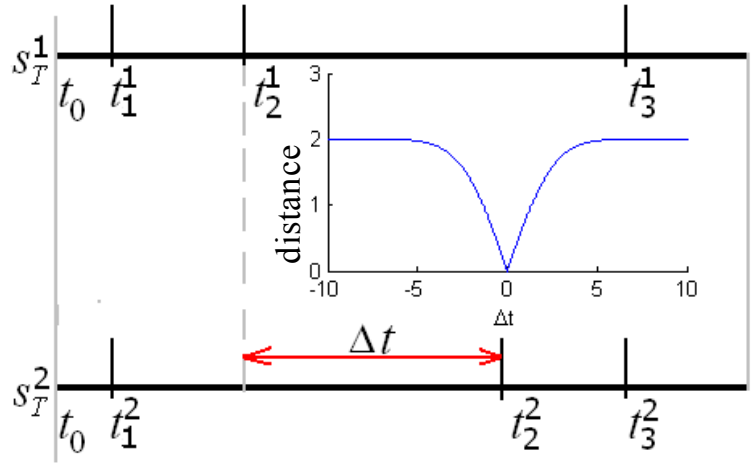

**Fig. 5.** Properties of the spike pattern distance function. Two patterns consisting of three spikes are shown on the top and the bottom of the figure. The first and the third spikes occurred simultaneously in both patterns. If we vary the interval between the second spikes we can build a plot of distance function value (in the middle). The distance is zero if the interval is zero. The distance is equal to two if the spikes are far appart.

In order to create such distance function let us consider a functional space $\Phi$, which is a subspace of continuous functions that have a limit of 0 when the argument approached infinity. Let us put the following constraints on $\Phi$: $\Phi = \{\varphi(t) : (\forall t, \varphi(t) = \varphi(-t)); \int \varphi(t)dt = 1\}$. In other words, functions $\varphi(x)$ are even and their integral is equal to one. Consider a single channel pattern $x_T$, defined as a set of spiking times $\{t_{x1}, t_{x2}, \ldots, t_{xl_x}\}$, where $l_x$ is the number of spikes in $x_T$. Let us choose some $\varphi \in \Phi$ and define a new function based on the spiking times as the following sum:

$f(t) = \sum_{i}^{l_x} \varphi(t - t_{xi})$. This function describes a density of spikes in $x_T$ at any given time. Using the same function $\varphi \in \Phi$ let us create $g(t)$ for $y_T$ in the same way: $g(t) = \sum_{i}^{l_y} \varphi(t - t_{yi})$, where $l_y$ is the number of spikes in $y_T$. We define the distance function $d^0(x_T, y_T)$ between the patterns $x_T$ and $y_T$ as:

$$d^0(x_T, y_T) = \int_{-\infty}^{\infty} |f(t) - g(t)| dt \qquad (2)$$

This function is non-negative for all pairs of elements in the space $\bar{S}_T^n$ and it satisfies all metric axioms:

1) $d^0(x_T, y_T) = \int_{-\infty}^{\infty} |f(t) - g(t)| dt = \int_{-\infty}^{\infty} |g(t) - f(t)| dt = d^0(y_T, x_T)$ - the symmetry axiom

2) $d^0(x_T, y_T) = 0 \Leftrightarrow x^T = y^T$ - the identity axiom

3) $d^0(x_T, y_T) + d^0(y_T, z_T) = \int_{-\infty}^{\infty} |f(t) - g(t)| + |g(t) - h(t)| dt \geq \int_{-\infty}^{\infty} |f(t) - h(t)| dt = d^0(x_T, z_T)$

   - the triangle inequality.

The constructed distance function also satisfies all additional requirements we have described. It continuously changes with varying spike times $t_i$. The value of the distance between $\bar{x}_T$ and an empty pattern is equal to the number of spikes in $\bar{x}_T$: $d^0(x_T, 0_T) = \int_{-\infty}^{\infty} |f(t)| dt = \sum_i \int_T \varphi(t - t_{xi}) dt = l_x$. These properties do not depend on the choice of $\varphi \in \Phi$. However, we can fine tune some of the characteristics of $d^0$ choosing various $\varphi$. For example, by choosing $\varphi(t) = \delta(t)$, where $\delta(t)$ is the Dirac's delta function, we obtain the distance $d_\delta$ that is equal to the number of non-matching spikes between the patterns. A convenient choice of $\varphi$ is the Gaussian function: $\varphi(t, \sigma) = \frac{1}{\sigma\sqrt{2\pi}} e^{-\frac{x^2}{2\sigma^2}}$ which sensitivity can be tuned with the parameter $\sigma$. An example of the distance between two sequences evaluated using $\varphi(t, 1)$ is shown on (Fig. 6).

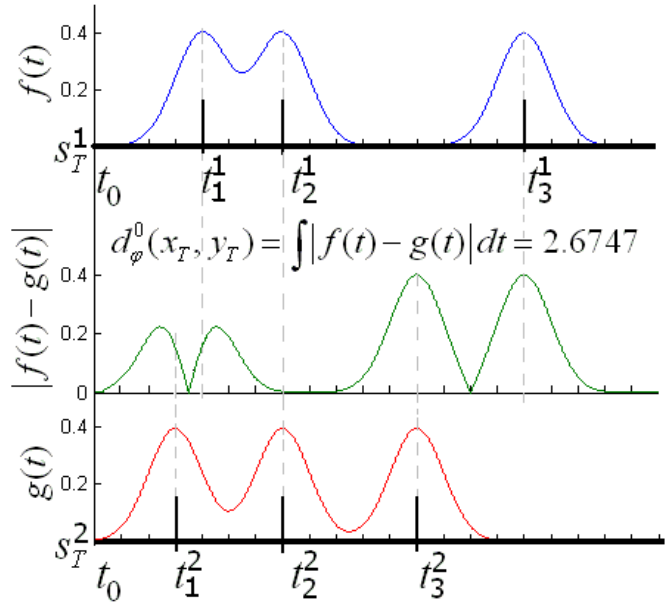

**Fig. 6.** The distance between two spiking patterns using the Gaussian function. Two patterns with three spikes each are shown on the top and the bottom of the figure. Each pattern is convolved with the Gaussian function (blue and red plots). The absolute difference between the convolved patterns is plotted in the middle in green. The distance value is the integral of the difference and is equal to 2.67 spikes for this example.

A distance between the multi-channel spiking patterns $\bar{x}_T, \bar{y}_T$ can be defined as a sum of the distances on each channel:

$$d(\bar{x}_T, \bar{y}_T) = \sum_{i}^{n} d^0(x_T^i, y_T^i) \qquad (3)$$

This distance function has all the same properties as the single channel distance function. Therefore, $\bar{S}_T^n$ is a metric space and for every spiking pattern $\bar{s}_T$ we can define a neighborhood of radius $r$ that corresponds to a set of "close enough" spiking patterns.

### 1.3. A generalized spiking neuron model

In general, a spiking neuron model has a state that deterministically depends on the input and output spikes and time $t$. Let us denote neuron's state space as $Q$ and the state at time $t$ as $q(t) \in Q$. Let us formalize the neuron's input-output behavior and its state evolution in discrete time. We introduce a function $\Lambda: Q \to [0,1]$ that defines a probability of the spike generation with the state $q(t_i) \in Q$ at time step $t_i$. Also we define a function that evolves neuron's state based on the previous state $q(t_{i-1})$, the presence of the output spike at the previous time step $y(t_{i-1}) \in B = \{0,1\}$ and the input spike s $\bar{x}(t_i) \in B^n$ at the current time step $t_i$: $F_q : (Q \times B \times B^n) \to Q$, where $B^n$ is the Boolean $n$-dimensional space. For the illustration, consider a neuron with a state $q(t_i)$. It is known if the neuron generated a spike $y(t_i)$ at $t_i$. The new state $q(t_{i+1})$ is computed using $F_q$ and the input spike train $x(t_{i+1})$ as follows: $q(t_{i+1}) = F_q(q(t_i), y(t_i), \bar{x}(t_{i+1}))$. Then we compute the probability of the output spike: $\Lambda_{i+1} = \Lambda(q(t_{i+1}))$. After that we generate an output spike $y(t_{i+1})$ at the new step $t_{i+1}$ with the probability $\Lambda_{i+1}$. Then we repeat the process: compute $q(t_{i+2}) = F_q(q(t_{i+1}), y(t_{i+1}), \bar{x}(t_{i+2}))$ and $\Lambda_{i+2} = \Lambda(q(t_{i+2}))$, generate a spike $y(t_{i+2})$ with probability $\Lambda_{i+2}$. We can include refractoriness when computing $\Lambda(q)$: if $t_{i+1} - t_i < \Delta t_{refr}$, then the probability of the spike generation is equal to 0, where $\Delta t_{refr}$ is the absolute refractory period.

Let us denote a probability of neuron being silent at $t_i$ as $L_i$. It is trivial that $L_i = 1 - \Lambda_i = 1 - \Lambda(q(t_i))$. Generation of a particular pattern $y_T$ on the interval $T$ is a stochastic event that is composed of all events at every time step $\Delta t$ that are consistent with pattern $y_T$: the neuron has to generate spikes at times that belong to the pattern $t_i \in y_T$ and to be silent at all other times $t_i \notin y_T$.

Let us compute the probability of a spike pattern $y_T$. In order to compute all $\Lambda_i$ and $L_i$ we need to compute all $q(t_i)$ on the interval $T$. Because of the dependency of the state on the output spikes we have to evolve the state as if the neuron actually have generated spikes according to $y_T$. After computing all $\Lambda_i = \Lambda(q(t_i))$ we just need to multiply the probabilities of the neuron outputs consistent with $y_T$ on all time steps:

$$P(y_T) = \prod_{t_i \in y_T} \Lambda(t_i) \cdot \prod_{t_i \notin y_T} (1 - \Lambda(t_i)) \qquad (4)$$

The set of all output spiking patterns $y_T$ (including the empty pattern $0_T$) conditioned on the state of the neuron and the input spike train is a probability space with elements $S_{T,\Delta t}$ where $\Delta t$ is a particular discretization of the interval $T$. If we split the interval $T$ on $n$ time steps $\Delta t$, then the total number of elements in $S_{T,\Delta t}$ is equal to $2^n$ and $\sum_{i=0}^{2^n - 1} P(y_{T,i}) = 1$.

Now let us consider the operation of the spiking neuron in continuous time. In continuous time $F_q$ is a shift operator on the solutions of a state evolution differential equation:

$$\frac{dq}{dt} = f_q(\bar{x}(t), y(t), q(t)) \qquad (5)$$

with initial conditions $q(0) = q_0$. Here $\bar{x}(t)$ and $y(t)$ are the functions that describe spiking patterns in continuous time and that can be defined as the sums of Dirac's delta functions:

$$\bar{x}(t) = \{\sum_j \delta(t - t_j^i), i \in 1...n\}; \; y(t) = \sum_j \delta(t - t_j^{out}).$$

The probability of a spike on a small interval $\Delta t$ can be computed as $\Lambda(t) = \lambda(t)\Delta t + \bar{o}(\Delta t)$, where $\lambda(t) = \lambda(q(t))$ is the point process intensity function. It can be proven that for the probability of an absence of a spike $L(T)$ on the interval $T$ the following holds [49]:

$$L(T) = e^{-\int_T \lambda(s) ds} \qquad (6)$$

Indeed, let us compute $L(T)$ as a product of probabilities of the spike absence at every time step $\Delta t$ in discrete time and then take the limit $\Delta t \to 0$:

$$L(T) = \prod_{t_i \in T} (1 - \Lambda(q(t_i))) = e^{\sum_{t_i \in T} \ln(1 - \lambda(q(t_i))\Delta t)} \underset{\Delta t \to 0}{=} e^{-\sum_{t_i \in T} \lambda(q(t_i))\Delta t + \bar{o}(\Delta t)} = e^{-\int_T \lambda(t) dt}$$

For example, we can compute the probability $L(t)$ that the neuron is not going to generate spikes on the interval $[0,t]$:

$$L(t) = e^{-\int_0^t \lambda(s)ds}$$

Notice that the following relation holds:

$$-\frac{dL(t)}{dt} / L(t) = \lambda(t) \qquad (7)$$

Let us introduce a probability density function $p_1(t_1|q_0)$ of the next spike being generated at time $t_1$ conditioned that the neuron has the initial state $q_0$ at the beginning of the interval $T$. The probability of generation of at least one spike on the interval $[0,t]$ is equal to

$$P(n(t) > 0) = 1 - L(t) = \int_0^t p_1(t_1|q_0)dt_1 \qquad (8)$$

where $n(t)$ is the number of spikes on the interval.

If we take a derivative of (8) with respect to $t$, we can find the probability density of generating the first spike at time $t_1$ using (7):

$$p_1(t_1|q_0) = -\frac{dL(t_1)}{dt} = \lambda(t_1)L(t_1) = \lambda(t_1)e^{-\int_0^{t_1} \lambda(s)ds} \qquad (9)$$

The probability density of generation of the second spike, provided the first spike has been generated at $t_1$, can be found using the same method:

$$p_2(t_2|t_1) = \lambda(t_2|t_1)e^{-\int_{t_1}^{t_2} \lambda(t_2|t_1)ds} \qquad (10)$$

The probability density of generating just a single spike at $t_1$ on the whole interval $T$ is equal to:

$$p(y_T = \{t_1\}|q_0) = p_1(t_1|q_0) \cdot e^{-\int_{t_1}^T \lambda(s|t_1)ds} = \lambda(t_1)e^{-\int_0^T \lambda(s|t_1)ds}$$

The combined two-dimensional probability density of generating only two spikes at times $t_1, t_2$ is equal to:

$$p(y_T = \{t_1, t_2\}|q_0) = p_1(t_1|q_0) \cdot p_2(t_2|t_1)e^{-\int_{t_2}^T \lambda(s|t_1,t_2)ds} =$$

$$= \lambda(t_1)\lambda(t_2|t_1)e^{-\int_0^T \lambda(s|t_1,t_2)ds}$$

The output spikes at times $t_1, t_2$ change the state of the neuron and therefore change $\lambda(t)$. The value of $\lambda(t)$ also depends on the input pattern $\bar{x}_T$. Therefore, given the input pattern $\bar{x}_T$, the output pattern $y_T$ and the initial state $q_0$ we can compute the evolution of $\lambda(t)$ on the interval $T$. Let us ignore the conditional notation for $\lambda(t)$ assuming that $\lambda(t)$ has been computed provided a particular input and output patterns and an initial state. By increasing the number of output spikes we can compute the probability density of generating any given output spiking pattern $y_T$:

$$p_T(y_T|\bar{x}_T, q_0) = \prod_{t_j^{out} \in y_T} \lambda(t_j^{out}) \cdot e^{-\int_T \lambda(s)ds} \quad (11)$$

If the pattern $y_T$ consists of $n$ spikes, the probability density of generating this pattern is $n$-dimensional. The whole set of all probability densities defines the probability space of all possible spiking patterns. An element of this space is a spiking pattern $y_T$. Various quantities can be computed by integrating those probability densities. It is important to remember that a spiking pattern is an ordered set of spiking times: $t_{i+1} > t_i$. For example, in order to compute the probability of generating of exactly two spikes on the interval $T$ one should use the following equation:

$$P(n(T) = 2) = \int_0^T \int_{t_1}^T \lambda(t_1|q_0)\lambda(t_2|t_1) e^{-\int_0^T \lambda(s|t_1,t_2)ds} dt_2 dt_1$$

Notice that in the continuous case there exist some patterns with two spikes being really close to each other (within some interval $\Delta t$). If $\Delta t < \Delta t_{refr}$ such patterns are impossible to generate because of the refractoriness so the probability density of such patterns has to be zero. Therefore a number of output spikes on interval $T$ is finite and limited by $n_{max} \leq T / \Delta t_{refr}$. This means that there is a limit on the maximum dimensionality of the elements of the space (it is not infinite dimensional).

Functions $F_q$ and $\Lambda$ for the discrete time case and operators $f_q$ and $\lambda$ for the continuous time case fully define the behavior and state evolution of the generalized spiking neuron with the initial state $q_0$ on the interval $T$. Such neuron performs stochastic transformation of an input spike pattern $\bar{x}_T$ into an output spike pattern $y_T$ (Fig. 7).

The neuron's state $q$ stores some amount of memory about the past events. The probability $P_T$ depends only on initial state $q_0$ and does not depend on the history of state evolution. Therefore, the behavior of the neuron on sequential time intervals is a Markov process (see equations (4) and (11)). The probability of generating the pattern $y_T$ on the whole interval T can be expressed via probabilities of generating $y_{\Delta T}$ on the parts $\Delta t$ of $T$ taking into account the state evolution based on the input and output spikes:

$$P_T\{\bar{y}_T | \bar{x}_T, q_0\} = \prod_{i}^{l} P_{\Delta T_i}\{y_{\Delta T_i} | \bar{x}_{\Delta T_i}, q_{i-1}\} \tag{12}$$

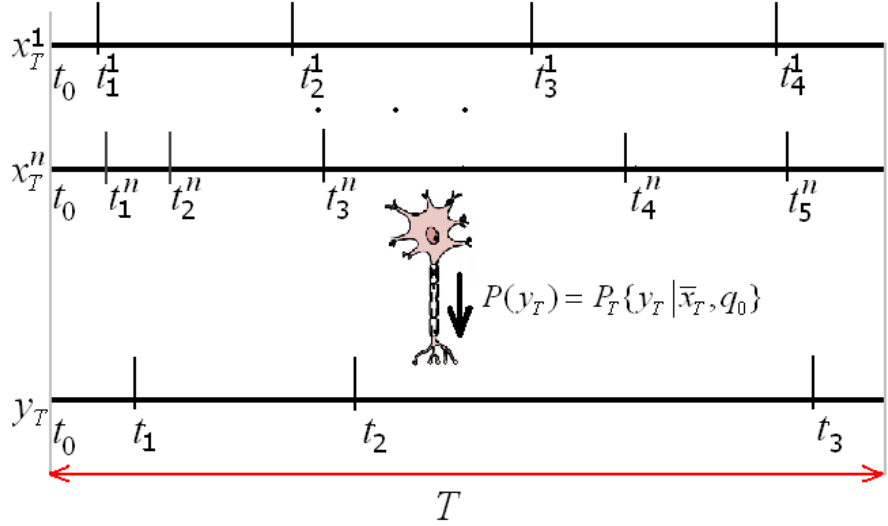

**Fig. 7.** Neuron as a stochastic processor of spiking patterns. A multidimensional input pattern *x* is shown on top. The neuron generates an output pattern *y* (on the bottom) on the interval *T* with a certain probability that depends on the input pattern and on the initial state of the neuron.

Consider a particular neuron model in continuous time that is defined using differential equations with the right hand side operator $f_q$ and the point process intensity function $\lambda$. Such model can be build by analyzing neurophysiological data or by simplifying/extending the already existing model. During the computer simulations it necessary to convert the model into the discrete time domain correctly. The task is to find functions $F_q$ and $\Lambda$ using their continuous counterparts $f_q$ and $\lambda$. The state evolution function $F_q$ can be obtained by the numeric integration of the differential equations $f_q$. Then we can find $\Lambda$ based on the integration step $\Delta t$ and the function $\lambda$ that depends on $q$. We assume that the value $\lambda(q(t_k)) = \lambda_k$ is constant during the $k$-th integration step. If the discretization step is small, then one can use the approximate equation:

$$\Lambda_k = \lambda_k \cdot \Delta t; \; L_k = 1 - \lambda_k \cdot \Delta t$$

If the discretization interval is large enough ($\lambda_k \cdot \Delta t > 1$), one can treat the spike generation process on this interval as the stationary Poisson point process with the intensity $\lambda_k$ [53]. The probability of the absence of a spike $L_k$ on the $k$-th step can be computed using:

$$L_k = e^{-\lambda_k \cdot \Delta t} \tag{13}$$

We can not distinguish one from multiple spikes on $\Delta t$ after discretization. Therefore we can find $\Lambda$ as the probability of generation at least one spike on the $k$-th step using:

$$\Lambda_k = 1 - e^{-\lambda_k \cdot \Delta t} \qquad (14)$$

### 1.4. Formulating learning tasks for the generalized spiking neuron

During the training neuron can make use of various sources of information that can help it to improve the performance in a particular task. Input and output signals are always available for the learning mechanisms inside the neuron. There are also special input signals that have the information about the task and that are used only during training. The learning task can be categorized depending on the availability and type of such extra teaching signals. If only input and output signals are available, the learning is unsupervised. If the extra teaching signal is present and it explicitly provides the desired output of the neuron, then it is a supervised learning task. Finally, in reinforcement learning task the neuron tries to optimize a certain external cost function. In this case a teaching signal provides information about the current cost function value.

Let us denote the whole set of neuron's parameters that can be changed during the learning as $\bar{W}$. We denote as $N(\bar{W})$ the generalized neuron with a particular value of parameters $\bar{W}$. The parameters $\bar{W}$ are included in the state of neuron and therefore they condition the probability distribution $P_T\{\bar{y}_T | \bar{x}_T, q(\bar{W})\}$ of generating $y_T$ (Fig. 8).

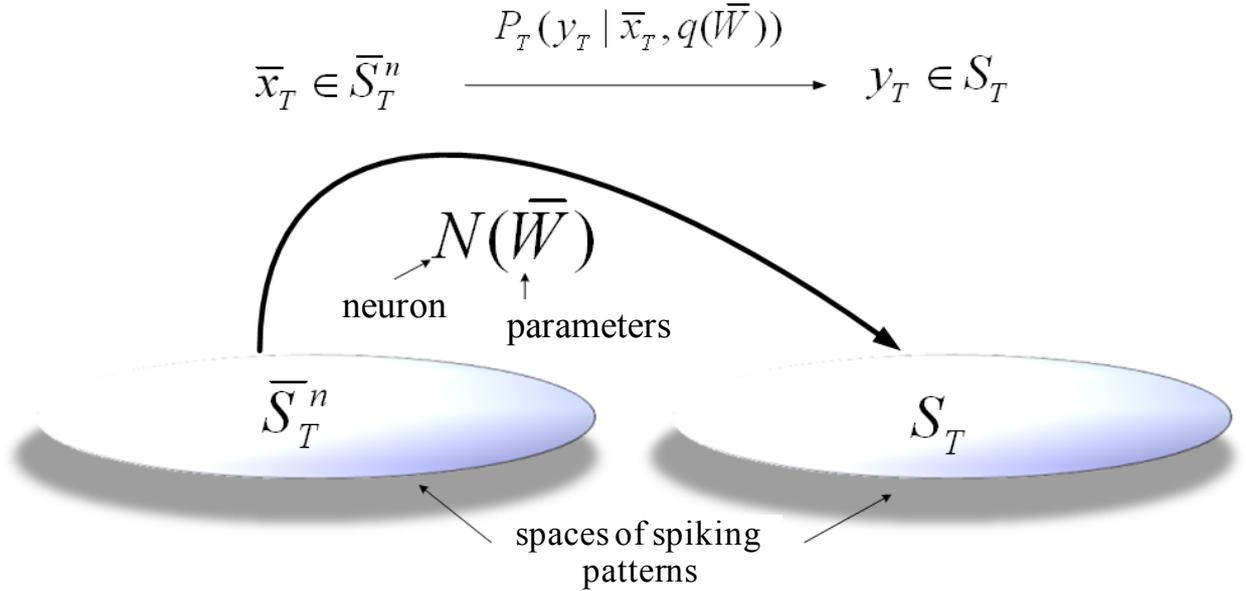

**Fig. 8.** A diagram of pattern transformation done by a neuron with parameters $\bar{W}$. The input pattern $x$ from the space of multidimensional input patterns undergoes a probabilistic transformation into the pattern $y$ from the one-dimensional space of patterns on the interval $T$. The transformation is parameterized by the neuron model $N$, its state $q$ and the parameters $\bar{W}$.

It is usually assumed that the processes of learning in biological neurons happens in neuron's synapses. By changing the synaptic efficacy of the weights the neuron can strengthen or weaken the influence of input spikes on it's state. The synaptic efficacy is usually characterized by a single number called synaptic weight $w_i$. The larger the synaptic weight $w_i$, the stronger is the influence of this particular synapse $w_i$. In this thesis we are going to consider more general case when every input channel (synapse) of the neuron can have multiple weights each characterizing a particular aspect of the input spike influence on the neuron's state. By $\bar{W}$ we are going to denote the whole set of neuron's weights.

There are different ways to split training and operation phases of the neuron. Usually in artificial neural networks training and operation (testing) are separated in time (offline learning). After the training is complete, the parameters are fixed and the neuron becomes a fixed data processor. Here we hold the hypothesis that in biological neural networks learning is not split on training/operation phases and neurons never stop adapting. In this work we would like to develop methods suitable for such learning (online learning). Also, similarly with biological networks we would like the develop methods that can be used in continuous time.

Now let us consider few examples of learning tasks for the spiking neuron. During the supervised learning a neuron has to generate an output pattern provided by the "teacher" (a teaching input) in response to a specific set of sensory input patterns. In an unsupervised learning task a neuron has to generate an output pattern that in some way better characterizes the properties of the input patterns. For example, the neuron can generate the output patterns which has the least amount of entropy given the input spikes. In the reinforcement learning a neuron has to "understand" relations between input, output and reinforcement signals in order to maximize the received reinforcement.

### 1.4.1. Supervised learning of the generalized spiking neuron

The supervised learning task can be formulated as follows. A neuron receives teaching spikes at times $t_i^d$ that indicate that the neuron itself should have generated a spike at $t_i^d$. The job of the neuron is to adapt its parameters in order to actually generate the desired spikes itself in the similar conditions – having the similar initial state and after receiving a similar input pattern. In other words, receiving a supervised spike means that the current input pattern is somehow special and the "teacher" wants the neuron to react on such pattern in the future. If the input $\bar{x}_T$ is fixed, the supervised spikes $t_i^d$ define the desired output pattern $y_T^d = \{t_1^d, t_2^d, \ldots, t_n^d\}$ that teacher wants the neuron to generate in response to $\bar{x}_T$. During the training iteration of length $T$ the neuron receives the input $\bar{x}_T$ together with the desired pattern $y_T^d$ and adapts its parameters. Notice that splitting the training process on iterations is not a necessary condition: the training can be set up in continuous time just concatenating the intervals $T$. However, some artifacts can possibly appear during the training due to the neuron carrying over its state to the next training interval.

Consider a generalized spiking neuron model characterized by the probability distribution $P_T\{y_T|\bar{x}_T, q_0\}$ on the interval $T$. The surprisal of a particular output pattern $y_T \in S_T$ is the quantity:

$$h(y_T) = -\ln(P\{y_T\}) \qquad (15)$$

Other names for this quantity are "negative log-likelihood" of $y_T$ or "self-information". The surprisal is a deterministic function of $y_T$ so it's value is a random variable which distribution depends on the input spike train and the neuron's initial state: $h(y_T) = h_T(y_T|\bar{x}_T, q_0) = -\ln(P_T\{\bar{y}_T|\bar{x}_T, q_0\})$. Elements that have the largest value of the surprisal will be generated less likely. The value $h(y_T)$ characterizes the degree of unpredictability of a particular neuron's output (hence the name "surprisal"). If the output $y_T$ has a very small value of the surprisal, it will be generated almost always in these conditions.

If the neuron's initial state $q_0$ is fixed, then for every pair of points $\{\bar{x}_T^*, y_T^*\}$ from the input and output pattern spaces we can compute the probability $P_T^* = P_T\{\bar{y}_T^*|\bar{x}_T^*, q_0\}$ and the surprisal $h_T^* = -\ln(P_T^*)$ (Fig. 9).

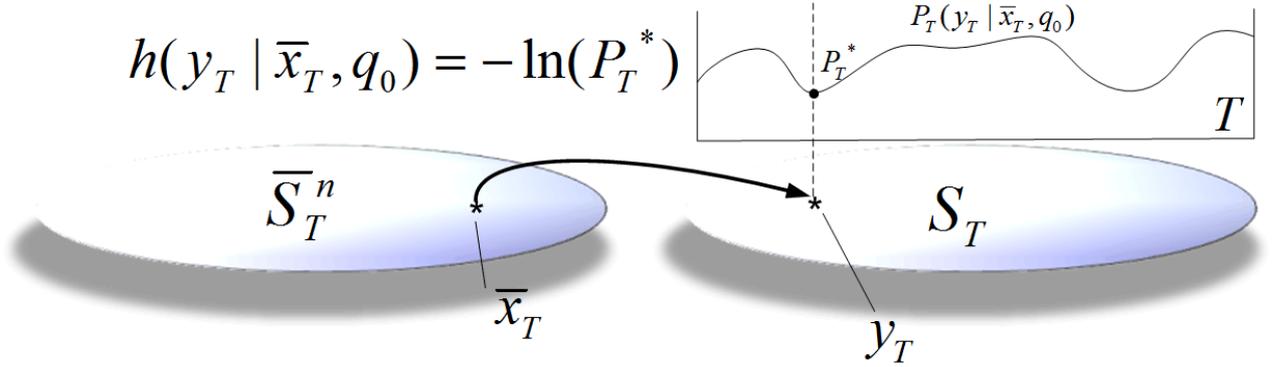

**Fig. 9.** Supervised learning of the generalized spiking neuron using the surprisal minimization. Each transformation of the input pattern $x$ from the multidimensional pattern space into the output pattern $y$ from the one dimensional pattern space is characterized by the probability $P$ and the surprisal value $h$. By minimizing the surprisal value of a particular transformation we can make such transformation more probable.

A supervised learning task can be formalized as the task of minimization of the surprisal of $y_T^d$ conditioned on $\bar{x}_T$ and the initial state $q_0$: $h_T(y_T^d|\bar{x}_T, q_0) \to \min$. By minimizing the surprisal $h_T(y_T^d|\bar{x}_T, q_0)$, the neuron will maximize the probability of generation of $y_T^d$.

The surprisal of $y_T$ on the interval $T$ is equal to the sum of surprisals on the parts of this interval:

$$h_T(y_T|\bar{x}_T,q_0) = \sum_i^l h_{\Delta T_i}(y_{\Delta T_i}|\bar{x}_{\Delta T_i},q_{i-1}) \tag{16}$$

Indeed, using the equation (12) we obtain:

$$h_T(y_T|\bar{x}_T,q_0) = -\ln(P_T\{\bar{y}_T|\bar{x}_T,q_0\}) = -\ln(\prod_i^l P_{\Delta T_i}\{y_{\Delta T_i}|\bar{x}_{\Delta T_i},q_{i-1}\}) =$$

$$= -\sum_i^l \ln(P_{\Delta T_i}\{y_{\Delta T_i}|\bar{x}_{\Delta T_i},q_{i-1}\}) = \sum_i^l h_{\Delta T_i}(y_{\Delta T_i}|\bar{x}_{\Delta T_i},q_{i-1})$$

The task of minimizing the surprisal at $y_T = y_T^d$ can be partitioned as a set of minimization tasks on the intervals $\Delta T_i$ (Fig. 10). Notice that they are conditioned on the neuron's state at the end of each interval. In practice this property is useful only if the interval is split in such a way that the state at the beginning of each part is independent on the events during other parts of the interval. For example, for integrate-and-fire neuron this can be achieved if the beginning of each part coincides with an output spike because the membrane potential will be reset to a fixed value which is independent of the history. In such case the long supervised learning session can be split into several short training iterations.

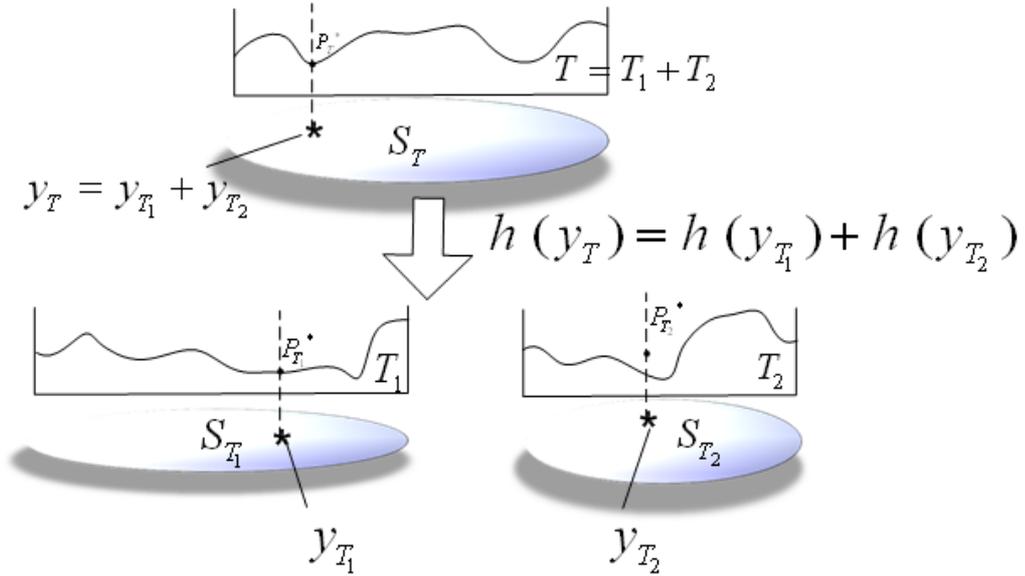

**Fig. 10.** The time additive property of the surprisal on the subsequent intervals. The surprisal of the output pattern on the long time interval is equal to the sum of surprisals of patterns on the parts of this interval provided that the neuron state transition from one interval to another is respected. Each part of the interval belongs to a new one dimensional space of output patterns characterized by the length of the part.

### 1.4.2. Unsupervised learning of the generalized spiking neuron

Another information-theoretic quantity of the neuron operation is the entropy $H_T$ of its output conditioned on the input and the initial state. The entropy on the interval $T$ can be computed as the expectation of surprisal of various output patterns $h(y_T)$:

$$H_T = \sum_{y_T \in S_T} P\{y_T | \bar{x}_T, q_0\} \cdot h(y_T | \bar{x}_T, q_0) \qquad (17)$$

where $h(y_T)$ is the surprisal of $y_T$, $P\{y_T\}$ is the probability of generation of $y_T$ and the summation is done for all possible output patterns. Since $y_T$ is integrated out, the quantity is a deterministic function of the input pattern $\bar{x}_T$ and the initial state $q_0$: $H_T = H_T(\bar{x}_T, q_0)$ (Fig. 11).

$$H_{|\bar{x}_T, q_0} = \sum_{y_T \in S_T} P\{y_T | \bar{x}_T, q_0\} \cdot h(y_T | \bar{x}_T, q_0)$$

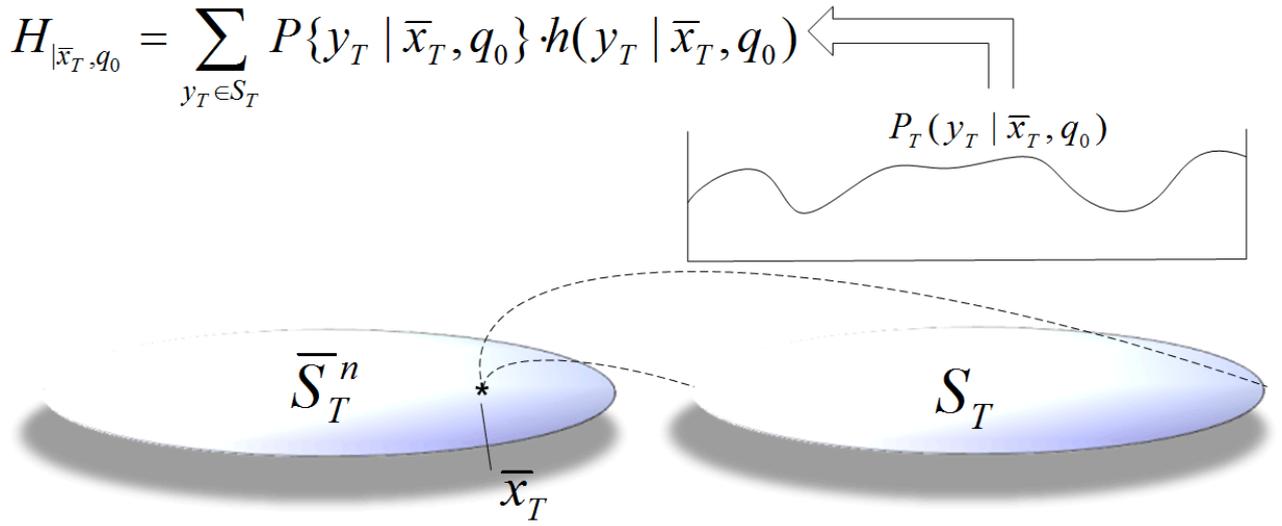

**Fig. 11.** The conditional entropy characterizes the neuron's behavior on the whole space of output patterns. A particular input pattern $x$ from the multidimensional pattern space and a particular initial state of the neuron $q$ defines a probability distribution on the one dimensional pattern space that is characterized by the entropy value.

The entropy $H_T$ characterizes the degree of uncertainty of the neuron's behavior [50]. When the entropy is large, neuron's output almost does not depend on its input. In fact, the entropy is maximized when all $y_T$ are equally likely after the input $\bar{x}_T$ and the neuron generates spikes according to a stationary Poisson process. The smaller the entropy, the more deterministic the neuron's output will be for a given input $\bar{x}_T$. Suppose that the neuron has several approximately equal choices on how to react on the input pattern. If the entropy decreases, the neuron starts to prefer one particular output more and more and the likelihood of other outputs decreases (Fig. 12). Notice that this output can be an empty pattern (the neuron is silent when $\bar{x}_T$ is presented).

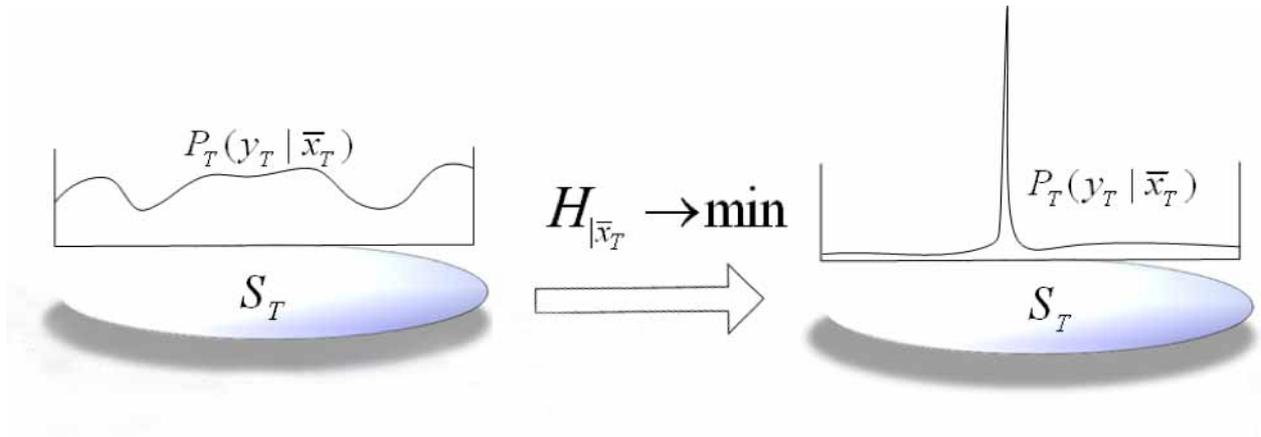

**Fig. 12.** Minimization of the conditional entropy makes the neuron prefer one particular output. A particular input pattern *x* from the multidimensional pattern space and a particular initial state of the neuron *q* defines a probability distribution on the one-dimensional pattern space that is characterized by the entropy value. The minimization of the entropy leads to a peak in the probability distribution and therefore to a stable generation of a particular output pattern (including an empty pattern without spikes).

### 1.4.3. Reinforcement learning of the generalized spiking neuron

A reinforcement learning theory [54] is a way to formalize a "rational behavior" of agents (e.g. animal, robots). An agent and the environment are in a feedback loop: the agent acts on the environment and the environment responds by changing its state and sending reinforcement signals (reward or punishment) back to the agent. The goal of the agent is to learn which action to take depending on the sensory context in order to maximize reinforcement. We will present this task in more details in Chapter 4, but here we can mention that the reinforcement learning methods for the network can be formulated in terms of individual neurons. A particular neuron receives a teaching signal that characterizes the value of its recent actions. Given the input pattern, the neuron can generate various outputs with a different probability (Fig. 13). The output patterns act on the environment via the agent's actuators. The environment changes its state and might respond with a global reinforcement signal. The neuron receives this signal and has to modify its parameters and therefore the probability distribution of its actions in order to receive more reinforcement on average in the future in the similar context:

$$R = \sum_{y_T \in S_T} P\{y_T | \bar{x}_T, q_0\} r(y_T | \bar{x}_T) \to \max \qquad (18)$$

Intuitively, if the reinforcement signal is positive, the neuron has to increase the probability of generating the same output. If the signal negative the probability of the recently generated pattern should be decreased. Notice that length of the interval $T$ controls whether the neuron pays attention only to the immediate rewards on a short time scale or to the cumulative rewards on a longer interval.

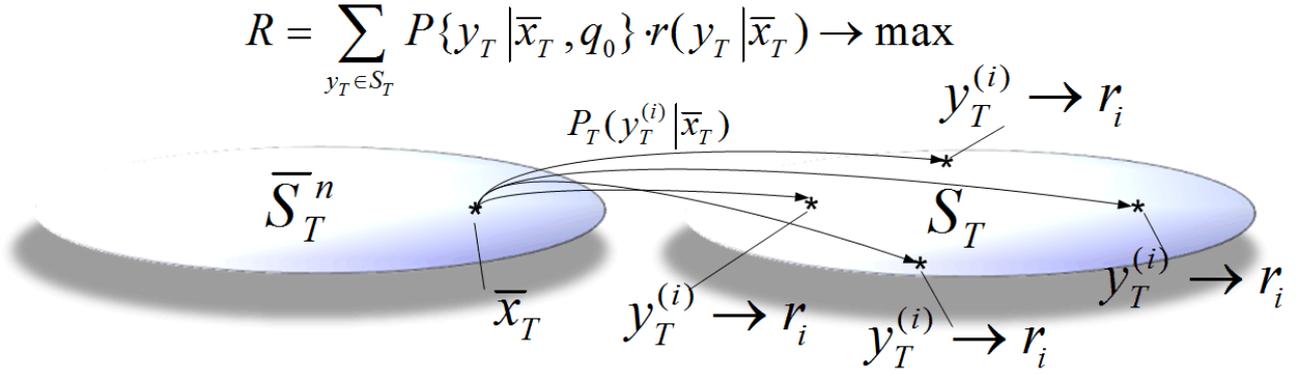

**Fig. 13.** In the reinforcement learning the task of the neuron is to maximize the cumulative reinforcement. In response to a particular input pattern $x$, the neuron tries to generate different output patterns $y$ ("exploration"). It receives different reinforcement signals $r$ depending on the output pattern and sensory context. The neuron needs to shape the input-output probability distribution in order to increase the average amount of received reinforcement.

### 1.5. Spike Multi Response Model

Now let us consider a particular implementation of the generalized spiking neuron model. The model we are going to describe belongs to the class of Spike-Response Models (SRM) [31]. The subset of such models that are described by a linear spike response kernel is marked with index "0" SRM0. The state of SRM0 consists of its membrane potential $u(t)$. The neuron generates spikes if its membrane potential crosses the threshold value $Th$.

First let us consider the Integrate-and-Fire model's differential equations:

$$C_1 \frac{du}{dt} = -u + \sum_{i,j} w_i e^{-(t-t_j^i)/C_2} H(t-t_j^i) + \sum_k (u_{refr} - u)\delta(t - t_k^{out}), \qquad (19)$$

where $C_1, C_2$ are the time constants, $i$ is the index of input channel, $j$ is the index of input spike on each input channel, $w_i$ is the synaptic weight, $t_j^i$ is the input spike time, $H(t)$ is the Heaviside function, $u_{refr}$ is the refractory potential value to which the potential is reset after an output spike, $t_k^{out}$ is the output spike time. In the simplest model described by the equation (1), an input spike instantaneously changes the potential value creating a jump discontinuity. In the model (19) we use a more complicated spike response shape: the potential gradually increases with a finite velocity that depends on the constants $C_1, C_2$. After that the potential again goes to zero.

The equation for the membrane potential for the SRM0 neuron is obtained by integrating equation (19). Since this is a linear differential equation, the solution can be computed as a sum of the responses on the individual spikes. The response function for this model is called "alpha-function" and is computed using:

$$\alpha(t) = \alpha_0 \cdot (e^{-t/\tau_m} - e^{-t/\tau_s}) \cdot H(t) \qquad (20)$$

where $H(t)$ is the Heaviside function, $\alpha_0$ is the scaling coefficient and $\tau_m, \tau_s$ are the parameters that determine the shape of the alpha-function (their value depend on $C_1, C_2$). The refractoriness is defined as the response function on the output spike:

$$\eta(t) = u_{refr} e^{-t/C_1} H(t),$$

where $H(t)$ is the Heaviside function, $u_{refr}$ is the refractory potential value.

The resulting equation for the membrane potential $u(t)$ for the SRM0 neuron can be computed using:

$$u(t) = \sum_i \sum_j w_i \alpha(t - t_j^i) + \sum_k \eta(t - t_k^{out}), \qquad (21)$$

where $i$ is the index of input channel, $t_j^i$ the input spike time, $w_i$ the weight of the $i$-th channel, $t_k^{out}$ the output spike time.

Now we are going to extend this model by adding more alpha-function responses per synapse using the following intuition. In real neurons special chemical substances ("neuromediators") act on the postsynaptic neuron during spike transmission [1]. There are multiple kinds of neuromediators in real neural networks but the role of such diversity is not yet clear [55]. There is evidence [1] that various neuromediators have different effects on the dynamics of the neuron. Different neuromediators can change the membrane potential with different speeds or have different properties of the persistence and the duration of such changes. This hypothesis is supported by some neurophysiological data [56, 57]. Also because of the complex spatial arrangement of dendrites, a simultaneous arrival of the same spike from the same axon on different branches of the dendritic tree might complel temporal effects on the membrane potential (e.g. proximal dendrites will propagate spike faster than distal ones). Based on that let us assume that every of $n$ input channels of the neuron has $m$ synapses with the axon of the other neuron. Each such connection has its own set of dynamic mechanisms (neuromediators or dendritic properties) so that an input spike creates several output responses in the membrane potential with the different time constants $C_{2,k}$. Let us assign a weight $w_{ik}$ to each response mechanism. We are going to call the described extended model of SRM0 neuron as "Spike Multi-Response Model" – SMRM. The modified differential equation for its membrane potential has the following form:

$$C_1 \frac{du}{dt} = -u + \sum_{i,j} \sum_k w_{ik} e^{-(t-t_j^i)/C_{2,k}} H(t - t_j^i) + \sum_k (u_{refr} - u)\delta(t - t_k^{out}), \qquad (22)$$

where the introduced additional summation by the index $k$ is performed for all dynamic mechanisms ($1...m$) with their own weights and time constants.

After the integration of the equation (22) we get a different alpha-function response for every dynamic mechanism:

$$\alpha_k(t) = \alpha_0 \cdot (e^{-t/\tau_{m,k}} - e^{-t/\tau_{s,k}}) \cdot H(t) \qquad (23)$$

where $H(t)$ is the Heaviside function, $\tau_{m,k}, \tau_{s,k}$ are the parameters that depend on the neuron's membrane constant $C_1$ and the time constant of the particular dynamic mechanism $C_{2,k}$, $\alpha_0$ is the scaling coefficient. A single spike arriving at the $i$-th input channel evokes $m$ alpha functions that all change the postsynaptic potential. We are going to call the total influence of a single spike as "spike postsynaptic potential". The spike postsynaptic potential is a weighted sum of all alpha functions from all dynamic mechanisms:

$$PSP_{i,j}(t, t_j^i) = \sum_k^m w_{ik} \alpha_k(t - t_j^i) \qquad (24)$$

where $i$ is the index of the input channel, $j$ is the index of the input spike time on that channel, $t_j^i$ is the input spike time, $\alpha_k$ is the alpha function of the $k$-th dynamic mechanism, $w_{ik}$ is the weight of the $k$-th dynamic mechanism on the $i$-th input channel. The total postsynaptic potential of the input channel is a sum of the postsynaptic potential from all spikes:

$$PSP_i(t) = \sum_{t_j^i \in x_T^i} PSP_{i,j}(t, t_j^i) \qquad (25)$$

The membrane potential is the sum of all input channel postsynaptic potentials and the refractory kernels:

$$u(t) = \sum_i^n PSP_i(t) + \sum_k \eta(t - t_k^{out})$$

If we expand all summations, then the membrane potential value for the described SMRM model is computed by the following equation:

$$u(t) = \sum_i^n \sum_{t_j^i \in x_T^i} \sum_k^m w_{ik} \alpha_k(t - t_j^i) + \sum_k \eta(t - t_k^{out}) \qquad (26)$$

In all experiments in this thesis we simplify the refractoriness term so that it just resets the membrane potential to zero after an output spike. After output spikes all history about input spikes is reset and effects of alpha functions do not propagate further. Such modification simplifies and speeds up computer simulations since one needs to store only recent spike times that came after the last output spike. Here is the modified definition of the alpha function that takes this simplification into account:

$$\alpha_k(t, t^{spike}) = \alpha_0 \cdot (e^{-t/\tau_{m,k}} - e^{-t/\tau_{s,k}}) \cdot H(t) \cdot H(t_{last}^{out} - t)$$

After this simplification the membrane potential equation takes the following form:

$$u(t) = \sum_i^n \sum_{t_j^i \in x_T^i} \sum_k^m w_{ik} \alpha_k(t - t_j^i, t_{last}^{out}); \ u(t_{last}^{out}) = 0 \qquad (27)$$

In the following we always omit the second alpha function argument implying that we cut the alpha function's effect by the last output spike.

The usage of the the set of alpha functions and the vector of weights for each input channel x achieves a greater flexibility in tuning neuron's response to a single spike. With a good set of alpha function's time constants and weights we can change the shape of post-synaptic response in a wide range. This allows us to shape the total membrane potential in order to generate an output spike at the required time depending on a precise timing of input spikes. In particular, we can tune the time of the postsynaptic potential maximums on every input channel. By aligning those maximums, we can make neuron to respond to a complex *n*-dimensional input patterns. Three alpha functions profiles are shown in Fig. 14, top. We plotted 2 examples of their weighted sums (spike postsynaptic potentials) in Fig. 14 (bottom). The input spike creates a different time response profile depending on the weights values which are shown in the center of the plots. The similar method of tuning the timing of the maximum effect on the input signal was also considered by other researchers. In particular, [58] uses a similar set of filtering kernels for a float output neuron. However, to the best of our knowledge we applied this method to a spiking neuron the first time. In [59] the shape of the postsynaptic response was changed by tuning the alpha function time constants. However, the weighted sum of fixed alpha functions can allow more flexible tuning of the postsynaptic potential time profile. The price of this flexibility is that the SMRM neuron has *m* times more weights that the standard SRM0 model, so the set of neuron's weights are defined by the *n×m* matrix.

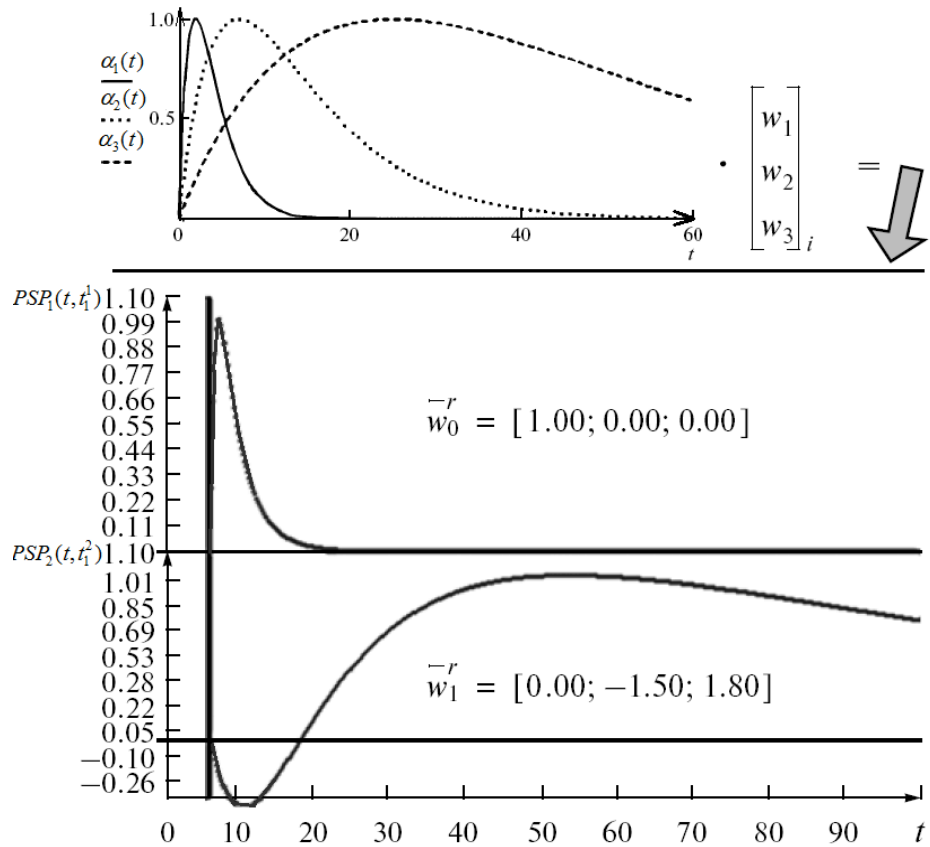

**Fig. 14.** Tuning the amplitude and timing of the postsynaptic potential from a single input spike using the weighted sum of alpha functions. The set of three alpha functions is shown on the top. Two input spikes coming simultaneously on two input channels are shown at the bottom. Time profile of the postsynaptic response is different for every set of weights (weight values are shown in the center of each plots).

Now let us introduce stochastic component into the SMRM neuron and derive an equation for $P_T\{y_T|\bar{x}_T, q_0\}$. There is neurophysiological data showing that the synaptic transmission is a noisy process [60]. There are other noise sources (e.g. thermal noise) that affect neuron's data processing. These facts serve as a justification in order to introduce a stochastic component into the model. Here we are going to use a stochastic threshold for the described SMRM neuron. A stochastic threshold can be viewed as a fuzzy threshold region. When the potential approaches to this region the probability of an output spike increases. Also there is a non-zero probability that a spike will be generated even if the membrane potential is lower than the threshold.

Here we assume that the probability of an output spike depends only on the value of the membrane potential. Let us define a point process intensity function that characterizes the probability of the spike generation at time $t$:

$$\lambda(t) = \lambda(u(t)),$$

where $\lambda(u)$ is the intensity as a function of the membrane potential. In this thesis we chose an exponential form of $\lambda(u)$:

$$\lambda(u) = e^{(u-Th)/\kappa}, \qquad (28)$$

where $Th$ is the threshold value, $\kappa$ is the constant that defines the slope of the exponential threshold and the degree of stochasticity (randomness) of the neuron. Plots of $\lambda(u) = \lambda(u,\kappa)$ for $Th = 1$ and various $\kappa$ are shown in Fig. 15.

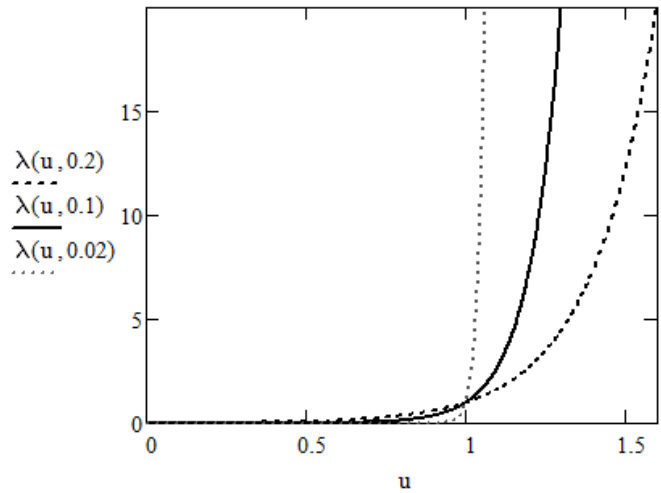

**Fig. 15.** The point process intensity function depends on the value of the membrane potential. Different plots corresponds to the different values of the stochasticity parameter $\kappa$.

The closer the membrane potential is to the threshold, the larger is the intensity function value. For the discrete time case with a time step $\Delta t$ the probability of generating of at least one spike can be computed using (14):

$$\Lambda(t_k) = \Lambda(u(t_k)) = 1 - e^{-\lambda(u(t_k))\Delta t}$$

The plots of $\Lambda(u) = \Lambda(u,\kappa)$ as functions of the membrane potential with $\Delta t = 1$, $Th = 1$ and various $\kappa$ are shown on Fig. 16:

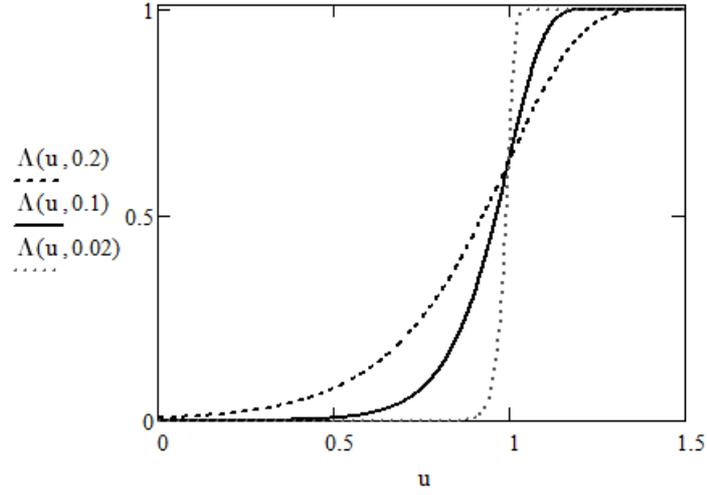

**Fig. 16.** The probability of the output spike in the discrete time neuron model as a function of the membrane potential. Different plots corresponds to the different values of the stochasticity parameter $\kappa$.

If the stochasticity parameter is close to zero, the model becomes deterministic: the neuron is going to generate a spike with probability "1" only if the potential crosses the *Th* value.

The probability of generating $y_T$ with a precision $\Delta t$ can be computed using (11):

$$P(y_T) = \prod_{t_k^{out} \in y_T} \lambda(t_k^{out}) \Delta t \cdot e^{-\int_T \lambda(s) ds}$$

SMRM is an implementation of the generalized spiking model. Its state space $Q$ consists of the current value of the membrane potential, the current weights $w_{ij}$, the history of the input spikes $\{t_j^i\}$, and the time of the last output spike $t_{last}^{out}$ that is taken into account during refractoriness: $Q = \{\{w_{ij}\}, \{t_j^i\}, t_{last}^{out}\}$. Notice that the number of input spike times is finite and is determined by the time of the last output spike and durations of the alpha functions. The state evolution function $F_q$ consists of the membrane potential evolution $u(t) \rightarrow u(t+dt)$ which is based on the input spike times and the weights $w_{ij}$. The probability of generating an output spike is determined by the intensity value $\lambda(t)$. The weights $w_{ij}$ can be changed during the learning, so in general the function $F_q$ also includes the learning rules $w_{ij}(t) \rightarrow w_{ij}(t+dt)$.

# Chapter 2. Supervised learning of spiking neurons

**Summary:**

In this chapter a supervised learning task is formulated for the generalized spiking neuron as a task of minimization of the surprisal of the desired output pattern.

We derived learning rules for the SMRM neuron and performed supervised training experiments that showed that the neuron can be trained to generate spikes at the desired times. We built plots of the surprisal minimization and the evolution of the membrane potential before and after the training.

In this chapter we developed an autoassociative memory network that is able to store several spatiotemporal spiking patterns and then recall them based on the initial segment of the pattern. In order to store the patterns we used the developed supervised learning rules for the SMRM model. We used the spiking pattern distance function to estimate the quality of the recall.

The memory network has been tested on the task of picture drawing on $8 \times 15$ pixel canvas. The network was able to predict and reconstruct the motion of a virtual pen for the whole image based on initial strokes consisting only of 2-3 pixels. It is shown that if the specific set of parameters is used, the network can act as a short term memory: it can memorize spatiotemporal patterns after the first presentation.

## 2.1. Supervised learning of the generalized spiking neuron in discrete time

Consider a supervised learning task in which the neuron has to generate a desired output pattern $y_T^d$ in response to the input pattern $\bar{x}_T$. First we are going to develop the learning rules for learning in discrete time. As it was shown in chapter 1, this task can be formulated as a task of surprisal minimization of the output pattern $y_T = y_T^d$:

$$h_T(y_T|\bar{x}_T, q_0) = -\ln(P(y_T|\bar{x}_T, q_0)) \to \min.$$

The surprisal $h_T(y_T|\bar{x}_T, q_0)$ is conditioned on the input pattern $\bar{x}_T$ and neuron's initial state $q_0$ on the interval $T$. Below we are going to omit the conditionals assuming that all calculations are performed with some fixed $\bar{x}_T$ and $q_0$.

The probability of the pattern $y_T^d = \{t_i^d\}$ on the interval $T$ can be computed using (4):

$$P(y_T^d) = \prod_{t_i \in y_T^d} \Lambda(t_i) \cdot \prod_{t_i \notin y_T^d}(1 - \Lambda(t_i)),$$

where $\Lambda(t_i)$ is the probability of spike generation on the $i$-th step. The probability of generating $y_T^d$ is equal to the product of probabilities that the neuron generates spikes at time steps $t_i \in y_T^d$ and does not generate spikes at other times $t_i \notin y_T^d$. Notice that $\Lambda(t_i)$ are not independent as they are all conditioned on the particular evolution of the neuron's state.

The surprisal of the desired output pattern is equal to:

$$h(y_T^d) = -\ln(P(y_T^d)) = -\sum_{t_i \in y_T^d} \ln(\Lambda(t_i)) - \sum_{t_i \notin y_T^d} \ln(1 - \Lambda(t_i)) \tag{29}$$

The probability of a spike can be computed using (14):

$$\Lambda(t_i) = 1 - e^{-\lambda(q(t_i))\Delta t},$$

where $\lambda$ is the point process intensity, $q(t_i)$ is the state of the neuron on the $i$-th time step. Below we are going to simplify the notation as follows: $\lambda(q(t_i)) \equiv \lambda_i$, $\Lambda(t_i) \equiv \Lambda_i$. The equation for the surprisal will take the form:

$$h(y_T) = -\sum_{t_i \in y_T^d} \ln(\Lambda_i) + \sum_{t_i \notin y_T^d} \lambda_i \cdot \Delta t$$

The minimization of the surprisal can be done using a stochastic gradient descent with respect to the neuron's parameters. Let us compute a derivative of the surprisal with respect to the parameter $w$:

$$\frac{\partial h(y_T^d)}{\partial w} = \sum_{t_i \in y_T^d} \frac{1}{\Lambda_i} \frac{\partial}{\partial w} e^{-\lambda_i \cdot \Delta t} + \sum_{t_i \notin y_T^d} \frac{\partial \lambda_i}{\partial w} \cdot \Delta t =$$

$$= \sum_{t_i \in y_T^d} \frac{-e^{-\lambda_i \cdot \Delta t}}{\Lambda_i} \frac{\partial \lambda_i}{\partial w} \Delta t + \sum_{t_i \notin y_T^d} \frac{\partial \lambda_i}{\partial w} \cdot \Delta t = \qquad (30)$$

$$= \sum_{t_i \in y_T^d} \frac{\Lambda(t_i)-1}{\Lambda(t_i)} \frac{\partial \lambda(t_i)}{\partial w} \Delta t + \sum_{t_i \notin y_T^d} \frac{\partial \lambda(t_i)}{\partial w} \cdot \Delta t$$

The derivative of the intensity $\lambda(t)$ is equal to:

$$\frac{\partial \lambda(t)}{\partial w} = \frac{\partial \lambda(q)}{\partial q} \frac{\partial q(t)}{\partial w}, \qquad (31)$$

where $q(t)$ is the state of the generalized spiking neuron.

The change in the parameters $w$ after training on the interval $T$ is equal to:

$$\Delta w^T = -\gamma \cdot \frac{\partial h(y_T)}{\partial w}, \qquad (32)$$

where $0 < \gamma < 1$ is the learning coefficient. By substituting the derivative we get the equation for the change in parameters of the generalized spiking neuron:

$$\Delta w^T = \gamma \Delta t \cdot \left( \sum_{t_i \in y_T^d} \frac{1-\Lambda(t_i)}{\Lambda(t_i)} \frac{\partial \lambda(q(t_i))}{\partial q} \frac{\partial q(t_i)}{\partial w} - \sum_{t_i \notin y_T^d} \frac{\partial \lambda(q(t_i))}{\partial q} \frac{\partial q(t_i)}{\partial w} \right) \qquad (33)$$

These learning rules are applicable to any neuron model if its possible to compute $\frac{\partial q(t)}{\partial w}$ for any time step and also if $\Lambda(t_i)$ is always larger than zero. Such requirements are satisfied for the developed SMRM model. The parameters $w$ are neuron's weights $w_{ik}$, where $i$ is the input channel index, $k$ is the alpha function index. Since the probability of a spike depends on the membrane potential which is a sum of alpha functions, the derivative of the state with respect to the weights can be computed as:

$$\frac{\partial q(t)}{\partial w_{ik}} = \frac{\partial u(t)}{\partial w_{ij}} = \sum_{t_j^i \in x_T^i} \frac{\partial}{\partial w_{ik}} \sum_i^n \sum_k^m w_{ik} \alpha_k(t-t_j^i) = \sum_{t_j^i \in x_T^i} \alpha_k(t-t_j^i)$$

The resulting learning rules for weights $w_{ik}$ of the SMRM neuron take the following form:

$$\Delta w_{ik}^T = \gamma \Delta t \cdot \left( \sum_{t_l \in y_T^d} \frac{1-\Lambda(t_l)}{\Lambda(t_l)} \frac{\partial \lambda(u(t_l))}{\partial u} \sum_{t_j^i \in x_T^i} \alpha_k(t_l - t_j^i) - \sum_{t_l \notin y_T^d} \frac{\partial \lambda(u(t_l))}{\partial u} \sum_{t_j^i \in x_T^i} \alpha_k(t_l - t_j^i) \right) (34)$$

This equation consists of two terms. The first term depends on the times of the desired spikes and defines the associative weight increase that happens when the input spike comes before the desired spike. The second term depends only on the input spikes and defines nonassociative weights decrease. This term tries to decrease the overall membrane potential to avoid generating spikes at all intervals except the desired spike times.

## 2.2. Implementation of the supervised learning rules

In this chapter we describe some details of implementation of the derived learning rules. We are going to describe two particular implementations: with and without simultaneous operation during training.

In order to implement the derived rules (34) we need to choose the interval $T$ at the end of which the neuron changes the weights. Since the neuron's state is reset after an output spike let us split the training samples on intervals $\Delta T_i$, where each interval begins after the desired output spike $t_i^d \in y_T^d$ and ends on the next desired spike $t_{i+1}^d \in y_T^d$, so that $\Delta T_i = (t_i^d, t_{i+1}^d]$. Since the surprisal is additive in time according to (16) we can minimize it on each interval. We can perform the weight changes using (34) at the end of every interval at times $t_i^d \in y_T^d$. Notice that this introduces an extra bias in the stochastic gradient descent estimation since we are going to do more weight updates comparing to the batch update at the end of the full interval. However, this allows us to represent the task differently. Let us introduce the concept of a "teaching" input channel of the neuron. The neuron receives "teaching" spikes at times $t_i^d$ on this channel according to the desired output pattern $y_T^d$. The arrival of a teaching spike makes the neuron to perform the weight updates according to (34). Using a separate teaching spikes in training of spiking neurons is not new. For example, it was used in the heuristic supervised learning rules ReSuMe [38]. ReSuMe also uses associative weight increase when the teaching spike arrives and non-associative weight decrease when the neuron is silent. However, the usage of the teaching input in this thesis now have a theoretical basis in terms of the task of the surprisal minimization.

Below we are going to describe in details two different implementations of learning with "teaching" spikes.

**Supervised algorithm with disabled spike generation.**

Spikes that the neuron generates at times $t^{self}$, that do not belong to the desired output pattern $y_T^d$ are going to be called "false positives" below. Notice that the computation of the probability distribution of the desired output pattern is correct only when the neuron actually generates spikes at the necessary time steps $t_i^d$. Otherwise the state evolution is not going to be the same: resets of the potential at times other than $t_i^d$ are going to alter the state evolution. One way to deal with this during the training is to completely disable the spike generation by the neuron itself and generate spikes only when the teacher spike arrives so that state evolves correctly. During the testing stage we turn back on the stochastic spike generation and turn off the weight changes. Such method requires explicit training and testing stages and can not be used in online learning setup.

Let us describe a simulation process in this case. Let's denote a current time step with $t_l$. First, we compute the membrane potential $u(t_l)$ based on on the previous and current input spikes. Then we compute $\frac{\partial \lambda(u(t_l))}{\partial u}$. Then for every weight we compute the gradient of the surprisal $g_{ik}$. In the absence of the teaching spike the gradient increases because the alpha function and intensity derivative are both positive:

$$g_{ik} = g_{ik} + \frac{\partial \lambda(u(t_l))}{\partial u} \sum_{t_j^i \in x_T^i} \alpha_k(t_l - t_j^i).$$

If there is no teaching spike we continue to the next simulation step.

If the teaching spike arrived at this time step, we compute the probability of the output spike $\Lambda(t_l)$ and decrease the gradient value:

$$g_{ik} = g_{ik} - \frac{1 - \Lambda(t_l)}{\Lambda(t_l)} \frac{\partial \lambda(u(t_l))}{\partial u} \sum_{t_j^i \in x_T^i} \alpha_k(t_l - t_j^i),$$

After that we change the weights:
$$w_{ik} = w_{ik} - \gamma \cdot g_{ik} \Delta t,$$
and reset the gradient values $g_{ik} \leftarrow 0$. The neuron generates output spikes, after that the neuron's state is reset and we can continue to the next simulation step.

**Supervised algorithm with enabled spike generation.**

If supervised learning has to happen in online settings together with operation, we use the following approximate algorithm. We can notice that a false positive spike has a limited influence on the neuron's state which is determined by the alpha functions and refractoriness parameters. Therefore, if a false positive spike happened long time ago we can ignore it. However, if the neuron generates it close to a desired teaching spike, it affects the gradient value because of the state reset. In this case it is necessary to increase the surprisal of the false positive spike pattern. To do that we can change the weights along with the gradient $\frac{\partial h(y_T^{self})}{\partial w_{ij}}$, changing update sign in (34). By increasing the surprisal we decrease the probability of generating such pattern. The lack of false positive spikes allows us to use rules (34) again. However, here we make another approximation and use rules (34) even if false positive spikes have not completely disappeared.

Let us describe the simulation process in this case. As before, let's denote a current time step with $t_l$. First, we compute the membrane potential $u(t_l)$. Then we compute $\lambda(u(t_l))$, $\Lambda(t_l)$, $\frac{\partial \lambda(u(t_l))}{\partial u}$. However now, depending on the output of a random number generator and $\Lambda(t_l)$, we determine whether the neuron generates a spike at this time step.

If neuron generates a spike itself and the teaching spike arrives at the same time, we do not change the weights since the neuron already performs the task and the gradients are reset to zero. If there is no teacher nor output spike the gradient increases:

$$g_{ik} = g_{ik} + \frac{\partial \lambda(u(t_l))}{\partial u} \sum_{t^i_j \in x^i_T} \alpha_k(t_l - t^i_j).$$

If there is a false positive spike or there is a teaching spike the gradient decreases:

$$g_{ik} = g_{ik} - \frac{1 - \Lambda(t_l)}{\Lambda(t_l)} \frac{\partial \lambda(u(t_l))}{\partial u} \sum_{t^i_j \in x^i_T} \alpha_k(t_l - t^i_j).$$

However, if there is only a teaching spike the weights change against the gradient:

$$w_{ik} = w_{ik} - \gamma \cdot g_{ik} \Delta t.$$

If there is only a self-generated false positive spike the weights change along with the gradient:

$$w_{ik} = w_{ik} + \gamma \cdot g_{ik} \Delta t.$$

If learning took place, the gradients are reset $g_{ik} \leftarrow 0$. After that we continue to the next simulation step.

### 2.3. Learning a desired delay between an input and output spike

Consider the simplest task of generating an output spike after a single input spike. Notice that using only a single alpha function puts limitations on the possible output spike times $t^{out}$. In particular the neuron can not learn the task of generating an output spike after the input spike if the time delta $t^{out} - t^{in}$ is larger than the time of the alpha function maximum: the increase of the weight will cause a generation of a false positive spike before the alpha-function maximum; the decrease of the weight will make the neuron silent. As we will see below, if we add several more alpha functions with gradually increasing times of maximums we can achieve a quite precise timing of the output spike. Moreover, it is even possible to teach the neuron to generate a spike after the time of the most distant alpha function maximum.

Let us consider a supervised learning task for the neuron with 3 alpha functions plotted in Fig. 14. The neuron receives an input spike at $t^{in} = 0$ and has to generate the output spike at $t^{out}$. The results of three experiments with learning various time intervals between the input and output spike $\Delta t_1 < \Delta t_2 < \Delta t_3$ are shown on Fig. 17. The surprisal value is plotted after every training trial. At the beginning of each trial the neuron received the input spike $t^{in} = 0$. Later in the trial the neuron received the teaching spike at $t^{out}$. The neuron has the ability to generate false positive spikes according to the stochastic threshold mechanism. However, the stochasticity parameter was chosen close to zero so that the neuron behaved almost like a deterministic one. The weights changes were performed according to the described algorithm with enabled spike generation. It can be seen from the surprisal plots that the value always decreases and the most abrupt decrease happens after only 7-10 training iterations. After the initial fall the entropy decreases slowly and the value becomes noisier because of the large amount of false positive spikes. The speed of training is different for various delay intervals $\Delta t_i$. It can be explained by different behaviors of the alpha functions at the point of the desired spike (Fig. 17 (right, top)). In order to learn the first delay $\Delta t_1$, the neuron can use all 3 alpha functions, while for learning the $\Delta t_2$ the impact of the first alpha function is smaller.

For the third experiment we chose an interval $\Delta t_3$ that is two times larger than the time of the maximum of the last alpha function. As noted before, the training would not be possible in the absence of other alpha functions: neuron would just start generating spikes near the time of maximum. However, in the bottom part of Fig. 17 we can see that the weights are tuned such that the first two alpha functions counteract the last alpha function at the beginning so no false positive spikes are generated. After the supressing effect is gone the last alpha function's tail makes the membrane potential raise at the desired time and generate the spike.

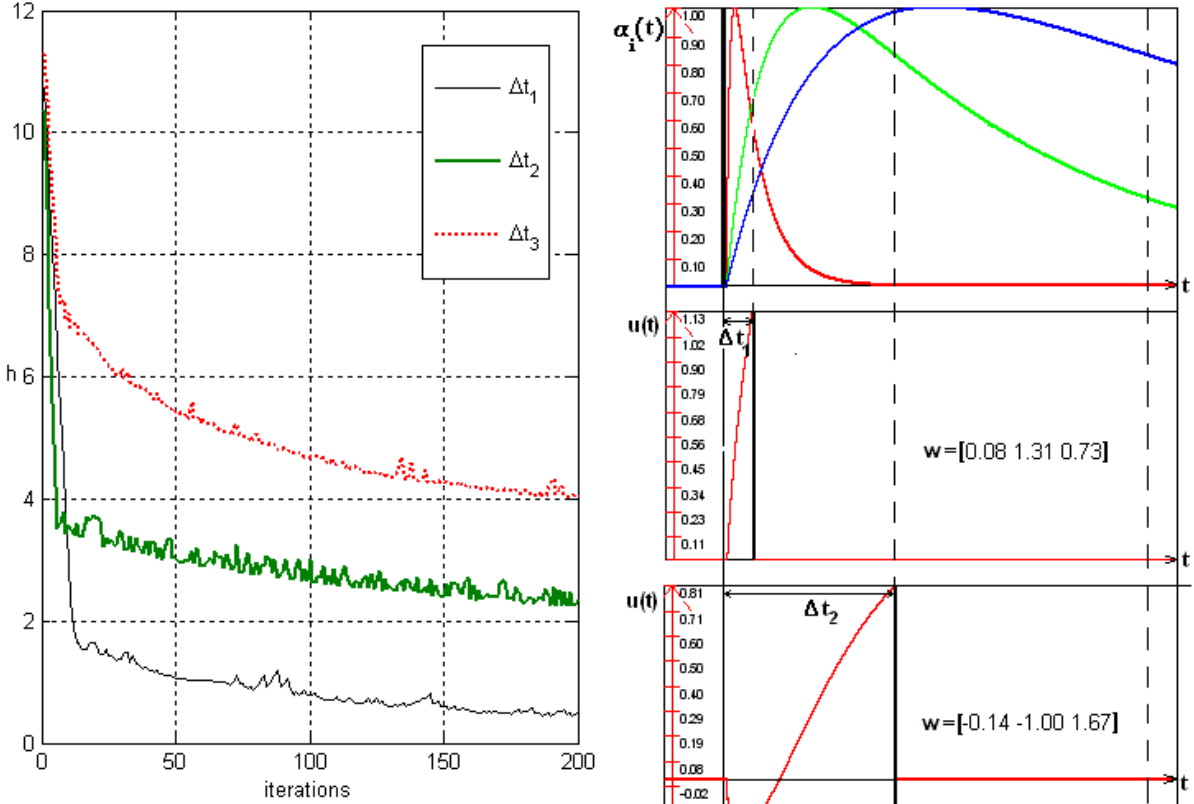

**Fig. 17.** The surprisal values during the training are shown on the left for three tasks of learning a delay. Three alpha-functions are shown on the right-top. The resulting membrane potential and the weights after training for the three tasks are shown on the right, below the alpha functions. The membrane potentials are shaped to produce the output spike after the required delay.

### 2.4. A pattern detection task

The task of generating a desired output pattern $y_T$ based on the input pattern $\bar{x}_T$ can be formulated as a set of tasks of generating a single output spike in response to the parts of the input pattern. Let's call an input pattern "simple" if it is desired to respond on it with a single output spike. Therefore, any supervised learning task can be formulated as a sum of tasks of detecting simple patterns with output spikes. The neuron is a stochastic pattern detector that detects some of the patterns with a spike and is silent in response to other patterns with a certain probability.

The result of training in the task of detecting a 5-channel pattern in a noisy stream of spikes is shown in Fig. 18. The stream of spikes shown in Fig. 18 (top) was received by the neuron on every training iteration. The pattern that neuron needs to detect is marked by a dashed rectangle. The desired delay $\Delta t$ between the beginning and the desired output spike is also shown in Fig. 18.

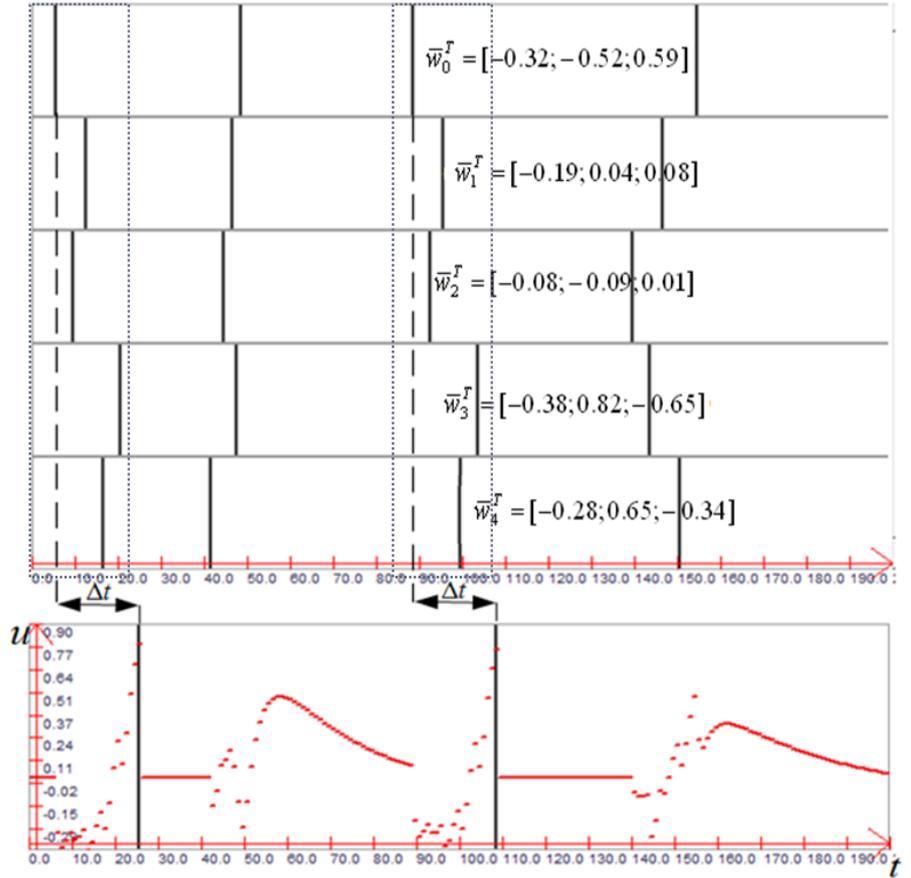

**Fig. 18.** The result of training in the pattern detection task. The input pattern that the neuron is required to detect is marked by a dashed rectangle. Other input spikes are noisy distractors. The resulting vector of alpha function weights after the training is shown for every input channel. The resulting membrane potential plot is shown at the bottom with dashed red line. The output spikes after the required pattern are shown with black bold lines in the bottom membrane potential plot.

During the training the neuron received a teacher spike after the target input pattern. This led to the weight changes according to the rules (34). This in turn led to the increased probability of the output spike after the target input pattern and the decreased probability of the spike after the distracting noisy patterns. The membrane potential after the training is shown on the bottom of Fig. 18. It can be seen that the membrane potential gets close to the threshold only after the selected pattern. The model has 3 alpha functions on each input channel so the total number of weights is $5 \times 3 = 15$. The weighted summation of alpha functions allowed the neuron to tune the delay $\Delta t_i$ and value $\Delta u_i$ of the postsynaptic potential maximums for every input channel. Correctly tuned values of five $\Delta t_i$ and $\Delta u_i$ corresponding to the temporal structure of the input pattern allowed the neuron to react on it with output spikes. The distractor patterns that also consisted of 5 spikes in all channels but did not have a required temporal structure were ignored. Therefore, the neuron detects the complete spatiotemporal structure of the pattern and not only the number of spikes per time interval (the firing rate) in the input stream.

The teacher spike indicates to the neuron which input pattern it should detect. Let's consider in details the process of training of detecting a very simple pattern. Consider a neuron with only 2 input channels and the input pattern that consists of two spikes (one spike per channel). The second spike comes after the first spike after the delay $\Delta t^{in}$. The task is to generate an output spike after the second input spike with the delay $\Delta t^{out}$ (see Fig. 19).

During the training the probability of the output spike $P(t)$ is shaped to have a maximum at the desired times and to be as low as possible at all other times. However, it is up to the neuron which input data it should use to achieve this shape of the probability distribution. On Fig. 19 (A) we show a result after the training on the simplest scenario: the input pattern $\bar{x}_T$ arrives and then the teacher sends a teaching spike. At the bottom plot in Fig. 19 (A) we show a probability of the output spike $P(t)$ after 20 training iterations in this simplest scenario. It can be seen that $P(t)$ does have a correct shape with the maximum at the desired time. The simulation discretization step is 1ms. If we sum the probabilities on interval $\Delta t_m$ = 3ms that is centered on the desired spike time we get the value of the probability of generating the output spike during this interval $P_{\Delta t_m}(y_T | \bar{x}_T) \approx 0.91$.

On the top plots in Fig. 19 (A) we show the probability of the spike generation $P(t)$ if only a single spike from the pattern is presented. Let's denote the first spike $x_T^1$ and the second $x_T^2$. From these plots you can see a drastic inequality of contributions from the $x_T^1$ and $x_T^2$ to the output spike generation $y_T$. The first spike $x_T^1$ has almost no effect on the output spike generation: $P_{\Delta t_m}(y_T | x_T^1) \approx 0.004$. While the second input spike can cause the output spike generation with a high probability: $P_{\Delta t_m}(y_T | x_T^2) \approx 0.86$. We can conclude that the neuron mainly uses only the second part of the pattern $x_T^2$ to spike. The first spike $x_T^1$ just serves the purpose of suppressing the probability of generating an output spike before the desired time with the help of negative alpha functions weights. The neuron is in fact a detector of $x_T^2$ and does not distinguish between $x_T^2$ and the full $\bar{x}_T$.

The described behavior might be undesirable if the task is to detect the whole pattern in a noisy stream of spikes. In the previous scenario neuron does not have the information that only the whole pattern has to be detected. This can be solved with additional training iterations directed to suppress the false positives explicitly. In fig. 19 (B) we present the results of a more complex training procedure. Every training epoch consists of three steps. First the neuron receives the full pattern $\bar{x}_T$ and then the desired teaching spike $y_T$. Then the neuron is presented with the parts $x_T^1$ and $x_T^2$ and does not receive any teaching spikes. This indicates to the neuron that the desired behavior is to not generate spikes at all (generate an empty pattern $y_T^0$). Therefore the neuron receives additional information that it should not detect the parts $x_T^1$ or $x_T^2$ separately. The plots of the probability of the output spike $P(t)$ after such training are shown in Fig. 19 (B). You can see that the probability of a spike after presenting only $x_T^1$ or $x_T^2$ is close to zero. However the probability of the spike after the full input pattern is still high $P_{\Delta t_m}(y_T | \bar{x}_T) \approx 0.82$.

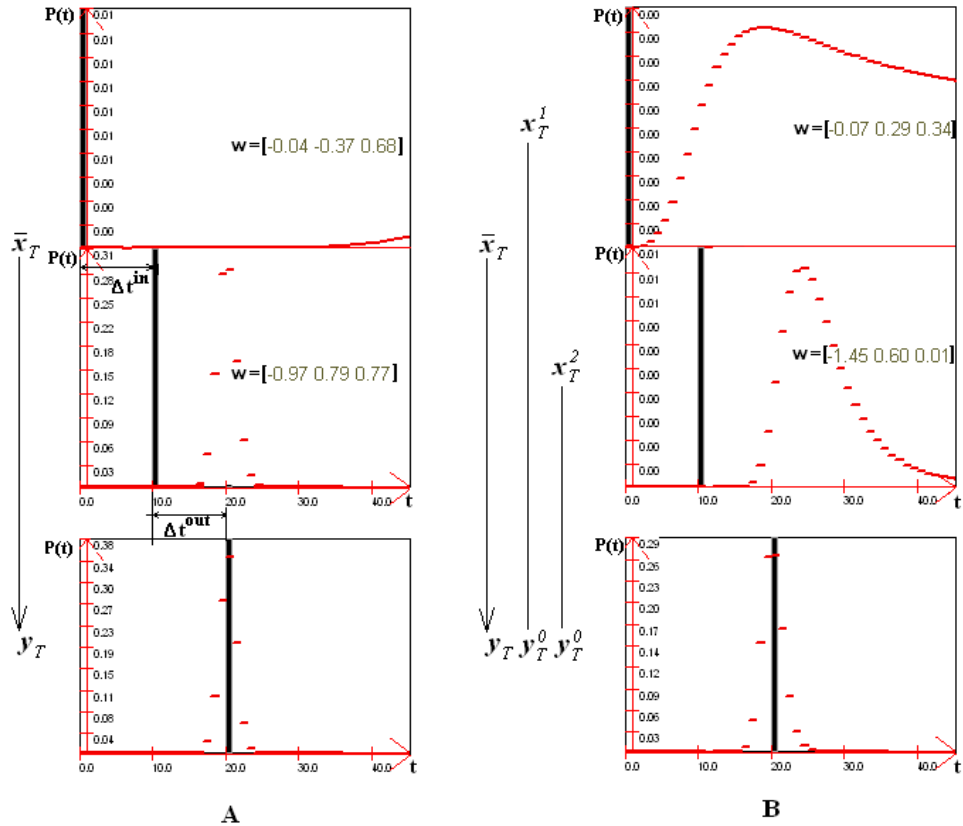

A       B

**Fig. 19.** The task of a simple pattern detection in different training scenarios. A – a simple scenario without the false positives suppression. B – a scenario with the false positives suppression. In both scenarios the neuron is trained to detect a pattern consisting of two spikes in two channels. Input spikes are shown with the black bold lines on the top of two plots separated by the interval $\Delta t^{in}$. The desired output spike is shown on the bottom plots with the delay $\Delta t^{out}$ after the last input spike. The dashed red lines show the probability of the output spike after training. At the bottom plots the probability is plotted when the whole pattern is presented. At the top plots the probability of the output spike given only a single input spike is shown. After the training using scenario A (left) the probability of an output spike after the second input spike only is as large as after the whole input pattern (notice the scale of the plots). After the training using scenario B (right) the probability of a spike after the whole pattern is large, while the probabilities of an output spike after a single input spike is really small (see the scale of the plots).

## 2.5. The spatiotemporal autoassociative memory

Autoassociative memory is a content-addressable memory where the item recall is based on the similarity between the input and stored data. In our case the data item is a multidimensional spiking pattern. If the memory network receives a corrupted spike pattern it should recall the original non-corrupted pattern. The set of possible "corruption" events is quite large: missing spikes, extra spikes, spike shifts, incomplete sequences etc.

One of the first developed autoassociative memory network is J. Hopfield's network [8]. This network is capable of restoring spatial binary image data. The network consists of the recurrently connected binary neurons. During the recording the weights are changed according to the Hebb's rule using which the neurons store the spatial associations between pixels. During the recall network receives a corrupted pixel image and restores it to the closest image in the memory using the spatial associations. Notice that the network has a non-trivial recurrent dynamics during the recall but can store only spatial patterns. In order to store temporal patterns some researches used extensions of J. Hopfield's ideas. For example, it was shown in [61] that the network of binary neurons with unstable internal dynamics can create attractors that allow it to store and recall sequences of impulses. However, the usage of binary neurons without a short-term memory does not allow the network to store delays between the impulses. There exist few models of spiking autoassociative memory that use heuristic learning rules [62, 63]. In [64] the developed network is able to record patterns from the first trial, however it has a complex multilayer structure, a lot of free parameters and heuristic learning rules. The network developed in this thesis is able to store patterns with delays between spikes, it has one simple recurrent layer and uses learning rules derived from the first principles. It will be shown that such network is also capable of recording a pattern from the first trial.

### 2.5.1. Graphical notation for describing spiking networks

For the simulation of the spiking networks we used a custom developed software described in Appendix 1. Below we are going to use graphical notation that is based on the user interface of this software (Fig. 20).

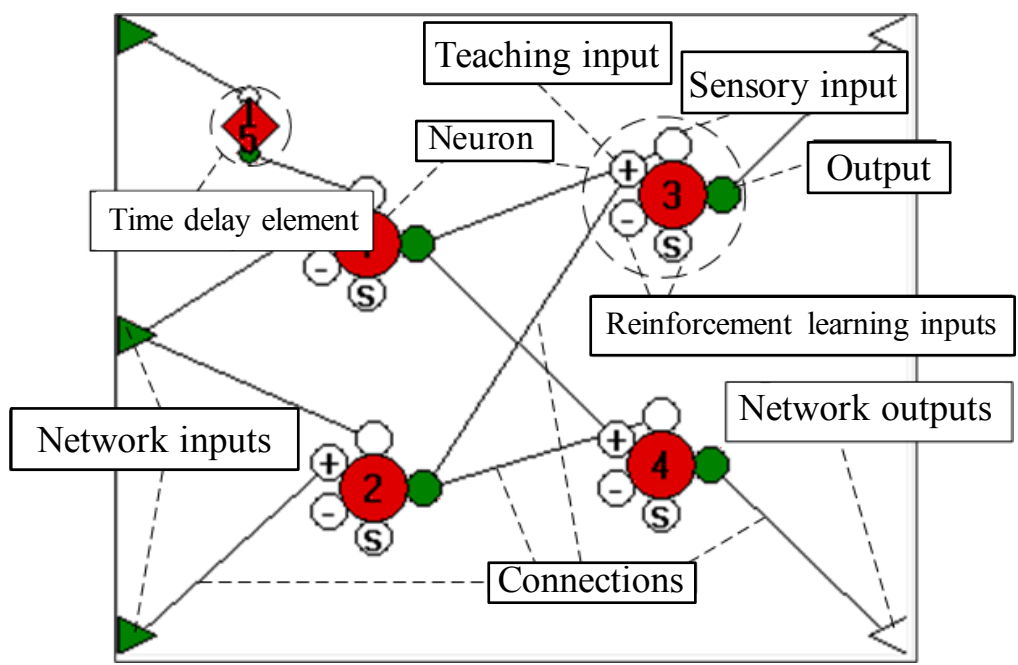

**Fig. 20.** The graphical notation for the elements of the spiking neural network.

Neuron's body is denoted with a red circle with its index. Its connection pins are denoted with smaller circles attached to its body. Output pins are placed on the right side of the body, and input pins are placed on the left. A sensory spiking input pin is denoted with an empty white circle. The neuron uses spikes coming on this input to generate output spikes. An input circle with a "+" is a teaching input. Spikes coming on this channel trigger learning (e.g. supervised surprisal minimization). Inputs with "-" and "s" are used during reinforcement learning and will be described in chapter 4. Connection lines between the pins are shown with solid black lines. Input sensory signals from the environment are denoted by triangles on the left side of the picture. Outputs into the environment are shown with triangles on the right side of the picture. Also there are spike delay elements that are shown with small diamonds. The value of the delay in time steps is displayed at the top of the diamond.

### 2.5.2. Training the spatiotemporal memory network

Spikes are a single type of a propagating signal in the networks developed in this thesis. A single object that generates spikes can be a source of the sensory input signals or a source of the teaching spikes (or both) depending on the connectivity. This means that one neuron can be a teacher for another neuron so supervised learning can be implemented between the neurons themselves and without any external supervised signal. In this case the network will look like an unsupervised adaptive system. Consider the simplest unsupervised learning task for the network with a single external input and a single neuron. Let's connect the external environmental input to the neuron's sensory input. Also let's connect the external input to the neuron's teaching input via a delay element (Fig. 21).

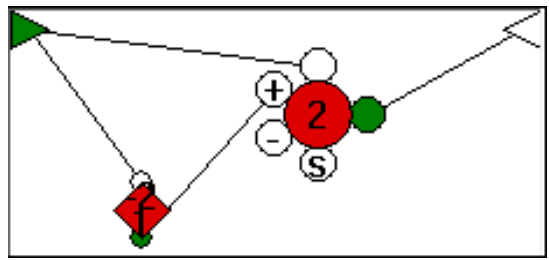

**Fig. 21.** Unsupervised learning to delay a spike with a simple spiking network. The network consists of the single neuron and the delay element. A spike from the network inputs comes to the sensory input of the neuron. Then after the delay this spike arrives at the supervised input of the neuron. The neuron learns to delay the input spike using its internal dynamics and the delay element can be removed after the training.

From the neuron's point of view every input spike is followed by a teaching spike after a specific delay. Therefore, this network will learn to repeat the input spikes with a specified delay.

Now let's construct similar network with *n* sensory inputs. We assign a neuron to every input channel. The task for the network is to learn the input pattern $\bar{x}_T = \{t_j^{i_j}, i_j \in 1..n, j = 1...m\}$, consisting of *m* spikes on *n* channels, where *j* is a spike index and $i_j$ is the index of the channel of the *j*-th spike. Let us connect every input channel with the corresponding neuron's sensory input and teaching input pin like it was done for the neuron shown in Fig. 21 with a delay element of 1 (not shown in the figure). As a result, every input spike that comes to the neuron at time $t_j^{i_j}$ will also arrive at its teaching input at the next step. From the neuron's point of view it will look like a teacher wants the neuron to generate an output spike immediately after the input spike $t_j^{i_j}$. Now let us connect outputs of all neurons with their sensory input (all-to-all connections). The resulting network for *n*=5 is shown in Fig. 22. Without the all-to-all connections the network can learn to repeat the input pattern $\bar{x}_T$ with a single step delay. However now neurons can use spikes from other neurons: if a corrupted input pattern $\bar{a}_T \approx \bar{x}_T$ has a missing spike $t_k^{i_k}$, then the $i_k$-th neuron can use the activity of all other neurons that repeated the previous spikes of the pattern $\{t_j^{i_j}, i_j \in 1..n, j = 1...k\}$ to restore the missing spike. Therefore, every neuron tries to remember the timing of its spike $t_k^{i_k}$ in the pattern $\bar{x}_T$ in relation to the activity of other neurons generating earlier spikes in $\bar{x}_T$. Such network is capable of restoring original patterns based on the initial spikes. For example, after the first two spikes of the pattern the third neuron will be active based on the activity of the first two neurons even in the absence of the external spike. Soon the fourth neuron is going to be active and so on until the end of the pattern. As a result, the whole pattern will be restored based on just two spikes. Since we use the SMRM that is sensitive to the spatiotemporal structure of the patterns we can store multiple spike patterns that start with spikes coming in different order.

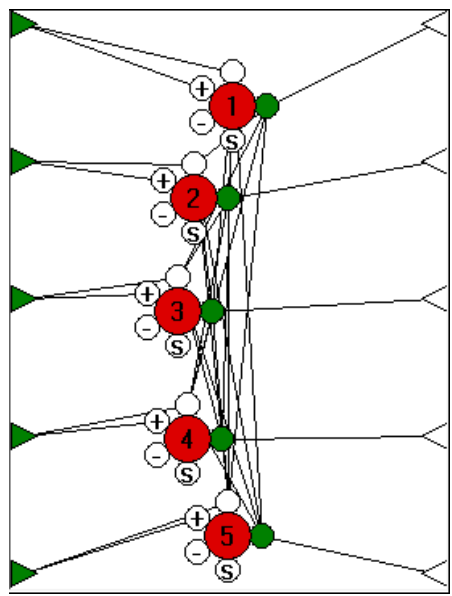

**Fig. 22.** An autoassociative spatiotemporal memory network with 5 input channels. The neurons are connected all-to-all: the output of every neuron is connected the sensory input of every other neuron.

The sensory input spikes arrive at the sensory and at the supervised inputs of the neurons. Every neuron learns to predict the external sensory spikes using the activity of other neurons.

Let's describe a process of learning and recall for the network shown in Fig. 22. The pattern to store is shown in Fig. 23.

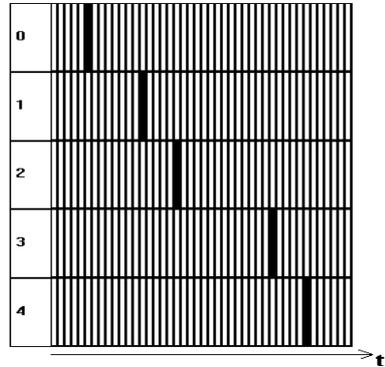

**Fig. 23.** The pattern used to test storing and recall properties of the spiking memory network. The pattern consists of 5 spikes subsequently appearing in the 5 channels with different interspike intervals.

First we present the pattern from Fig. 23 to the network $k$ times. After that we fix the network's weights. After $k=5$ the network is already capable of repeating the pattern but is not able to fill in the missing spikes. After $k=10$ network is able to restore some missing spikes but it also generates extra spikes which are not present in the original pattern. Example of the input and restored pattern after $k=10$ is shown on Fig. 24.

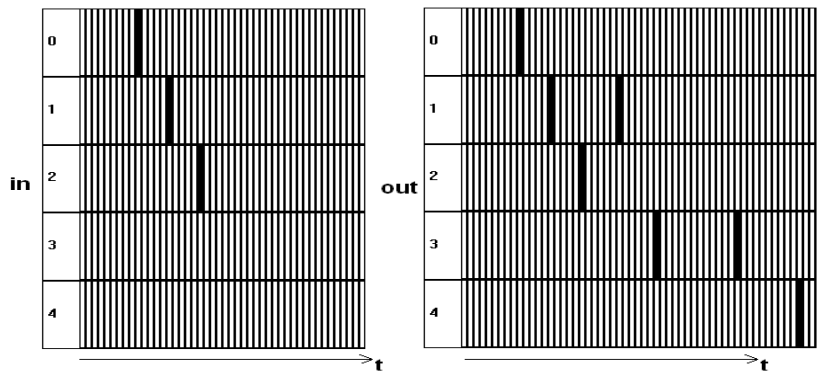

**Fig. 24.** Nosy restoration of the full pattern by the spiking memory network after 10 training iterations. The initial spikes given to the network are shown on the left. The network response pattern is shown on the right. Notice the extra spikes in the second and the fourth channel and unreliable generation of the last spike.

After $k=20$ the network is capable of restoring the full pattern from the first two spikes (Fig. 25).

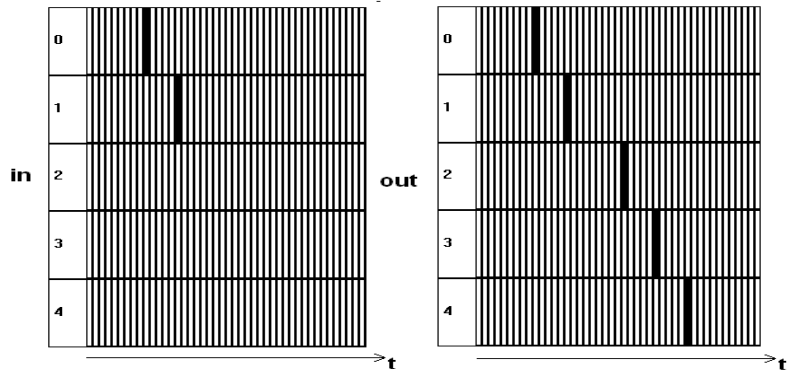

**Fig. 25.** Restoring the full pattern from the 2 spikes after 20 learning iterations. The initial spikes given to the network are shown on the left. The network response pattern is shown on the right. Notice the similarity of the response pattern with the original pattern from the Fig. 23.

Also we performed an experiment with the network trained to store two patterns, shown in Fig. 26. The network is capable of storing two patterns that differ in the spike order but not in the number of spikes. The recall is done by presenting the first two spikes of the particular pattern.

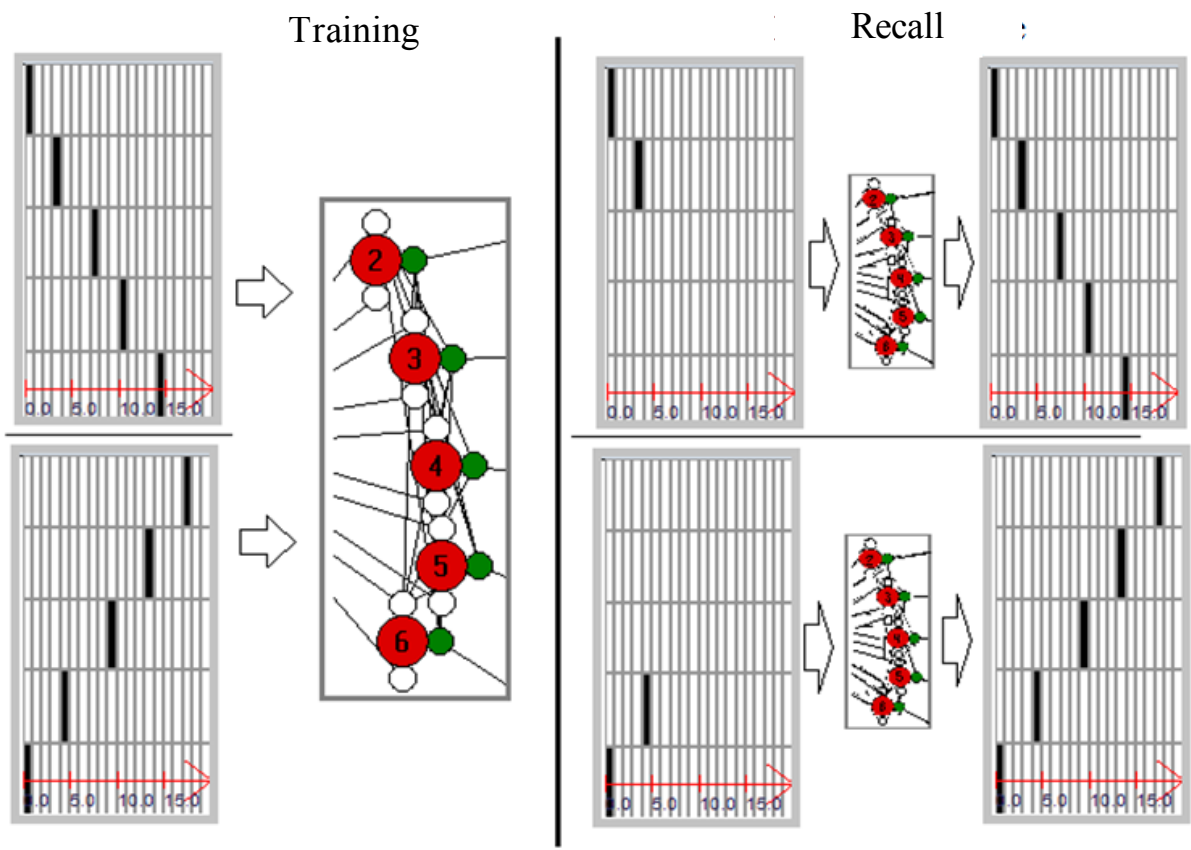

**Fig. 26.** Storing and recall of two different patterns with the different temporal structure. Two test patterns are shown to the network (left). After the training two different initial spikes (left patterns on the right part of the figure) evoke correct original patterns (right patterns on the right part of the figure). Notice the similarity of the most left (original) and the most right patterns (restoration).

In another experiment we used the spiking patterns distance developed in Chapter 1 for measuring the quality of the recall. The synfire chain test pattern for a 20-channel memory network is presented in Fig. 27. It is expected that the neurons will be activated in sequence starting from a single spike clue. Every iteration consisted of presenting the pattern, changing the weights and then testing the recall by presenting only the first spike. After that the distance between the desired and recalled pattern was computed.

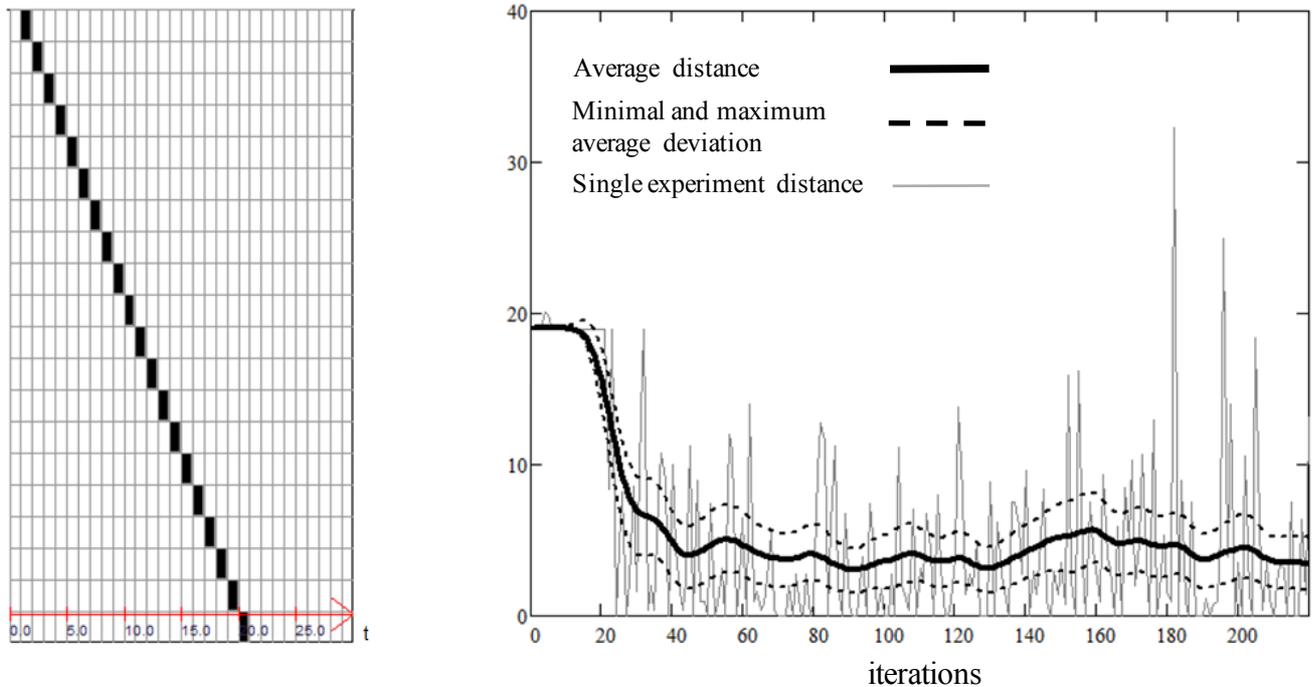

**Fig. 27.** The distance between the desired and recalled patterns. The testing pattern consisting of 20 sequential spikes in 20 input channels is shown on the left. The recall process consisted of presenting only the first spike in the sequence. The average spiking distance between the desired and recalled pattern during the training (bold line) and maximum and minimum average deviation from the average distance (dashed lines) are shown on the right. The averaging took place after ten runs of the training and recall. The grey plot is an example of distance evolution in a single experiment.

Initially the distance is equal to 19 since only the first spike out of 20 is present and the network doesn't generate any spikes by itself. During the training the distance quickly decreases. The recall process is a stochastic process since SMRM neurons are stochastic. Sometimes perfect recall with zero distance can happen only after 22 presentations (see an example shown with a grey line on Fig. 27). On average a good recall with the distance of 4 happens after about 40 training iterations.

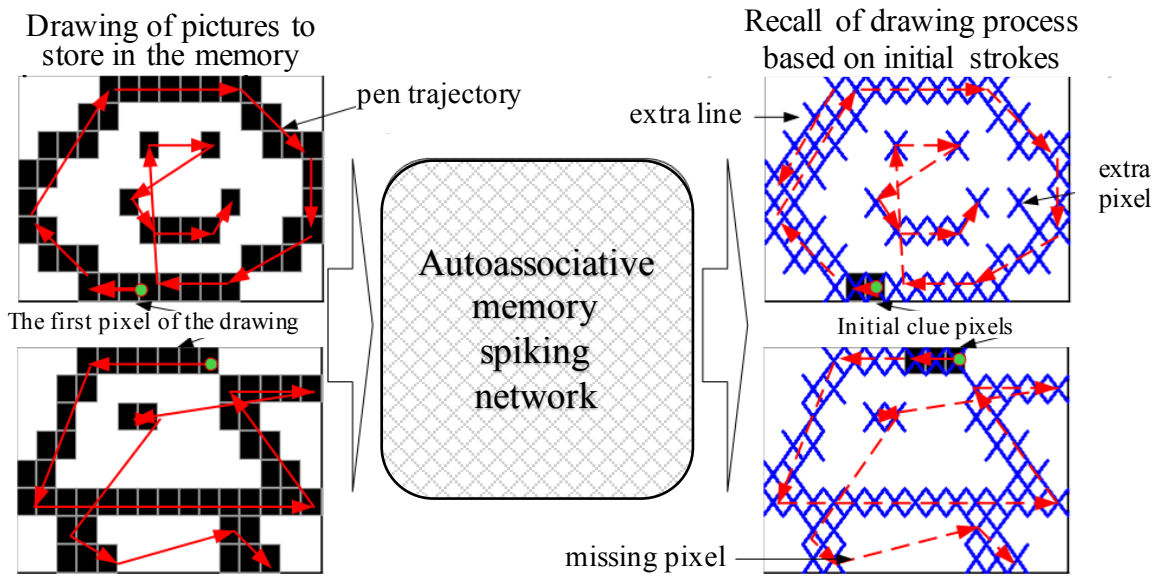

**Fig. 28.** Storing and recall of the drawing patterns. Two original human drawings are shown on the left. The red arrows show the order of the pixels in the drawings. During the training the original patterns were presented to the network several times. After the training the network was able to restore drawing process based on 2-3 initial pixels. Notice extra/missing pixels in the recalled patterns.

The autoassociative memory network has been also tested on the task of learning the drawing patterns (Fig. 28). A teacher draws a simple picture on a virtual board of $8 \times 15$ pixels initialized with zeroes. A drawing process consists of the set of events of marking the coordinates $x_i, y_i$ at times $t_i$ with the value "1". This process was represented as a spatiotemporal pattern of spikes $(n_i, t_i)$, where $n_i$ is the input channel index of the spike at $t_i$. The channel index $n_i$ for the $i$-th pixel is determined by flattening out the pixel matrix (e.g. $n_i = 15x_i + y_i$). The number of channels in the pattern is equal to the total number of pixels - 160. The process of recording of a pattern is shown in the following video: http://www.youtube.com/watch?v=O_Hz_mCNUec. The recorded spiking pattern has the information about the spatial trajectory of the "pen" and information about the speed of drawing stored in the interspike intervals. The recorded pattern was presented several times to the autoassociative memory network consisting of 160 neurons. The task of the network is to learn the process of drawing and to be able to restore the drawings based on the initial strokes of the pen. The network was trained and tested on two pictures shown in Fig. 28, left. The network was able to successfully recall the movements of the virtual pen based on the first pen stroke consisting of just 2-3 pixels (http://www.youtube.com/watch?v=IpO4TqLp3dk). The starting strokes were chosen to ease the recall process. Notice that the pictures have overlapping spatial lines. The network was not confused between them because of the different direction of drawing that were stored in the temporal relations between the spikes. If the initial clue comes from the middle of the drawing process, the networks restores the remaining pixels. For example, if you mark two pixels of the eyes from the "face" drawing, the network finishes the drawing with the mouth line (http://www.youtube.com/watch?v=ARHxRmbExr8).

A special choice of network's parameters (increased stochasticity and learning rate) makes the network capable of storing a picture based on just one presentation of the drawing process (http://www.youtube.com/watch?v=NaTEKB3fL5w). In this case the network works as a short-term memory. However, in this case the capacity of the network significantly drops, so the network cannot store multiple drawings if they have overlapping pixels. If the stored pictures share pixels, the network restores both drawings simultaneously.

## 2.6. Supervised learning of the generalized spiking neuron in continuous time

Let us consider a supervised learning task for a spiking neuron in continuous time. The neuron has to learn to generate a desired pattern $y_T^d$ in response to the input pattern $\bar{x}_T$. We assume that the interspike intervals in the desired pattern are always larger than the absolute refractoriness interval $\Delta t_{refr}$ since it is impossible to learn the pattern where this doesn't hold.

The probability of generating $y_T^d$ where each spike is generated with accuracy $\Delta t$ can be computed using (11):

$$P_{T,\Delta t}(y_T^d) = \prod_{t_j^{out} \in y_T^d} (\lambda(t_j^{out})\Delta t) \cdot e^{-\int_T \lambda(s)ds}.$$

The surprisal can be computed using:

$$h_{T,\Delta t}(y_T^d) = -\ln(P_{T,\Delta t}(y_T^d)) =$$

$$= -\sum_{t_j^{out} \in y_T^d} \ln(\lambda(t_j^{out})\Delta t) + \int_T \lambda(s)ds =$$

$$= -\sum_{t_j^{out} \in y_T^d} \ln(\lambda(t_j^{out})) + \int_T \lambda(s)ds - n\ln(\Delta t),$$

where $n$ is the number of the output spikes.

Let us introduce differential surprisal $h^{diff}(y_T^d)$ [49, 51]:

$$h^{diff}(y_T^d) = -\sum_{t_q^{out} \in y_T^d} \ln(\lambda(t_q^{out})) + \int_T \lambda(s)ds.$$

The value of the surprisal $h_{T,\Delta t}(y_T^d)$ can be determined using the $h^{diff}(y_T^d)$ up to a constant, which depends on the discretization step $\Delta t$ and number of spikes $n$ in the pattern:

$$h_{T,\Delta t}(y_T^d) = h^{diff}(y_T^d) - n\ln(\Delta t).$$

Since the constant $n\ln(\Delta t)$ does not depend on the neuron's parameters, the minimization of the differential surprisal $h^{diff}(y_T^d)$ is equivalent to the minimization of $h_{T,\Delta t}(y_T^d)$. We are going to minimize the differential surprisal using the stochastic gradient descent. Let's find a partial derivative of the differential surprisal with respect to a parameter $w$:

$$\frac{\partial h_{T,\Delta t}(y_T^d)}{\partial w} = \frac{\partial h^{diff}(y_T^d)}{\partial w} = -\sum_{t_l^d \in y_T^d} \frac{1}{\lambda(t_l^d)} \frac{\partial \lambda(t_l^d)}{\partial w} + \int_T \frac{\partial \lambda(s)}{\partial w} ds.$$

The derivative of the intensity $\lambda(t)$ with respect to the parameter is equal to:

$$\frac{\partial \lambda(t)}{\partial w} = \frac{\partial \lambda(q)}{\partial q} \frac{\partial q(t)}{\partial w},$$

where $q(t)$ is the state of the generalized neuron model at time $t$.

Using stochastic gradient descent, the change in parameters is equal to:

$$\Delta w = -\gamma \cdot \frac{\partial h(y_T^d)}{\partial w},$$

where $0 < \gamma < 1$ is the learning coefficient. The learning rules take the following form:

$$\Delta w^T = \gamma \cdot \left( \sum_{t_l^d \in y_T^d} \frac{1}{\lambda(t_l^d)} \frac{\partial \lambda(q(t_l^d))}{\partial q} \frac{\partial q(t)}{\partial w} - \int_T \frac{\partial \lambda(q(s))}{\partial q} \frac{\partial q(s)}{\partial w} ds \right) \quad (35)$$

This learning rule is valid for any stochastic neuron model if $\frac{\partial q(t)}{\partial w}$ exists and the intensity $\lambda(t)$ is not equal to zero at the desired spike times.

The derivative of the SMRM neuron's state with respect to its weights is equal to:

$$\frac{\partial q(t)}{\partial w_{ik}} = \frac{\partial u(t)}{\partial w_{ik}} = \sum_{t_j^i \in x_T^i} \frac{\partial}{\partial w_{ik}} \sum_i^n \sum_k^m w_{ik} \alpha_k(t - t_j^i) = \sum_{t_j^i \in x_T^i} \alpha_k(t - t_j^i)$$

By substituting this into the learning rule (35), we get the supervised learning rules for the SMRM neuron in continuous time:

$$\Delta w_{ik}^T = \gamma \cdot \left( \sum_{t_l^d \in y_T^d} \sum_{t_j^i \in x_T^i} \frac{1}{\lambda(t_l^d)} \frac{\partial \lambda(u(t_l^d))}{\partial u} \alpha_k(t_l^d - t_j^i) - \sum_{t_j^i \in x_T^i} \int_T \frac{\partial \lambda(u(s))}{\partial u} \alpha_k(s - t_j^i) ds \right) \quad (36)$$

Let's check that the obtained equation is equivalent to the discrete time equation in the limit $\Delta t \to 0$:

$$\Delta w_{ik}^T = \gamma \cdot \left( \sum_{t_l^d \in y_T^d} \frac{1 - \Lambda(t_l^d)}{\Lambda(t_l^d)} \frac{\partial \lambda(u(t_l^d))}{\partial u} \sum_{t_j^i \in x_T^i} \alpha_k(t_l^d - t_j^i) \Delta t - \sum_{t_l \notin y_T^d} \frac{\partial \lambda(u(t_l))}{\partial u} \sum_{t_j^i \in x_T^i} \alpha_k(t_l - t_j^i) \Delta t \right) =$$

$$\stackrel{\Delta t \to 0}{=} \gamma \cdot \left( \sum_{t_l^d \in y_T^d} \frac{1 - \lambda(t_l^d) \Delta t}{\lambda(t_l^d) \Delta t} \frac{\partial \lambda(u(t_l^d))}{\partial u} \sum_{t_j^i \in x_T^i} \alpha_k(t_l^d - t_j^i) \Delta t - \int_T \frac{\partial \lambda(u(s))}{\partial u} \sum_{t_j^i \in x_T^i} \alpha_k(s - t_j^i) ds \right) =$$

$$= \gamma \cdot \left( \sum_{t_l^d \in y_T^d} \frac{1 - \bar{o}(\Delta t)}{\lambda(t_l^d)} \frac{\partial \lambda(u(t_l^d))}{\partial u} \sum_{t_j^i \in x_T^i} \alpha_k(t_l^d - t_j^i) - \int_T \frac{\partial \lambda(u(s))}{\partial u} \sum_{t_j^i \in x_T^i} \alpha_k(s - t_j^i) ds \right) =$$

$$= \gamma \cdot \left( \sum_{t_l^d \in y_T^d} \sum_{t_j^i \in x_T^i} \frac{1}{\lambda(t_l^d)} \frac{\partial \lambda(u(t_l^d))}{\partial u} \alpha_k(t_l^d - t_j^i) - \sum_{t_j^i \in x_T^i} \int_T \frac{\partial \lambda(u(s))}{\partial u} \alpha_k(s - t_j^i) ds \right).$$

We can simplify it further if the intensity $\lambda(u)$ is the exponent (28). In this case using $\frac{\partial \lambda(u)}{\partial u} = \frac{1}{\kappa} \lambda(u)$, we get:

$$\Delta w_{ik}^T = \frac{\gamma}{\kappa} \cdot \left( \sum_{t_l^d \in y_T^d} \sum_{t_j^i \in x_T^i} \alpha_k(t_l^d - t_j^i) - \sum_{t_j^i \in x_T^i} \int_T \lambda(u(s)) \alpha_k(s - t_j^i) ds \right). \qquad (37)$$

## 2.7. Relations between the change of weight and pre-/post-synaptic spike times

Consider the simplest training task that can be solved using the rule (37). Let the neuron be in a rest state with the membrane potential equal to zero and with no history of input spikes. At time $t^{in}$ the neuron receives an input spike. Also here let's get back to the neuron model with only one alpha function per input channel (SRM0). In this case the membrane potential changes according to $u(t) = w \cdot \alpha(t - t^{in})$. The neuron receives a teaching pattern $y_{[-T,T]}^d$ that has only a single teaching spike at $t^{out} = 0$. The single weight of the neuron changes according to (37):

$$\Delta w(t^{in}) = \frac{\gamma}{\kappa} \cdot \left( \alpha(t^{out} - t^{in}) - \int_{-T}^{T} \lambda(u(s)) \cdot \alpha(s - t^{in}) ds \right) \qquad (38)$$

By changing the time of the input spike $t^{in}$ and fixing the teaching spike $t^{out} = 0$ and the weight value $w$, we can plot $\Delta w$ depending on the time difference between the input and output spikes $\Delta t = t^{out} - t^{in}$ (Fig. 29 (left)).

It can be seen from the Fig. 29 (left) that the weight change is qualitatively different depending on the sign of $\Delta t$. When the time of the input spike $t^{in}$ is smaller than the time of the output spike $t^{out}$, the weight should be increased in order to increase the probability of a spike. The magnitude of the increase is roughly proportional to the alpha function value at time $t^{out}$ (this can be seen from the plot when $\Delta t$ is positive). When the $\Delta t$ is negative the increase of weight value will lead to the generation of an unwanted extra spike $t^{fail} > t^{in}$, which will decrease the probability of generating only a single spike at $t^{out}$. Therefore, the change of the weight is negative when $\Delta t < 0$. The weight's change in this case almost does not depend on the $t^{out}$ and is coming from the second term of equation (38). The shape and relations of the positive and negative parts of the plot depend on the current value of the weight $w$ and the parameters of the stochastic threshold. The increase in stochasticity or weight value leads to a larger magnitude of the negative weight change $\Delta w(\Delta t)$ when $\Delta t < 0$ because of the larger value of the second term in equation (38).

Neurophysiological data shows that the changes in the amount of neurotransmitter that is released in the synapse during the learning depends on the difference between the input and output spikes. This dependency is called Spike Timing Dependent Plasticity (STDP) [23, 65, 66]. It is interesting to compare theoretically derived dependency $\Delta w(\Delta t)$ with the experimentally obtained STDP curve (Fig. 29, right). One can notice their similarity when the output spike is generated after the input, $\Delta t > 0$. However, plots differ when $\Delta t < 0$ and the output spike is generated before the input. The data shows the dependency from the $\Delta t$ but the theoretical curve does not have this dependency except the short region due to the refractory events.

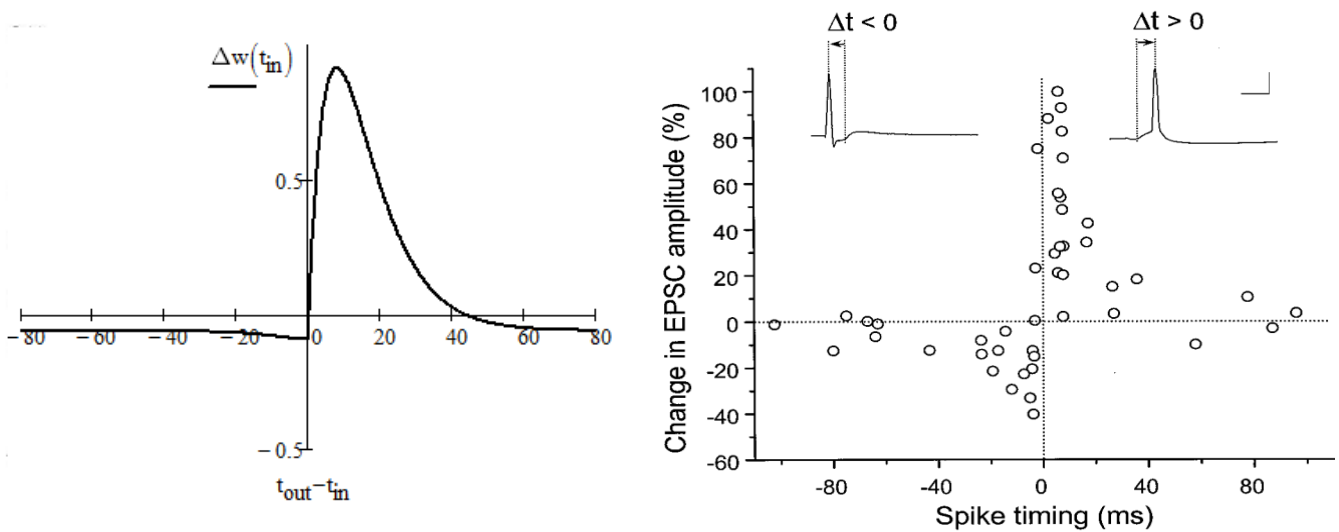

**Fig. 29.** Theoretical (left) and experimental (STDP) (right) dependence between the weight changes and the interval between the input and output spikes. The left curve is build by integrating the supervised learning surprisal minimization equation. The right figure is taken from [23] with permission. Notice the similarity between the built curve and the shape of the data points cloud.

## 2.8. Convergence of the supervised learning algorithm

If the gradient descent is used to minimize the surprisal $h(y_T)$ it becomes important whether the surprisal is a convex function of parameters $\bar{W}$. The surprisal $h(y_T)$ is bounded from below, so if $h(y_T, \bar{W})$ is a convex function of $\bar{W}$, then it's local minimum is necessarily global. If the learning coefficient $\gamma$ is small enough then the parameters $\bar{W}$ will converge to the point $\bar{W}^*$ where the surprisal is in its global minimum $h(y_T, \bar{W}^*) = h^*(y_T)$. Below we are going to show when the surprisal is a convex function of the weights $\bar{W}$ of the SMRM neuron.

The surprisal function is convex if and only if its Hessian matrix is positive semidefinite. Let's find a hessian matrix of $h(y_T)$ with respect to the weights $w_{ij}$ of the SMRM neuron:

$$\frac{\partial^2 h(y_T)}{\partial w_{ij} \partial w_{lm}} = \int_T \frac{\partial^2 \lambda(u(s))}{\partial u^2} A_{ijlm}(s) ds - \sum_{t_q^{s\_out} \in y_T} \frac{\partial^2 \ln(\lambda(u(t^d)))}{\partial u^2} \cdot A_{ijlm}(t_q^d), \quad (39)$$

where $A_{ijlm}(t) = \sum_{t_{k1}^i \in x_T^i} \alpha_j(t - t_{k1}^i) \sum_{t_{k2}^l \in x_T^l} \alpha_m(t - t_{k2}^l)$; $ij$ and $lm$ run through $n \cdot m$ weights in the flattened matrix $\bar{W}$. The matrix $A_{ijlm}(t)$ of $n \cdot m \times n \cdot m$ dimensions is positive semidefinite for every $t$, because it is the Gramian matrix for alpha functions values at time $t$.

The second order derivative of the exponential stochastic threshold is strictly positive:

$$\frac{\partial^2 \lambda(u(s))}{\partial u^2} = (1/\kappa^2) e^{(u(s) - Th)/\kappa}$$

The second order derivative of the log of the intensity function is zero:

$$\frac{\partial^2 \ln(\lambda(u(t)))}{\partial u^2} = \frac{\partial^2 (u(t) - Th)/\kappa}{\partial u^2} = 0$$

If the stochastic threshold is exponential then the Hessian (39) has only the first term which is a linear combination of positive semidefinite matrices with positive coefficients. Therefore, the Hessian of the surprisal is positive semidefinite and the surprisal $h(y_T)$ is a convex function of the weights of the SMRM neuron.

## 2.9. Choosing alpha-function parameters

The properties of the supervised learning process depend on the number of alpha-functions and their parameters. The set of alpha functions is a vector basis of the representation of the postsynaptic potential $PSP_i(t) = \sum_{t_k^i \in x_i^T} \sum_j^m w_{ij} \alpha_j(t - t_k^i)$. A perfect training result is obtained when the intensity function $\lambda(t)$ peaks during the desired spike and is minimal at all other times. With the exponential stochastic threshold the membrane potential has to go to minus infinity at all points on the interval $T$ except the desired spike times $t_k^d$, where the the membrane potential has to be really large in order to increase $\lambda(t)$ enough for a spike. Let's choose an interval of the potential values $U_{\mathit{eff}} = [u_{\min}, u_{\max}]$ such that if $u < u_{\min}$ the spike probability is close to zero and if $u > u_{\max}$ then the spike probability is high enough. The "perfect" membrane shape with such approximations can be described with:

$$\begin{cases} u(t) \geq u_{\max}, & \text{if } t \in y^T \\ u(t) \leq u_{\min}, & \text{otherwise} \end{cases} \qquad (40)$$

where $y^T$ is the desired output pattern.

Consider a spiking pattern detection task. The neuron receives an input pattern $\bar{x}_T^*$ starting at time 0, the pattern's last spike comes at $t_{\mathit{last}}^{\mathit{in}} < T$. The neuron should generate a single output spike after receiving this pattern at time $t^*$ such that $t_{\mathit{last}}^{\mathit{in}} < t^* \leq T$ and it should not generate spikes before and after the desired time $t^*$. If the input pattern has as single spike per input channel, then the constraints on the membrane potential (40) will take the following form:

$$\begin{cases} \sum_{i}^{n} \sum_{j}^{m} w_{ij} \alpha_j (t^* - t_i^{\mathit{in}}) \geq u_{\max} \\ \sum_{i}^{n} \sum_{j}^{m} w_{ij} \alpha_j (t - t_i^{\mathit{in}}) \leq u_{\min}, \; npu \; t \in [0, T], t \neq t^* \end{cases} \qquad (41)$$

If at least one spike from the input pattern is not present, the neuron should not generate the spike at $t^*$ and it's potential $u(t^*)$ should be lower than $u_{\max}$. The condition of equal contribution of all input spikes is expressed with the relations $\forall i \; PSP_i(t^*) = u_{\max}/n = PSP_{\max}$ and $\forall i \; PSP_i(t) \leq u_{\min}/n = PSP_{\min}$ when $t \neq t^*$. Notice that $PSP_i(t) = 0$ for $t \leq t_i^{\mathit{in}}$. Now we can express constraints (41) on each postsynaptic potential as follows:

$$\begin{cases} \sum_{j}^{m} w_{ij} \alpha_j (t^* - t_i^{\mathit{in}}) = PSP_{\max} \\ \sum_{j}^{m} w_{ij} \alpha_j (t - t_i^{\mathit{in}}) \leq PSP_{\min}, \; \forall t \in [0, T], t \neq t^* \end{cases} \qquad (42)$$

Equations (42) require that the postsynaptic potential has a large discontinuous jump at $t^*$ that is not possible because alpha-functions are continuous. Let's choose a finite value of time accuracy of the output spike generation $\Delta_s$ and relax (42): postsynaptic potential $PSP_i(t)$ has to be lower $u_{min}$ when $t \in [t_i^{in}, t^* - \Delta_s] \cup [t^* + \Delta_s, T]$. Therefore the membrane potential has to grow from the value below $u_{min}$ to the value $u_{max}$ during $\Delta_s$ around the desired spike and then go down again below $u_{min}$. If we split the interval $[t_i^{in}, T]$ into the small intervals where the membrane potential is roughly constant then the conditions (42) will turn into the system consisting of one equation and the set of linear inequalities. Let's extract $w_{i1}$ from the first equation (42) and substitute it into the inequalities. As a result we will get a system of inequalities with $m-1$ unknowns (weights $w_{i2}...w_{im}$):

$$\sum_k \sum_{j=2}^m w_{ij} \left( \alpha_j(t_k - t_i^{in}) - \frac{\alpha_j(T - t_i^{in})}{\alpha_1(T - t_i^{in})} \right) \leq u_{min} - \frac{PSP_{max}}{\alpha_1(T - t_i^{in})}, \forall t_k \in [t_i^{in}, T], t_k \neq t^*$$

The solution of these equations is a set of possible weight values at the $i$-th input channel. The neuron would be a perfect detector of the input pattern if the weights belong to this solution set.

For a neuron with two alpha functions the solution is:

$$\begin{cases} w_{i2} \geq \dfrac{PSP_{max} \alpha_1(t - t_i^{in}) - PSP_{min} \alpha_1(t^* - t_i^{in})}{\alpha_2(t^* - t_i^{in})\alpha_1(t - t_i^{in}) - \alpha_2(t - t_i^{in})\alpha_1(t^* - t_i^{in})}, \forall t \in [0, T], t \neq t^* \\ w_{i1} = \dfrac{PSP_{max} - w_{i2}\alpha_2(t^* - t_i^{in})}{\alpha_1(t^* - t_i^{in})} \end{cases} \quad (43)$$

The inequalities (43) can be solved in a narrow space of other parameters $\{PSP_{min}, PSP_{max}, t^*, t_i^{in}, \{\tau_{\alpha 1}^{max}, \tau_{\alpha 2}^{max}\}\}$, where $\{\tau_{\alpha 1}^{max}, \tau_{\alpha 2}^{max}\}$ are maximums of alpha functions. For example, the system cannot be solved when: $\{PSP_{min} = 0.1, PSP_{max} = 0.2, t^* = 6, \{\tau_{\alpha 1}^{max} = 2, \tau_{\alpha 2}^{max} = 6\}\}$. This is the case because when $t^* > \tau_{\alpha 1}^{max}$, the weight of the second alpha-function is positive. In order to compensate the increase of the second alpha function before the time $t^*$ the weight of the first alpha function has to be negative. However, it is not enough to use the first alpha function to decrease the potential after $t^*$ because it dies out quicker than the second one. The minimal number of alpha functions to get the desired behavior of the postsynaptic potential is three. Since the larger number of alpha functions is computationally expensive, we use three alpha functions in all experiments in this thesis. Below we are going to try to choose the best parameters for the three alpha functions.

The solution region of the inequalities for three alpha functions with parameters $\{PSP_{min} = 0.1, PSP_{max} = 0.2, t^* = 6, \{\tau_{\alpha 1}^{max} = 2, \tau_{\alpha 2}^{max} = 6, \tau_{\alpha 3}^{max} = 15\}\}$ is shown in Fig. 30.

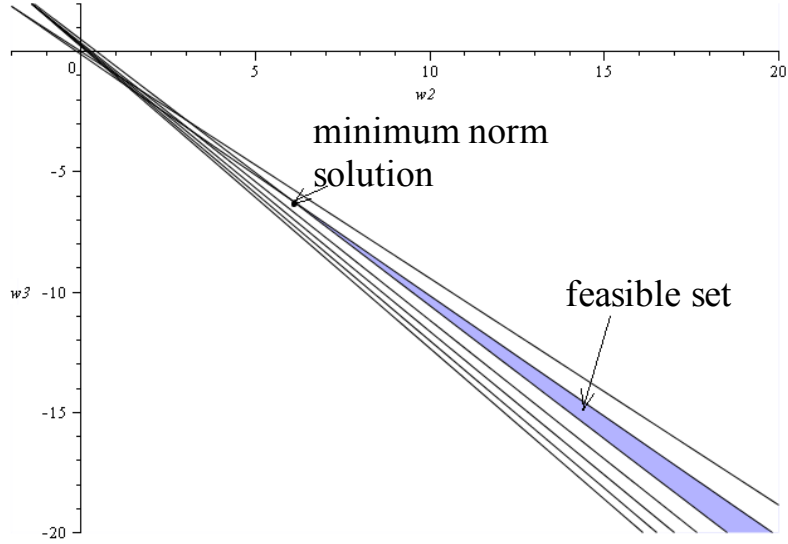

**Fig. 30.** The solution set of the system of inequalities for three alpha functions (see text).

As it can be seen from Fig. 30, there is a point in the weights solution set that is the closest to the origin. To find it we introduce a helper function:

$$D_{pq}(t_1, t_2) = \alpha_p(t_1)\alpha_q(t_2) - \alpha_p(t_2)\alpha_q(t_1)$$

Let us use the following notation:

$$D_{pq}(\tau) = D_{pq}(\tau + \Delta_s, \tau - \Delta_s);$$
$$G_{pq}(\tau) = D_{pq}(\tau, \tau - \Delta_s) + D_{pq}(\tau + \Delta_s, \tau)$$

where $\Delta_s$ is the spike time precision interval, $\Delta\tau$ is the interval between the input and output spike. Also let's denote $\tau_i = t^* - t_i^{in}$. One can find an exact solution for the inequalities (42) given that the norm of the weights is minimized:

$$\begin{cases} w_{i1}^* = \dfrac{PSP_{\min}G_{32}(\tau_i) - PSP_{\max}D_{32}(\tau_i)}{\alpha_3(\tau_i)D_{12}(\tau_i) + \alpha_2(\tau_i)D_{31}(\tau_i) - \alpha_1(\tau_i)D_{32}(\tau_i)} \\ w_{i2}^* = \dfrac{PSP_{\min}G_{13}(\tau_i) - PSP_{\max}D_{13}(\tau_i)}{\alpha_1(\tau_i)D_{23}(\tau_i) + \alpha_3(\tau_i)D_{12}(\tau_i) - \alpha_2(\tau_i)D_{13}(\tau_i)} \\ w_{i3}^* = \dfrac{PSP_{\min}G_{21}(\tau_i) - PSP_{\max}D_{21}(\tau_i)}{\alpha_2(\tau_i)D_{31}(\tau_i) + \alpha_1(\tau_i)D_{23}(\tau_i) - \alpha_3(\tau_i)D_{21}(\tau_i)} \end{cases} \quad (44)$$

An example of the postsynaptic potential obtained from solving (44) given the parameters:

$$\{PSP_{\min} = 0.1, PSP_{\max} = 0.2, \tau_i = 6, \Delta_s = 1, \{\tau_{\alpha 1}^{\max} = 2, \tau_{\alpha 2}^{\max} = 6, \tau_{\alpha 3}^{\max} = 15\}\},$$

is shown in Fig. 31. It can be seen that the plot satisfies the constraints: maximum of 0.2 is reached at *6ms* and at all other times that are further than 1ms from the maximum the potential is lower than 0.1.

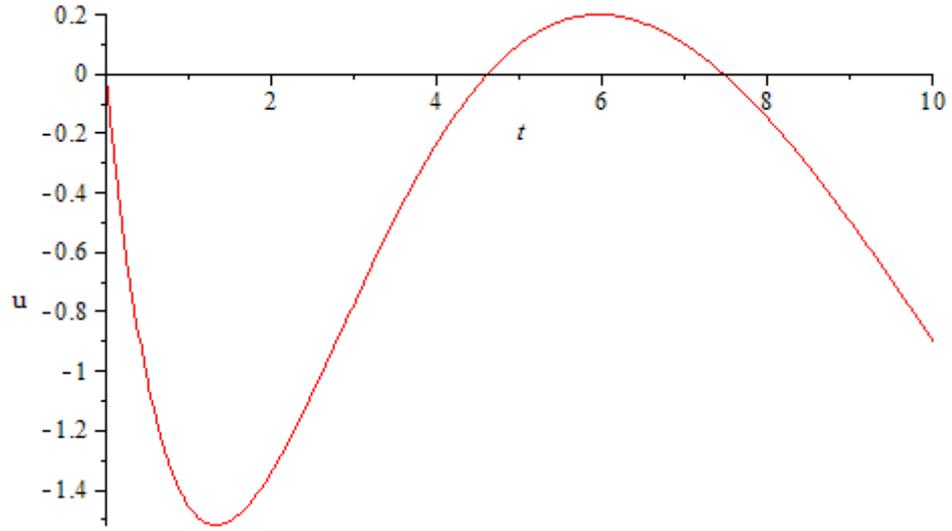

**Fig. 31.** The weighted sum of three alpha functions with the weights obtained by solving the equation (44) given the 6ms interval between the input and desired output spike.

We assume that the minimal norm solution is preferable because it is faster to reach during the training and it is a common regularization technique for the supervised learning tasks usually achieved by the weights decay. Let's denote the weights norm: $l_i(\tau_i) = \sqrt{w_{i1}^{*2} + w_{i2}^{*2} + w_{i3}^{*2}}$, where the weights are obtained by solving (44). Assume that the neuron is going to be trained on patterns with duration no longer than $t_{max}$. Let's define a cost function $L = \frac{1}{t_{max}} \int_{\Delta s}^{t_{max}} l_i(\tau) d\tau$ that is the average weights norm for detecting all patterns no longer than the maximum duration. We minimized this cost with respect to the alpha function parameters using the off-the-shelve numeric optimization package. We obtained the following values for the alpha function maximums that leads to the best solutions of the supervised learning with respect to the defined cost: $\{\tau_{\alpha 1}^{max} = 1.8, \tau_{\alpha 2}^{max} = 3.3, \tau_{\alpha 3}^{max} = 9.3\}$. These parameters were used for the experiments throughout the thesis.

# Chapter 3. Unsupervised learning of spiking neurons

**Summary:**

In this chapter we formulate an unsupervised learning task to increase the robustness of the spike generation process. We take the entropy of the neuron's output as a robustness measure. If the entropy is high, then a lot of output patterns are equally likely and a particular output is not generated robustly. If the entropy is low then the neuron will generate a particular output based on the input more reliably. We develop an original unsupervised learning algorithm that allows us to approximately solve the entropy minimization task during online learning with the assumption that the output firing rate is small.

We performed experiments of entropy minimization with the SMRM neuron and plotted the entropy evolution during training. The results show that after training the likely output patterns are generated robustly while other less likely patterns are not generated anymore. We developed a composite training algorithm for speeding up the supervised learning. This algorithm interleaves supervised surprisal minimization and unsupervised entropy minimization. We show that the developed algorithm allows the neuron to learn the delay between the input and output spike.

## 3.1. Unsupervised minimization of entropy of the neuron's output

Here we describe an unsupervised learning task in which entropy $H$ of a neuron's output decreases. The entropy of the neuron's output $H_T$ on an interval $T$ is defined as the expectation of the surprisals of all possible output patterns:

$$H_T = \sum_{y_T \in S_T} P\{y_T | \bar{x}_T, q_0\} \cdot h(y_T | \bar{x}_T, q_0) \tag{45}$$

The entropy $H_T = H_T(\bar{x}_T, q_0)$ characterizes a variability of the neuron's behavior given the input pattern $\bar{x}_T$ and initial state $q_0$ [50]. Below we are going to drop the conditional notation assuming in all derivations that $\bar{x}_T$ and $q_0$ are fixed.

We will develop a learning algorithm only for a discrete time simulation with the discretization interval $\Delta t$. Let's denote a membrane potential value on a particular interval as $u_{\Delta t}$. The probability of an output spike on this interval is denoted as $\Lambda_{\Delta t}$, which can be computed using (14). There are two possible outputs on a single interval: a spike or silence. Therefore, the entropy on the interval $\Delta t$ equals to $H_{\Delta t} = -\Lambda_{\Delta t} \cdot \ln(\Lambda_{\Delta t}) - (1 - \Lambda_{\Delta t}) \cdot \ln(1 - \Lambda_{\Delta t})$. The value of $H_{\Delta t}$ is maximal when $\Lambda_{\Delta t} = 0.5$ (a spike or silence are equally likely on $\Delta t$). The value of $H_{\Delta t}$ is minimal and is equal to zero when $\Lambda_{\Delta t} = 1$ or $\Lambda_{\Delta t} = 0$. In this case the neuron behaves deterministically – it either always generates the spike ($\Lambda_{\Delta t} = 1$) or is always silent ($\Lambda_{\Delta t} = 0$). Provided that $\Lambda_{\Delta t}$ depends monotonically on the potential $u_{\Delta t}$, the entropy $H_{\Delta t}$ is close to zero if the potential is very small or very high (Fig. 32). The entropy is maximal when the potential is near the threshold.

For a fixed input pattern, the membrane potential $u_{\Delta t}$ depends on the weights $\bar{W}$. If the weights are high, we need to increase the weights more in order to decrease the entropy, and if the weights are small, we need to decrease them.

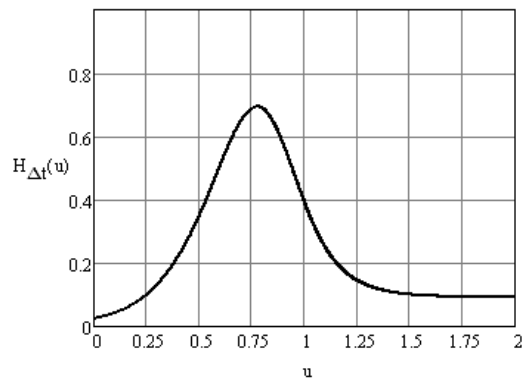

**Fig. 32.** The entropy of the neuron's output on a single time step $\Delta t$ depending on the neuron's membrane potential. Notice that the entropy is maximal when the membrane potential is close to the threshold "1". The entropy is minimal if the potential is small (high probability of silence) or very large (high probability of a spike).

If the interval $T$ consists of $n$ time steps $\Delta t$, the number of possible output patterns is $2^n$. In order to compute the entropy one needs to compute $2^n$ probabilities of every output pattern using (4), and then use the equation (45).

In order to minimize the entropy $H_T$ we can use the stochastic gradient descent and compute a single change of weight $\Delta w_{ij}$ as:

$$\Delta w_{ij} = -\gamma \cdot \frac{\partial H_T}{\partial w_{ij}}$$

where $0 < \gamma < 1$ is the learning coefficient. Using the equality $\frac{\partial P\{y_T\}}{\partial w_{ij}} \equiv P\{y_T\} \cdot \frac{\partial h(y_T)}{\partial w_{ij}}$, we get the following equation for the weight's change:

$$\Delta w_{ij} = -\gamma \cdot \sum_{y_T \in S_T}^{2^n} \frac{\partial h(y_T)}{\partial w_{ij}} \cdot (1 + h(y_T)) \cdot P\{y_T\} \qquad (46)$$

This approach to the entropy minimization has been applied in [50] where it has been shown that the entropy minimization can be related to the STDP in real neurons. However, computational complexity of the equation (46) grows exponentially with the growth of the interval $T$. This limits its use in practical applications. Also this equation is especially hard to use during online learning since the evaluation of probabilities of all possible output patterns on the interval $T$ can be done only at the end of the interval when the whole input pattern is known. This makes a computational load quite uneven. Also it is not always obvious how to choose the interval $T$ in the first place.

Let us try to find a learning algorithm for the entropy minimization that is practical to use during the online learning. First we are going to reformulate the entropy evaluation technique and then we will apply some reasonable approximations to make the rules scale linearly with the length of the interval $T$. Consider the output patterns space $S_T$ on the interval $T$. We can compute the probability of every element $y_T$ in this space using (4) or (11). Next let's introduce a new space $\hat{S}_T$ as follows. We include in this space the pattern without spikes $y_T^0$ and all the subsets of the original space $\hat{y}_T(t) \subset S_T$ which have a spike at time $t$ and an arbitrary number of spikes after $t$: $\hat{y}_T(t) = (y_T | t \in T \cup y_T = \{t_1 = t, t_2 > t_1, ...\})$. Notice again that the elements $\hat{y}_T(t)$ are not individual patterns but sets of patterns. To compute the probability of such elements we need to ignore the probabilities of events happening after *t*. In discrete time the probability of $\hat{y}_T(t)$, where *t* belongs to the *k*-th time step is equal to:

$$P\{\hat{y}_T(t)\} = \Lambda_k \prod_i^{k-1}(1-\Lambda_i)$$

where $\Lambda_k \equiv \Lambda(t_k)$ is the probability of spike on *k*-th time step. Notice that multiplication is performed up to index *k-1* and not till the end of the interval *T*.

The probability of spike $P\{\hat{y}_T(t)\}$ is equivalent to the probability of spike at *t* with silence before *t* so we are going to denote it simply $P\{t\}$. Let's also assume that interval *T* starts at $t = 0$: $T \equiv [0,T]$.

The probability of an empty pattern $y_T^0$ is equal to $P\{y_T^0\} = \prod_i^T(1-\Lambda_i)$.

The entropy of the described simplified pattern space $\hat{S}_T$ can be computed as follows:

$$\hat{H}_{[0,T]} = -P\{\hat{y}_T^0\} \cdot \ln(P\{\hat{y}_T^0\}) - \sum_{t \in [0,T]} P\{t\} \cdot \ln(P\{t\}) \tag{47}$$

Substituting the probabilities, we get:

$$\hat{H}_{[0,T]} = \prod_i^T(\Lambda_i - 1) \cdot \sum_i^T \lambda_i \cdot \Delta t +$$

$$+ \sum_k \Lambda_k \prod_i^{k-1}(\Lambda_i - 1) \cdot (\ln(\Lambda_k) + \sum_i^{k-1} \lambda_i \cdot \Delta t)$$

where $\lambda_i \equiv \lambda(t_i)$ is the point process intensity function on the *i*-th step.

Next we are going to show how minimization of the entropy $\hat{H}_{[0,T]}$ of the restricted output space $\hat{S}_T$ is related to minimization of the full entropy $H_{[0,T]}$ of the space $S_T$. Notice that the following relation holds: $\hat{H}_{[0,T]} \leq H_{[0,T]}$. Indeed, during the compressing of several individual patterns $y_T$ into the subsets $\hat{y}_T(t)$ we loose the information $\Delta\hat{H}_{[0,T]}$ about the differences between $y_T$ in the subsets:

$$H_{[0,T]} = \hat{H}_{[0,T]} + \Delta\hat{H}_{[0,T]}$$

Using the property of hierarchical additivity of entropy [67], we notice that the value $\Delta\hat{H}_{[0,T]}$ can be expressed as a weighted sum of entropies $H_{[t,T]}(|t)$:

$$\Delta\hat{H}_{[0,T]} = \sum_{t \in [0,T]} P\{t\} \cdot H_{[t,T]}(|t) \tag{48}$$

where $H_{[t,T]}(|t)$ is the entropy of the interval $[t, T]$ given that the neuron generated the spike at time $t$, $P\{t\}$ is the probability of the spike at time $t$ and silence before that.

Therefore, for the full entropy the following holds:

$$H_{[0,T]} = \hat{H}_{[0,T]} + \sum_{t \in [0,T]} P\{t\} \cdot H_{[t,T]}(|t) \tag{49}$$

The entropy $H_{[t,T]}(|t)$ by its structure is completely analogous to the full entropy $H_{[0,T]}$ except the length of the interval. Therefore, we can use the same arguments for splitting it onto the simplified subsets of patterns $\hat{S}_{[t,T]}$:

$$H_{[t,T]}(|t) = \hat{H}_{[t,T]}(|t) + \Delta \hat{H}_{[t,T]}(|t)$$

The full entropy $H_{[0,T]}$ on the whole interval $T$ can be computed using the following recursive relation:

$$H_{[0,T]} = \hat{H}_{[0,T]} + \sum_{t_1 \in [0,T]} P\{t_1\} \cdot (\hat{H}_{[t_1,T]}(|t_1) + \sum_{t_2 \in [t_1,T]} P\{t_2 | t_1\} \cdot (\hat{H}_{[t_2,T]}(|t_2) + ...))$$

In order to minimize $H_{[0,T]}$, we need to minimize both terms in (49). Let's find the derivative of $H_{[0,T]}$ with respect to the neuron's parameters:

$$\frac{\partial}{\partial w_{ij}} H_{[0,T]} = \frac{\partial}{\partial w_{ij}} \hat{H}_{[0,T]} + \frac{\partial}{\partial w_{ij}} \Delta \hat{H}_{[0,T]} \tag{50}$$

The first term of (50) equals to:

$$\frac{\partial \hat{H}_{[0,T]}}{\partial w_{ij}} = -\frac{\partial \ln(P\{y_T^0\})}{\partial w_{ij}} \cdot P\{y_T^0\} \cdot (1 + \ln(P\{y_T^0\})) -$$
$$- \sum_{t \in [0,T]} \frac{\partial \ln(P\{t\})}{\partial w_{ij}} \cdot P\{t\} \cdot (1 + \ln(P\{t\})) \tag{51}$$

The negative log-probabilities $-\ln(P\{t\})$ are the surprisals $h(\hat{y}_T(t))$ of $\hat{y}_T(t)$ and can be computed using (34). Here is an explicit derivative of $h(\hat{y}_T(t))$ for the SMRM model:

$$\frac{\partial h(\hat{y}_T(t_k))}{\partial w_{ij}} = -\sum_q^{k-1} \frac{\partial \ln((1-\Lambda_q))}{\partial w_{ij}} - \frac{\partial \ln(\Lambda_k)}{\partial w_{ij}} =$$
$$= \sum_q^{k-1} \frac{\partial \lambda_k \cdot \Delta t}{\partial w_{ij}} - \frac{1}{\Lambda_k} \frac{\partial (1-e^{-\lambda_k \cdot \Delta t})}{\partial w_{ij}} = \left( \sum_q^{k-1} \frac{\partial \lambda_q}{\partial w_{ij}} - \frac{1-\Lambda_k}{\Lambda_k} \frac{\partial \lambda_k}{\partial w_{ij}} \right) \Delta t = \tag{52}$$
$$= \left( \sum_q^{k-1} \frac{\partial \lambda_q}{\partial u_q} \sum_{t_l^i \in x_T^i} \alpha_j(t_q - t_l^i) - \frac{1-\Lambda_k}{\Lambda_k} \frac{\partial \lambda_k}{\partial u_k} \sum_{t_l^i \in x_T^i} \alpha_j(t_k - t_l^i) \right) \Delta t$$

where $\frac{\partial \lambda_k}{\partial u_k}$ is a derivative of the intensity function on the $k$-th step, $t_l^i$ is the time of the input spike on the $i$-th input channel, $\alpha_j(t)$ is the alpha-function, $\Delta t$ is the time discretization step.

The derivative of the surprisal of an empty pattern can be computed using:

$$\frac{\partial h(y_T^0)}{\partial w_{ij}} = \left( \sum_q^T \frac{\partial \lambda_q}{\partial u_q} \sum_{t_l^{in} \in x_T^i} \alpha_j(t_q - t_l^{in}) \right) \Delta t \qquad (53)$$

where the summation is performed for all time steps $q$ of interval $T$.

The computation of the second term (51) can be performed on each step for every $t_k$ by gradually summing up the gradients of the surprisals (52). At the end of interval $T$ we need to compute the first term in (51) using (53).

For the second term in (50) the following equation holds:

$$\frac{\partial}{\partial w_{ij}} H_{[0,T]} = \frac{\partial}{\partial w_{ij}} \hat{H}_{[0,t_1]} + P\{t_1\}(\frac{\partial}{\partial w_{ij}} \hat{H}_{[t_1,t_2]} - \frac{\partial h(\hat{y}_T(t_1))}{\partial w_{ij}} \cdot H_{[t_1,t_2]}(|t_1) + \\ + P\{t_2\}(\frac{\partial}{\partial w_{ij}} \hat{H}_{[t_2,t_3]} - \frac{\partial h(\hat{y}_T(t_2|t_1))}{\partial w_{ij}} \cdot H_{[t_2,t_3]}(|t_1) + \qquad (54)$$

Every entropy $H_{[t,T]}(|t_1)$ can be represented again as a sum of the two terms like in (50), so we can minimize it using the same approach again: first we minimize $\hat{H}_{[t_1,T]}(|t_1)$ using (51), (52) and (53), and then we recursively take care of $\Delta \hat{H}_{[t_1,T]}(|t_1)$.

The obtained equations for the full entropy derivative are based on the property of hierarchical additivity of entropy. They are not approximation yet and allow us to find the precise value of the gradient. They also suffer from the same drawbacks as the direct gradient computation method using (46): the number of the possible output patterns and computations grows exponentially with the length of $T$. However, the hierarchical representation of the derivative allows us to construct a reasonable approximation.

We can compute gradient $\frac{\partial \hat{H}_{[t_k,T]}}{\partial w_{ij}}$ online since we need only the current evolution of the state. If the probability of the spike at a certain time $t$ is high then the probability of the pattern $\hat{y}_T(t^{out} > t)$, in which the first spike happens after $t$ is small since the neuron most likely will generate the spike now as opposed to later:

$$P\{\hat{y}_T(t^{out} > t) | \Lambda(t) = \max\} \to 0.$$

Therefore, if at some point $t$ the probability of the spike is high, we can break the computation of the gradient $\dfrac{\partial \hat{H}_{[t_k,T]}}{\partial w_{ij}}$ because the contributions of the terms after $t$ are going to be negligible. It is likely that the neuron will actually generate a spike at $t$. However, it would be a mistake to break the computation of derivative just based on the output spike because it is likely that the peak of the spike probability is located after the actual spike. Let's assume that the approximate duration of the neuron's short-term memory about its input spikes is $t_M$. This duration characterizes how long an input spike can influence the neuron's output. For the SMRM neuron this time is equal to the time of the longest alpha function. We will use the short-term memory duration as a criterion to break the computation of the gradient after the output spike: $\dfrac{\partial \hat{H}_{[t_k,T]}}{\partial w_{ij}} \approx \dfrac{\partial \hat{H}_{[t_k,t^{out}+t_M]}}{\partial w_{ij}}$.

The remainder $\Delta \hat{H}_{[0,T]}$ characterizes the uncertainty of the neuron's behavior conditioned that the first spike happened at the beginning of the interval. If average intervals between the output spikes of the neurons are longer than the short-term memory duration $t_M$, then the output spike generation $t_1$ almost does not influence the output spike generation after $t_1 + t_M$. Therefore, the remainder can be approximately computed as the full entropy after the last output spike: $\Delta \hat{H}_{[0,T]} \approx H_{[t^{out},T]}$. Using these approximations and the hierarchical additivity property we get the following approximation for the gradient of the full entropy:

$$\frac{\partial}{\partial w_{ij}} H_{[0,T]} \approx \frac{\partial}{\partial w_{ij}} \hat{H}_{[0,t_1+t_M]} + \frac{\partial}{\partial w_{ij}} \hat{H}_{[t_1,t_2+t_M]} + \frac{\partial}{\partial w_{ij}} \hat{H}_{[t_2,t_3+t_M]} + ... \quad (55)$$

where $t_1, t_2, t_3, ...$ are the output spike times and the terms $\dfrac{\partial}{\partial w_{ij}} \hat{H}_{[t_i,t_i+t_M]}$ are computed using (51).

During the simulation an output spike initiates the computation of the new term in (55) while the old terms are kept being computed for the duration $t_M$ after the spike. Also we can assume now that interval $T$ is infinitely long because it does not appear on the right hand side of (55). In practice, if the neuron's firing rate is not too high, there are only a few terms of (55) being computed at each step (usually, only two terms are being computed). This allows us to use this algorithm online without breaking the training onto intervals.

### 3.2. An example of the unsupervised entropy minimization

Now we are going to consider a toy example of the unsupervised entropy minimization $H_{[0,T]}$ using the approximate method (55). We use the SMRM neuron described in the first chapter.

In this example an input pattern consists of two spikes arriving on two different input channels (Fig. 33, right-top). The weight vectors for the input channels are initialized with the following handpicked values: $w_1 = [0.1, 0.7, 0.3]; w_2 = [0.1, 0.6, 0.3]$. Notice that the weight vectors differ only in the second alpha function's weight. The weight values are picked to make the neuron's entropy initially high. Indeed, the probability distributions of the two output spikes after each input spike are shown in Fig. 33, right-center. The accumulated probability of the first output spike is slightly bigger that the second one. Also, the probability distributions are quite broad so the uncertainty of the neuron's output and therefore its entropy $H_{[0,T]}$ is high.

We performed 100 iterations of the entropy minimization with the stochastic gradient descent using (55). After each iteration we computed the full entropy using (45), the values are shown in Fig. 33 (left). After the training the weight vectors became equal to: $w_1 = [0.04, 1.73, 0.49]; w_2 = [0.01, -0.09, -0.14]$. The resulting probability distribution of the spike generation after the training is shown in Fig. 33 (right, bottom). Inspite of the initial proximity of the weight values, the final values are quite different. Initially the probability of an output spike after the first input spike was slightly larger than the silence, so in order to minimize the entropy the neuron have increased the spike probability and decreased the width of the probability distribution by increasing the weights on the first input channel. The probability of an output spike after the second input spike was smaller than the probability of silence, so the neuron decreased the weights on the second channel in order to increase the probability of an empty pattern.

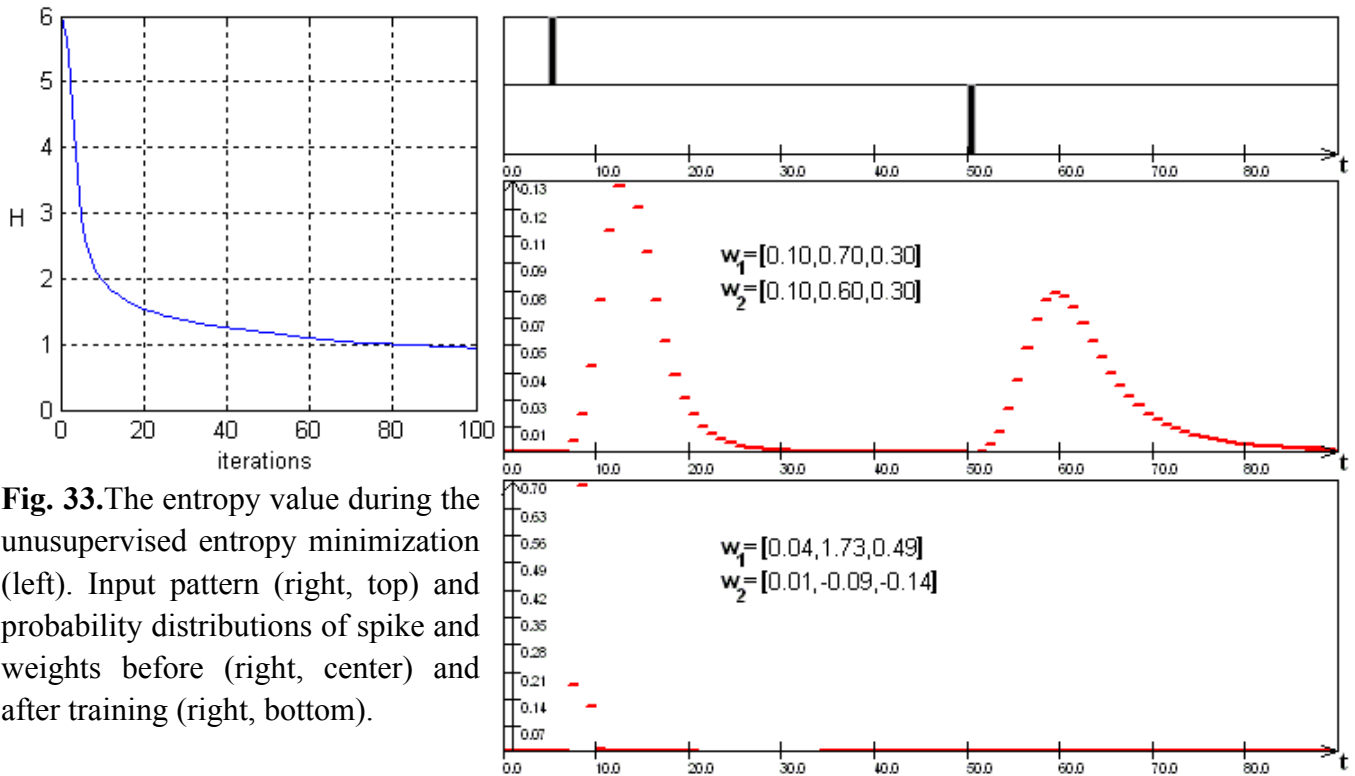

**Fig. 33.** The entropy value during the unusupervised entropy minimization (left). Input pattern (right, top) and probability distributions of spike and weights before (right, center) and after training (right, bottom).

## 3.3. The composite supervised and unsupervised learning algorithm

It is commonly assumed during supervised training that a "teacher" reacts on each input pattern with a supervised spike or silence. However, during online learning this assumption might not hold: the teacher might not have any information about the label of a particular input pattern. When the teacher is not active, the neuron can still adapt its weight in an unsupervised manner. Here we propose the following composite learning algorithm. When the neuron receives supervised spikes from the teacher, it tries to minimize the surprisal $h_T$ of the desired output. In the absence of the teacher's spikes, the neuron performs the unsupervised minimization of the entropy $H_T$. In other words, the teacher sometimes shows to the neuron the desired patterns and at all other times the neuron tries to learn these patterns without the teacher's help. The qualitative scheme of the probability distribution shaping during the composite training is shown in Fig. 34.

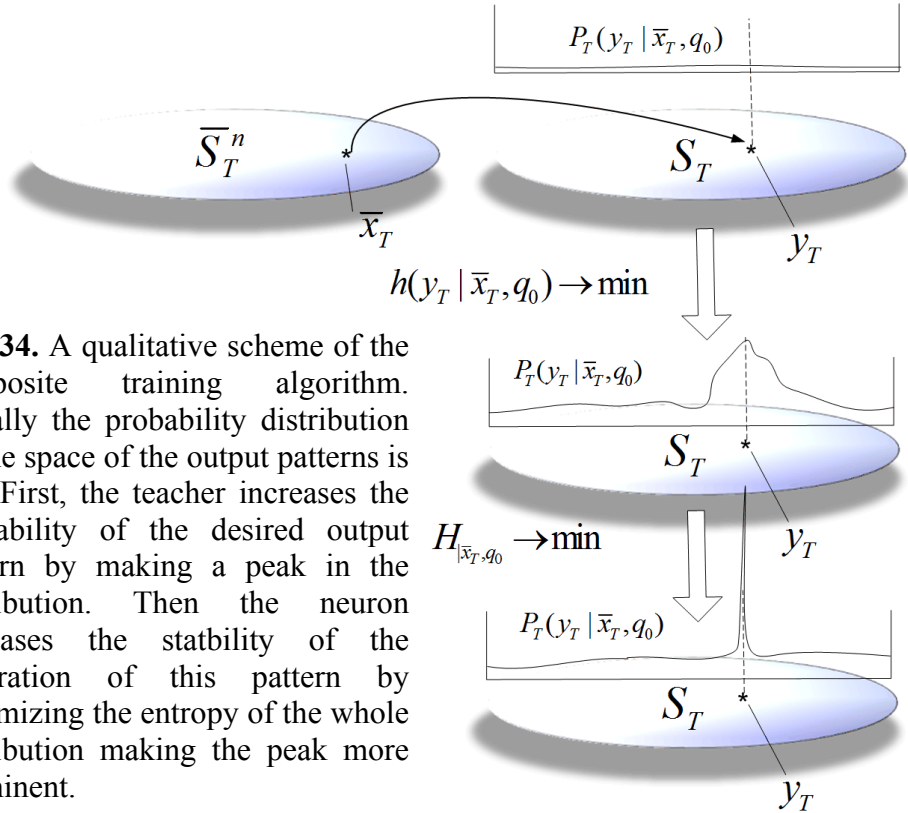

**Fig. 34.** A qualitative scheme of the composite training algorithm. Initially the probability distribution on the space of the output patterns is flat. First, the teacher increases the probability of the desired output pattern by making a peak in the distribution. Then the neuron increases the statbility of the generation of this pattern by minimizing the entropy of the whole distribution making the peak more prominent.

Let's assume that initially the neuron's weights are small and it does not generate spikes after the input pattern $\bar{x}_T$: the probability of an empty pattern $y_T^0$ is close to one. After that the teacher shows the desired response $y_T$ after the input pattern $\bar{x}_T$. The neuron adapts its parameters $\bar{W}$ using the surprisal minimization algorithm (34) so that the probability of the desired pattern $P_T(y_T|\bar{x}_T)$ makes a peak in the probability distribution of all possible output patterns. After that the neuron adapts the weights using the entropy minimization algorithm (55). A decrease of $H_T$ makes the peak in the probability distribution more prominent. If the probability of $P_T(y_T|\bar{x}_T)$ was high, then after the minimization of $H_T$ we can expect it to be even higher. Effectively the neuron learns the teacher's pattern without explicit teaching signals. As a result, the required number of presentations of the desired pattern decreases. If a peak in $P_T(y_T|\bar{x}_T)$ was not high enough after the supervised phase, the minimization of the entropy $H_T$ might lead to the increase of the probability of an empty pattern $y_T^0$ again as we have seen in the toy example in the previous chapter. This property might have a positive effect – if the teacher's signal is noisy or the teacher has made a mistake selecting the wrong label, the mistake will be suppressed during the entropy minimization and such desired pattern will be forgotten.

Stochastic gradient descent process is characterized by the trajectory of parameters $\bar{W}$ on the cost function landscape. During the surprisal minimization the cost landscape is given by the surprisal $h_T(\bar{W})$ as the function of the parameters. During the entropy minimization the cost landscape is given by $H_T(\bar{W})$. During the composite training we would like the minimization of $H_T(\bar{W})$ to lead to minimization of the surprisal $h_T(\bar{W})$. Therefore, the result of the composite training algorithm depends on the properties $H_T(\bar{W})$ and $h_T(\bar{W})$. In particular, it depends on the proximity of minima of the both costs in the parameter space $\bar{W}$. The trajectory of the parameters should lead to minima of both functions.

For a particular output $y_T^*$ the surprisal $h_T(\bar{W})$ is a convex function of the neuron's weights and every its local minimum is also global (see chapter 2). The entropy $H_T(\bar{W})$ has a more complex landscape: the value is close to zero $H_T \to 0$ for every output pattern $y_T$ that is generated robustly enough ($P(y_T) \to 1$). Intuitively, the number of local minima of $H_T(\bar{W})$ equals to the number of the possible output patterns on the interval $T$. The convergence of the composite algorithm depends on the locations of the global minimum of surprisal $h_T(\bar{W})$ and the local minimum $H_T(\bar{W})$ that corresponds to the desired output $y_T^*$. The strict requirements of convergence of the composite algorithm require further research, however the experiments described below show that for the SMRM neuron the minima of both functions are indeed close in the parameters space.

Let's consider a toy task of learning a delay between the input and the output in order to illustrate the composite learning algorithm. Assume that the SMRM neuron has only one input channel and two alpha functions with weights $w_1$ and $w_2$. Alpha functions maximums are equal to 1ms and 10ms. The time discretization step is equal to 1ms. The input pattern consists of a single input spike at the beginning of interval $\bar{x}_T = \{t_0^{in} = 0\}$. The desired output pattern consists of a single output spike at 6ms $y_T^* = \{t_0^d = 6\}$. The task is to train the neuron to generate the desired spike using as little supervised iterations as possible.

The surprisal landscape for the described task is shown in Fig. 35. It can be seen that $h_T(w_1, w_2)$ is convex but it has a lot of plateaus where the speed of the stochastic gradient descent is very low.

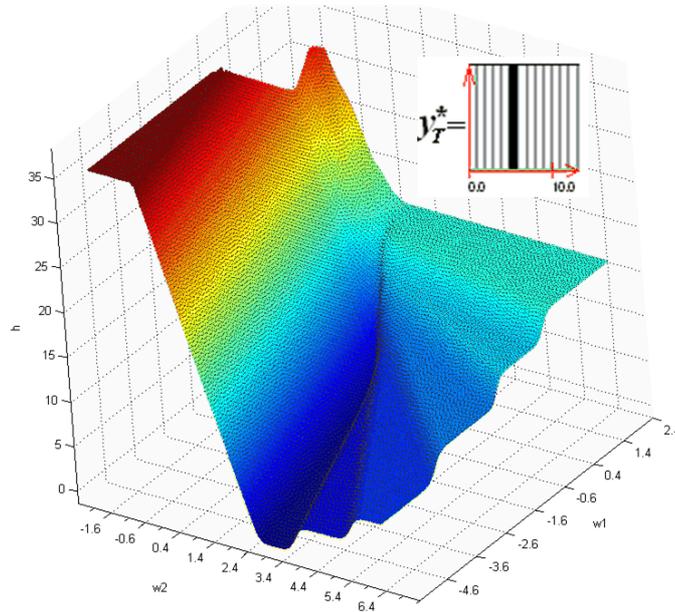

**Fig. 35.** The surprisal $h_T(w_1, w_2)$ landscape as a function of two weights. The desired output pattern $y_T^*$ is displayed on top. Notice the convexity of the surprisal and the existence of plateaus.

The landscape of the entropy $H_T(w_1, w_2)$ is shown in Fig. 36. Troughs of local minima correspond to the robust generation of possible output patterns from the space $S_T$.

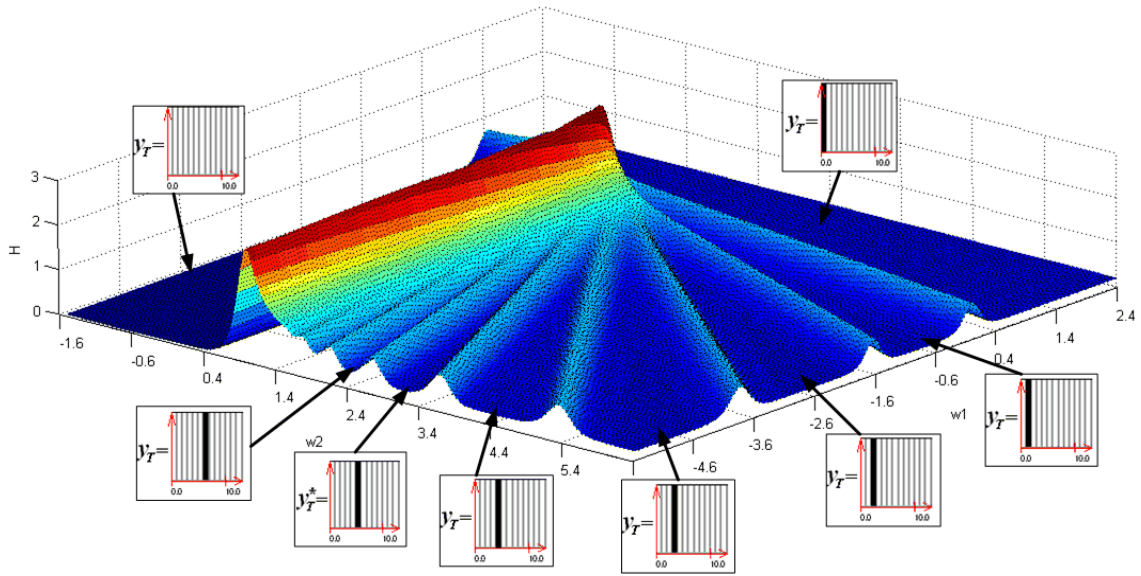

**Fig. 36.** The entropy $H_T(w_1, w_2)$ landscape as a function of weights. Every output pattern has a corresponding local minimum.

The trajectory of the weights evolution can be quite different depending on the initial conditions, the order and the duration of the supervised and unsupervised training phases. We are going to describe several examples of the weights evolution below.

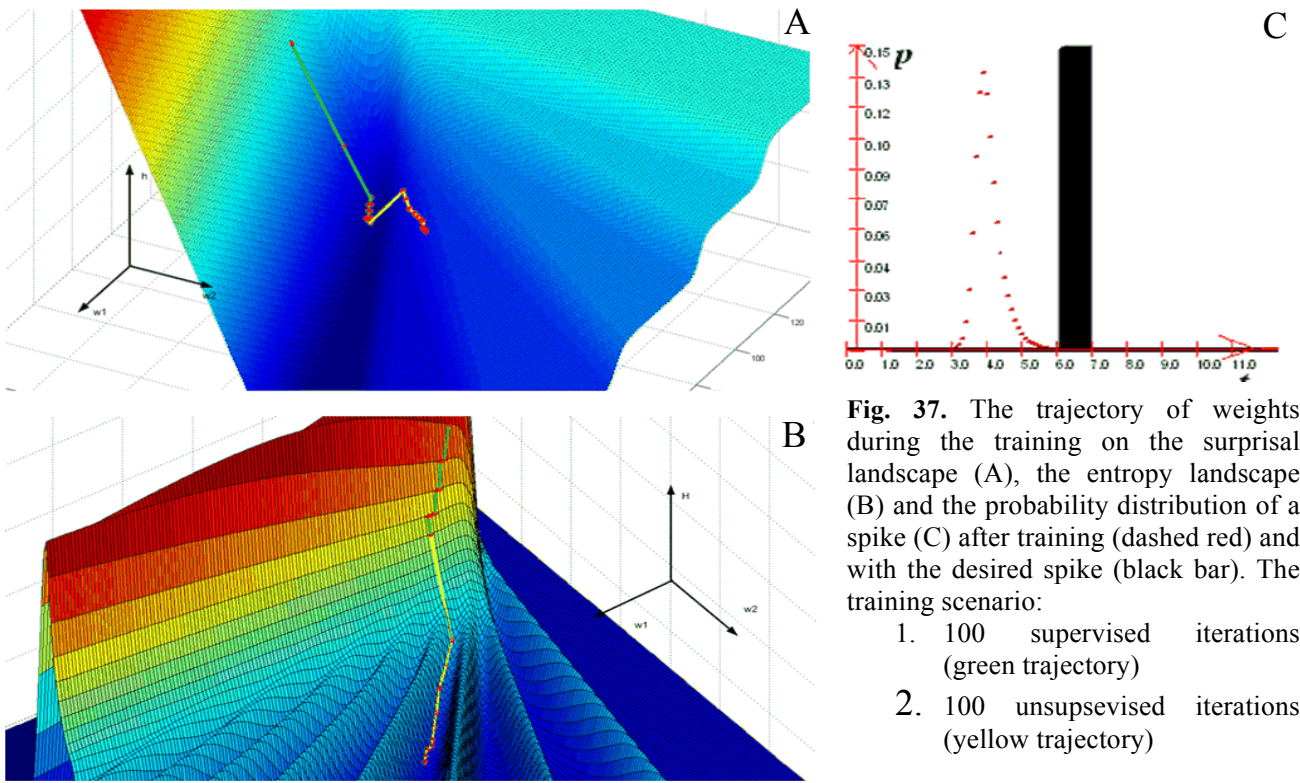

**Fig. 37.** The trajectory of weights during the training on the surprisal landscape (A), the entropy landscape (B) and the probability distribution of a spike (C) after training (dashed red) and with the desired spike (black bar). The training scenario:
1. 100 supervised iterations (green trajectory)
2. 100 unsupsevised iterations (yellow trajectory)

The weights trajectory after 100 supervised iterations and 100 unsupervised iterations is shown in Fig. 37. During the supervised training the parameters move towards the surprisal global minimum. After reaching the plateau the speed of training slows down and the value of $h_T(w_1, w_2)$ and probability of the desired pattern doesn't change much. After 100 iterations we start an unsupervised training phase during which the parameters move to the closest local minimum of $H_T(w_1, w_2)$. However, it happened that the closest minimum is not the minimum of the desired pattern $y_T^*$, so after the unsupervised training the neuron robustly generates the wrong pattern $y_T = \{t_0^{s\_out} = 4\}$. The probability distribution of the output spike time is shown in Fig. 37 (C). We can see that in this example the neuron was not trained enough with the teacher so the peak in the probability distribution of all possible patterns was not prominent enough for the entropy minimization to work as desired.

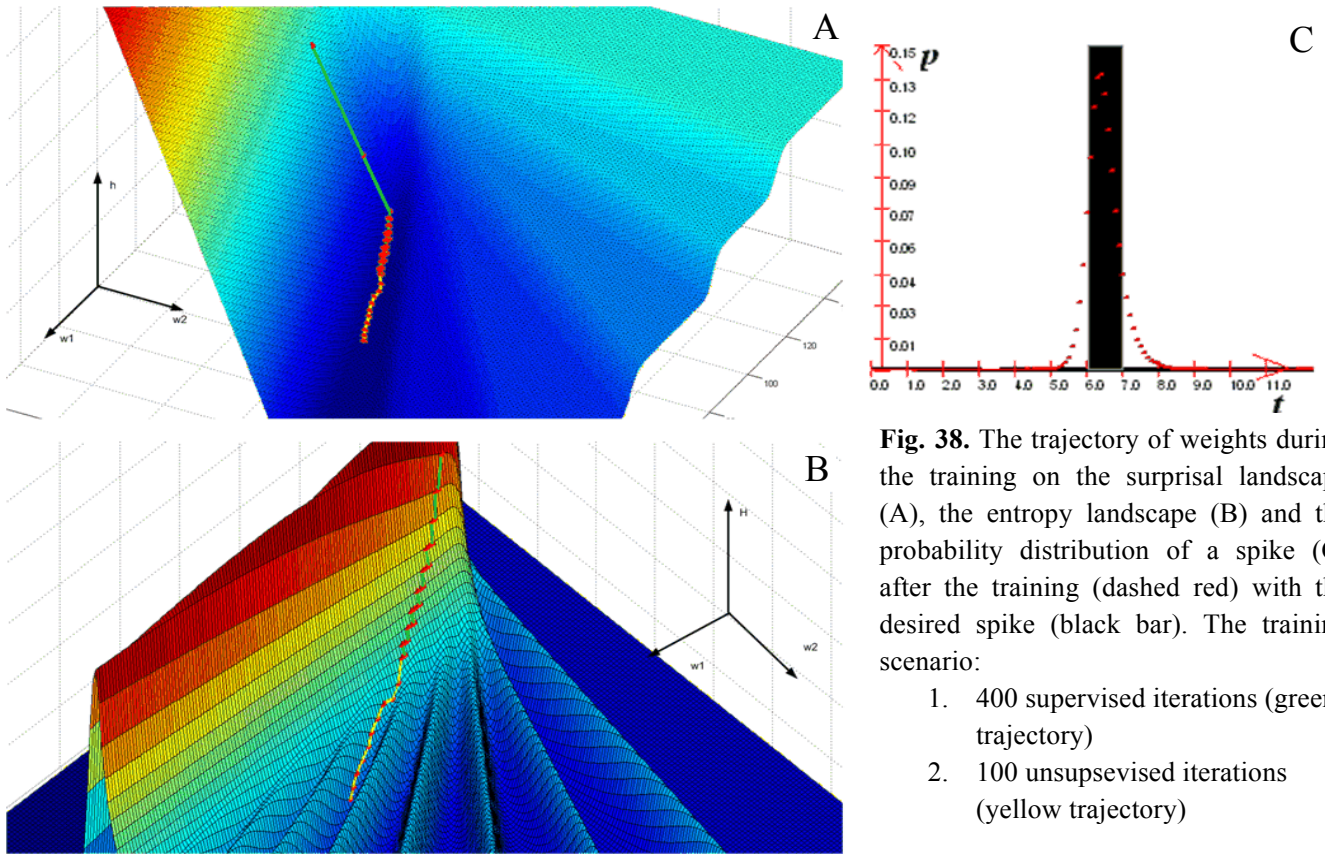

Fig. 38. The trajectory of weights during the training on the surprisal landscape (A), the entropy landscape (B) and the probability distribution of a spike (C) after the training (dashed red) with the desired spike (black bar). The training scenario:
1. 400 supervised iterations (green trajectory)
2. 100 unsupsevised iterations (yellow trajectory)

The weights trajectory after a much longer supervised training (400 iterations) and the same unsupervised training (100 iterations) is shown in Fig. 38. The longer supervised training process creates a larger peak in the output patterns probability distribution at the point of the desired pattern $y_T^*$. Because of that the closest local minimum of $H_T(w_1, w_2)$ corresponds to this desired pattern. The resulting probability distribution of the output (Fig. 39 (C)) is indeed shaped as desired with the peak at $y_T^* = \{t_0^{s\_out} = 6\}$. We can see that because of the plateaus in $h_T(w_1, w_2)$ landscape the supervised training has to continue for a long time for the composite algorithm to work. Notice that moving on the $H_T(w_1, w_2)$ landscape during the unsupervised learning corresponds to a much faster movement on the plateau of $h_T(w_1, w_2)$. This property is used in the experiment below.

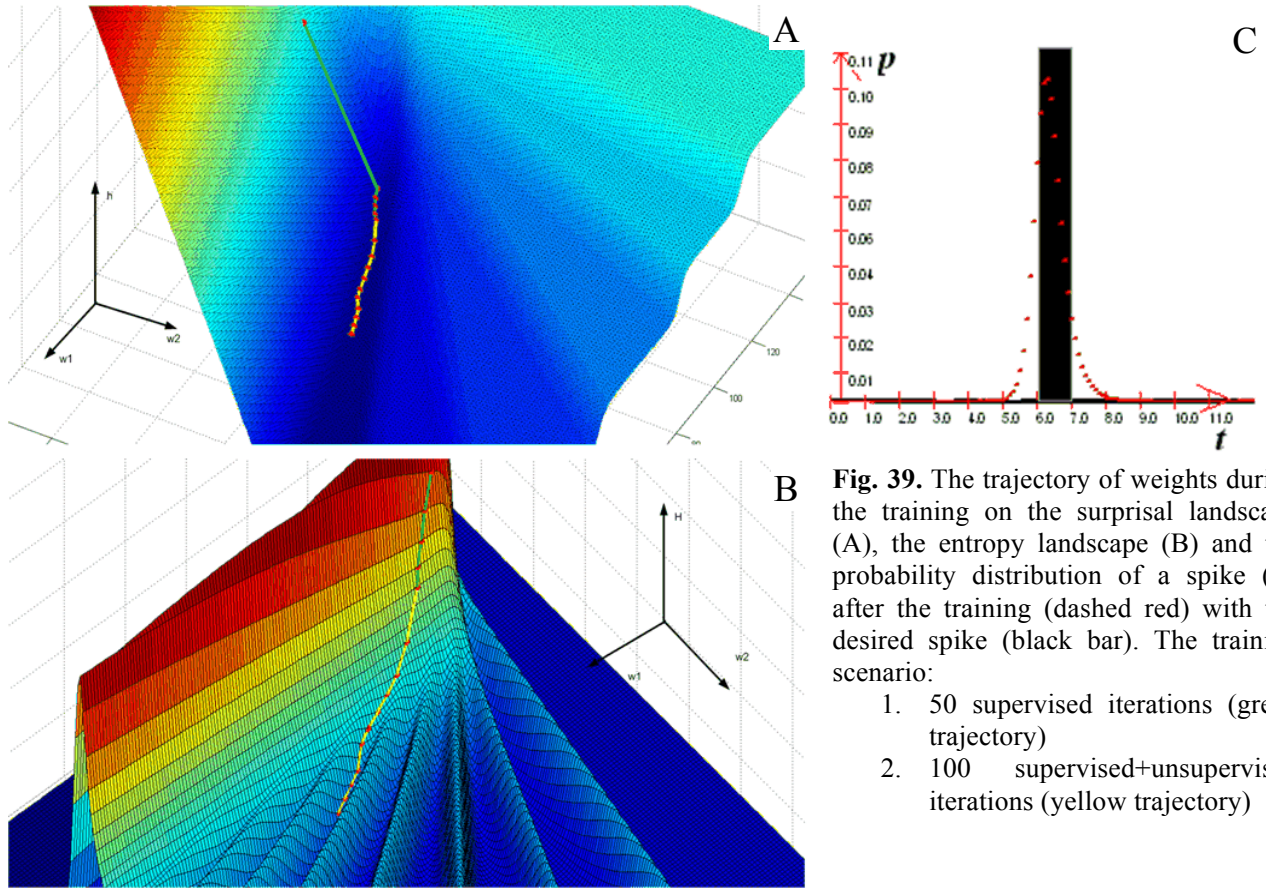

**Fig. 39.** The trajectory of weights during the training on the surprisal landscape (A), the entropy landscape (B) and the probability distribution of a spike (C) after the training (dashed red) with the desired spike (black bar). The training scenario:
1. 50 supervised iterations (green trajectory)
2. 100 supervised+unsupervised iterations (yellow trajectory)

In the last experiment we first perform a short supervised pretraining for 50 iterations to avoid the empty pattern local minimum. Then we apply the supervised and unsupervised weight changes at each iteration for 100 iterations minimizing $h_T(w_1, w_2)$ and $H_T(w_1, w_2)$ simultaneously (Fig. 39). This training method also achieves the desired probability distribution (Fig. 39 (C)). The neuron robustly generates the desired pattern $y_T^* = \{t_0^{s\_out} = 6\}$ using less supervised examples (150 vs 400). As before, the weight trajectory reaches the plateau of $h_T(w_1, w_2)$ during the supervised pretraining. However now the minimization of $H_T(w_1, w_2)$ moves the parameters much faster on the plateau while the constant supervised corrections do not allow the trajectory to fall into the wrong local minimum of $H_T(w_1, w_2)$.

# Chapter 4. Reinforcement learning of spiking neurons

**Summary:**

In this chapter we develop a spiking neural networks for controlling a virtual agent. The network uses a reinforcement feedback from environment to improve its control strategy. Reinforcement learning rules for spiking neurons are formulated using the framework of entropic cost functions.

The reinforcement learning rules for spiking neurons are tested on several simple control tasks. First, we solved a task of resource collecting in an empty discrete grid environment. We tried several network architectures and showed that a two-layer spiking neural network is able to solve the task. The network uses the temporal structure of input patterns in the lack of spatial information. We performed experiments in which agent's actuators fail. The network is able to find a new control strategy after one of the agent's actuators stopped functioning correctly. Next, we solved a virtual soccer player robot navigation task on an empty field. The robot has to reach a ball without touching the field borders. Finally, we performed an experiment in which we change the object controlled by the network. We trained a network to control the agent in the grid environment and then replaced the agent with the virtual soccer robot. The network successfully relearned to control the new object.

## 4.1. A spiking neural network as a control system

Real biological control systems of animals consisting of spiking neurons usually perform better than the state of the art artificial control systems in tasks that require a quick adaptation in unknown environments, an adaptation to the animal body changes and changes in the task objectives. However the usage of biologically inspired artificial spiking networks as control systems is quite limited. The majority of the control spiking neural networks are used in the simple navigation and obstacle avoidance tasks in small virtual environment [68-71, 45], in simplified real environments [72–74, 41–44] and in two joint manipulator reaching tasks [75-77]. We speculate that their limited use is caused by the lack of well-developed methods for their training.

The spiking neural network has to have information about the task in order to adapt and improve its control strategy. Such information can be prewired into the network based on the neuroscientific data [41,42,78,76]. Alternatively, if the control law is known, a researcher can use the supervised learning [75,76]. Also it is popular to use randomly connected recurrent spiking networks - Liquid State Machines (LSM) [79]. Rich internal dynamics of LSMs allows them to represent history of inputs and their temporal relations in the instantaneous activity of a large number of neurons (a reservoir). One can train readout elements using simple learning rules to learn the control policy [75,68,72,39]. If the control law is unknown, one can use genetic algorithms to evolve the spiking network controllers [43-45].

An alternative way to train the spiking network controller is to use the reinforcement learning. The reinforcement learning algorithms are usually developed for the Markov decision processes [54,80,81]. An extension of such algorithms to the spiking neurons domain has happened quite recently [71,82,83,69]. In this chapter we will show the connection between the learning rules based on Markov decision process formalism and the entropic cost function minimization algorithms developed in the previous chapters.

Consider a spiking network controlling an object (e.g. a robot) (Fig. 40).

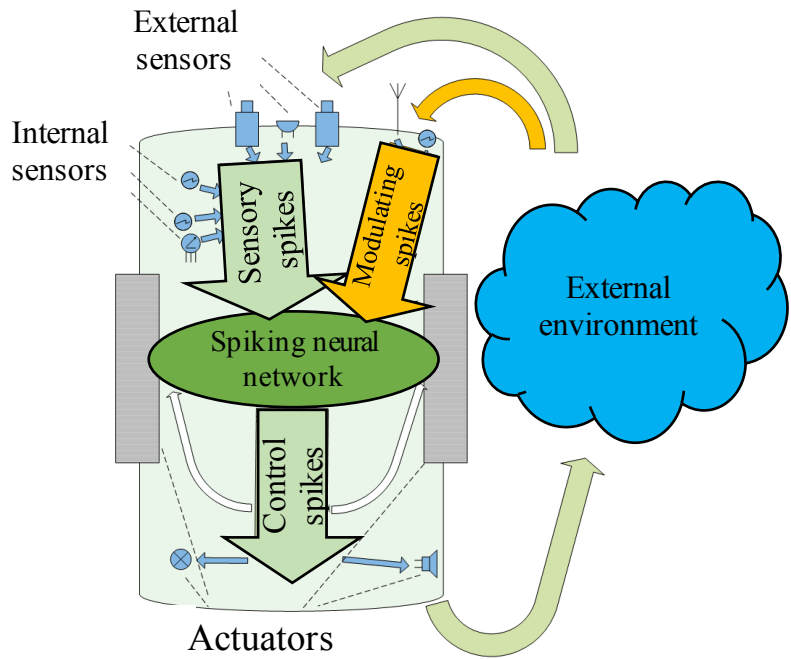

**Fig. 40.** The spiking network controlling the robot. The robot provides sensory and modulatory spikes to the network. Output spikes of the network control the robot's actuators.

The network receives a multidimensional stream of spikes from the robot's sensors (for example, visual and distance measurements, positions of actuators or a power level). Each sensor has to encode data into the stream of spikes. For example, an analog signal can be encoded using biologically inspired receptive fields approach [84]. The input spike stream from all sensors can have complex spatiotemporal structure that mirrors the structure of external and internal events of the robot. Outputs of the network are connected to the actuators of the robot or to the internal systems of the robot (e.g. power system). The actuators have to decode the spiking signals from the network in order to apply control commands.

The network also receives an input stream of modulating signals such as reinforcement signals with information about the network's performance. If the task is performed well the network receives a positive reinforcement signal. If the network makes undesired actions, it receives a negative reinforcement signal. The reinforcement input is separated from the sensory input since it has a different semantics and typically used only during learning. The network training is based on the action-reinforcement feedback loop. If positive reinforcement is received, the network has to analyze the history of its input and output spikes. After that the network has to infer how to change its parameters in order to generate control signals that lead to reinforcement. After the negative reinforcement the network should change its parameters in the opposite way in order to prevent generation of the similar actions in the similar sensory context.

One of the core problems in the reinforcement learning theory is an exploration/exploitation dilemma [54, 85]. The controller has to find a tradeoff between the exploratory behavior that can lead to a discovery of new sources of reinforcement and the exploitation behavior using already obtained knowledge about the environment that guarantees a certain level of reinforcement. In this thesis we introduce an additional modulatory signal indicating that the current control strategy does not lead to enough reinforcement. This signal controls the exploration/exploitation bias in behavior. Intuitively, this signal can be viewed as a "hunger". After receiving such signal, the network has to explore the possible control strategies more actively.

### 4.2. A neuron model for a spiking network controller

We use the SMRM neuron with 3 alpha functions as a basic element of the control network. The membrane potential evolves according to (26) and probabilities of output spike patterns are computed using (11).

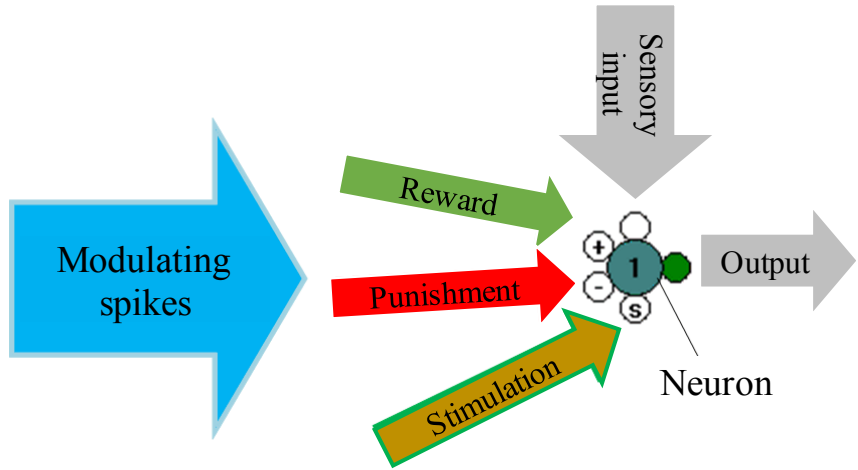

**Fig. 41.** Different types of the sensory and modulatory signals arrive at different types of input channels. Sensory spikes arrive at the sensory input. Modulating spikes arrive at three different input channels depending on their semantics.

Sensory and modulatory signals are all represented as spiking sequences. However, since those signals have different semantics the neurons use separate types of input channel pins to receive the signals (Fig. 41). The sensory input pin receives a multidimensional stream of spikes from the sensors of the robot and other neurons. Positive and negative reinforcement spikes arrive to the channel pins denoted by «+» and «-» respectively. After receiving the reinforcement spikes the neuron adapts its parameters in order to increase future cumulative reinforcement. The last type of a modulatory input signal pin denoted by "s" (from "stimulating" or "search") increases the exploratory activity of the neuron. Here we use a simple random exploration technique. A spike that arrives on this input adds the following value to the membrane potential $\Delta u^{st}(t) = w^{st}\alpha^{st}(t - t^{st})$, where $\alpha^{st}(t), w^{st}$ are the alpha function and the weight of the stimulating input, $t^{st}$ is the time of the stimulating spike. Such membrane potential increase makes the neuron generate an output spike with a higher probability. The parameters of simulating input are chosen to increase the randomness of spike times on average and therefore to increase the exploratory nature of the controller's actions.

### 4.3. Reinforcement learning as modulated entropy minimization

Reinforcement learning is the process of finding a control policy that maximizes a cumulative future reward [54]. The developed in [80] «OLPOMDP» algorithm maximizes the cumulative future reward when a policy is a continuous function of some parameters. It was shown in [86] that when this algorithm is applied to the individual elements of the network, it will maximize the cumulative reinforcements of the whole network. A stochastic spiking neuron activity can be formalized as a Markov decision process where the policy actions are generated spikes in discrete time and the parameters of the policy are neuron's weights [71, 82, 83, 69]. An application of OLPOMDP algorithm to the spiking neuron in continuous time is done in [71]. In the notation of [71] the learning rules for the parameter $w_j$ have the following form:

$$\frac{dw_j}{dt} = \gamma r(t) z_j(t)$$

$$\tau_z \frac{dz_j(t)}{dt} = -z_j(t) + \left( \sum_{t_q^{out} \in y_T} \frac{\delta(t - t_q^{out})}{\lambda(u(t))} - 1 \right) \frac{\partial \lambda(u(t))}{\partial w_j} \qquad (56)$$

where $r(t)$ is the reinforcement signal value, $z_j(t)$ is the eligibility trace, $\gamma$ is the learning coefficient, $\tau_z$ is the reinforcement discount parameter that characterized the memory of Markov decision process. The eligibility trace value characterizes the compressed history of the spikes on a given input channel.

In this thesis the reinforcement signal comes as positive or negative reinforcement spikes and can be computed using the delta functions: $r(t) = \sum_{reward} \delta(t - t^r) - \sum_{pain} \delta(t - t^p)$, where $t^r$ and $t^p$ are the times of positive and negative reinforcement spikes.

Recall, that in the second chapter we derived the learning rules for the SMRM model in continuous time based on the surprisal minimization on the interval $T$:

$$\Delta w_{ik}^T = \gamma \sum_{t_l^{out} \in y_T} \sum_{t_j^i \in x_T^i} \frac{1}{\lambda(u(t_l^{out}))} \frac{\partial \lambda(u(t_l^{out}))}{\partial u} \alpha_k(t_l^{out} - t_j^i) - \gamma \sum_{t_j^i \in x_T^i} \int_T \frac{\partial \lambda(u(s))}{\partial u} \alpha_k(s - t_j^i) ds \qquad (57)$$

where $t_l^{out} \in y_T$ is the desired output spike time, $t_j^i \in x_T^i$ is the input spike time in the $i$-th input channel, and $i, j$ are indices in the $n \times 3$ weight matrix of the neuron.

Let's extract the gradient from the (57):

$$\frac{dw_{ik}}{dt} = -\gamma g_{ik}(t)$$

$$g_{ik}(t) = (1 - \sum_{t_l^{out} \in y_T} \frac{\delta(t - t_l^{out})}{\lambda(u(t))}) \sum_{t_j^i \in x_T^i} \frac{\partial \lambda(u(t))}{\partial u} \alpha_k(t - t_j^i) \tag{58}$$

where $g_{ik}(t)$ is the gradient of the surprisal with respect to the weight $w_{ij}$ at the time moment $t$.

Notice that reinforcement learning rules (56) contain a very similar term to the gradient $g_{ik}(t)$ so the reinforcement learning rules can be rewritten as:

$$\frac{dw_{ik}}{dt} = \gamma (\sum_{reward} \delta(t - t^r) - \sum_{pain} \delta(t - t^p)) z_{ik}(t)$$

$$\tau_z \frac{dz_{ik}(t)}{dt} = -z_{ik}(t) - g_{ik}(t) \tag{59}$$

The solution of the eligibility traces function $z_{ik}(t)$ with zero initial conditions is a negative convolution of the surprisal gradient with the exponential kernel $e^{-t/\tau_z}$: $z_{ik}(t) = -\frac{1}{\tau_z} e^{-t/\tau_z} * g_{ik}(t)$. The eligibility trace function accumulates the surprisal gradient roughly in a direction of a gradient of the probability of the recent output pattern $y$ conditioned on the input pattern $\bar{x}$. The most recent outputs are amplified and the old history of the output pattern $y$ is forgotten. When the reinforcement spike comes, the weights change proportionally to the surprisal gradient filtered in this way. In particular, after the positive reinforcement the probability of the recent output increases (the surprisal is decreased) and after the negative reinforcement the probability of the recent output decreases. The schematic process of the modulated surprisal change is shown in Fig. 42.

Using the same technique, we can obtain the reinforcement learning rules in discrete time from the continuous time case using the gradients computed for the supervised learning. For that we need to compute the gradient of the surprisal of the current neuron's behavior as opposed to the surprisal of the desired pattern. Another difference with supervised learning gradient is the exponential discount of the processing history. The discount time $\tau_z$ depends on the average response time of the controlled object [80] which indicates the duration after which the network gets a reinforcement signal in response to the previously generated control.

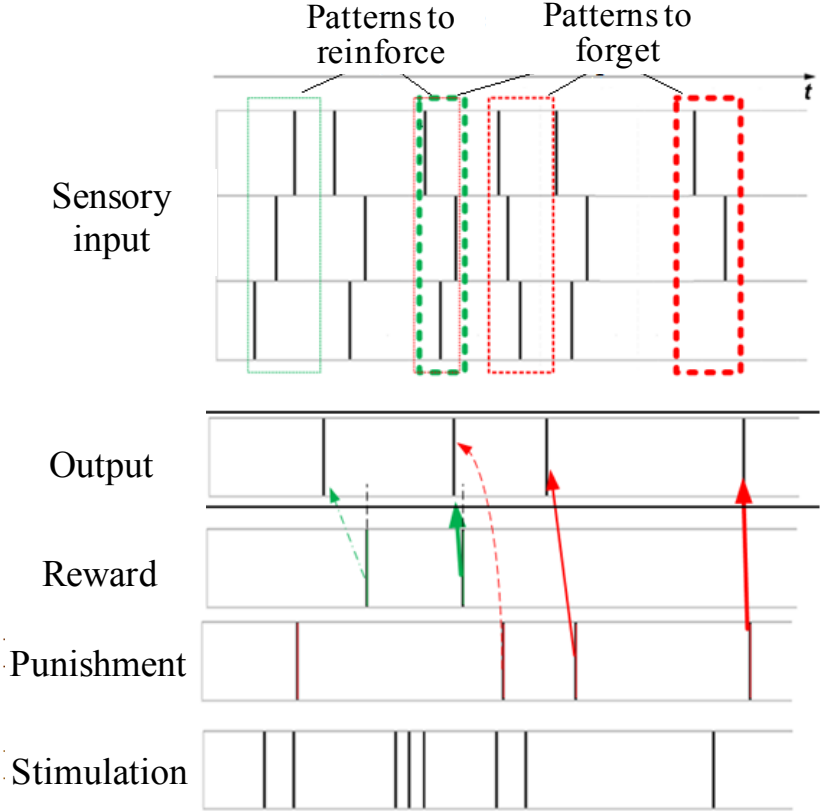

**Fig. 42.** Modulated pattern detection during reinforcement learning. The sensory spikes arrive at the sensory input. The stimulation and the input spikes lead to the generation of the output spikes. The information about the most recent input and output spikes is stored in the eligibility traces and old input and output spikes are forgotten. When the reinforcement arrives, the strength of the response to the particular input patterns is changed. The positive/negative reinforcement spike (reward/punishment) leads to the increase/decrease in the strength of the recent input-output transformations. If the neuron have generated output spikes, the preceding input patterns (shown with arrows) are reinforced or forgotten. The width of the arrow indicates the strength of the reinforcement or forgetting of the input pattern.

Every simulation step $t_l$ for the SMRM neuron during reinforcement learning contains the following operations. First, the membrane potential value $u(t_l)$ is computed based on the sensory spike history. After that we compute the derivative $\frac{\partial \lambda(u(t_l))}{\partial u}$. Then we compute the surprisal gradient $g_{ik}$ for every weight. Next we compute $\Lambda(t_l)$ and determine whether the neuron should generate an output spike. In the absence of an output spike the gradient value increases:

$$g_{ik} = e^{-\Delta t/\tau_z} g_{ik} + \frac{\Delta t}{\tau_z} \frac{\partial \lambda(u(t_l))}{\partial u} \sum_{t_j^i \in x_T^i} \alpha_k(t_l - t_j^i).$$

If there is an output spike, the gradient decreases:

$$g_{ik} = e^{-\Delta t/\tau_z} g_{ik} - \frac{\Delta t}{\tau_z} \frac{1 - \Lambda(t_l)}{\Lambda(t_l)} \frac{\partial \lambda(u(t_l))}{\partial u} \sum_{t_j^i \in x_T^i} \alpha_k(t_l - t_j^i),$$

If the reinforcement spike arrives, we change the weights as follows. If the reward (positive reinforcement) spike arrives, we change the parameters against the gradient:

$$w_{ik} = w_{ik} - \gamma g_{ik},$$

and if the punishment (negative reinforcement) spike arrives we change the parameters along the gradient:

$$w_{ik} = w_{ik} + \gamma g_{ik},$$

after that we reset the gradient: $g_{ik} \leftarrow 0$.

In chapter 2 we showed that after the integration of the supervised learning rules (57) the weight change curve resembles the biological Spike Timing Dependent Plasticity (STDP) [23]. During the reinforcement learning the exponential decay of the eligibility traces in (59) and the modulation of the weight changes in (57) resemble Modulated STDP [71], that was also observed in real neurons and further investigated in simulations [87-89].

## 4.4. A discrete grid virtual agent controlled by a spiking neural network

To test spiking network controllers we developed a simple virtual agent in an empty discrete grid dimensions (a 3x3 grid is shown in Fig. 43). This environment is similar to the environment used in [70] for investigating some heuristic reinforcement learning rules for spiking networks. The agent can freely move between the grid cells. The boundaries are surrounded by walls. The position of the agent is depicted with a black circle. The grid cells might contain resources that can be consumed by the agent. A resource location is depicted by a small green circle.

At the beginning of simulation, the agent has a certain amount of energy. The energy slowly decays and the agent starts to feel "hunger" (depicted by the change of agent's color in the videos below). To order to replenish the energy the agent needs to find the cell with a resource. After that the agent stops feeling hunger. The task of the agent is to supply itself with the energy and do not bump into the walls.

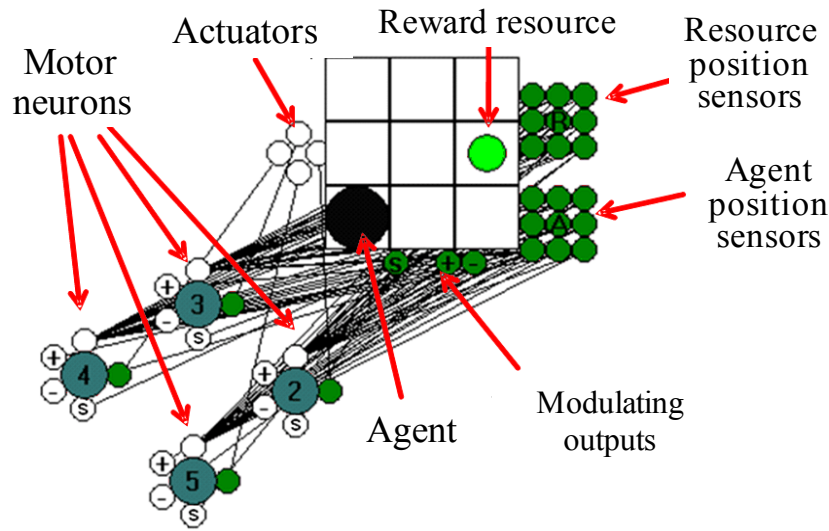

**Fig. 43.** A virtual agent in a grid environment controlled by a spiking network consisting of 4 neurons.

The agent is controlled by a spiking network that receives input and modulatory spike signals. Such spiking neural network consisting of four neurons is shown in Fig. 43. There are two global position sensors for the agent and for the resource (shown on the right of the grid with dark green circles). Each global position sensor has a number of outputs being equal to the dimensions of the grid. In the 3x3 environment the network receives 9+9=18 inputs. If a cell is occupied with an object, the corresponding neuron in the sensor array will start to fire with a certain rate. For example, if the agent is on the bottom left, the bottom left sensory neuron of the agent sensor array is going to generate spikes.

The agent is controlled by the means of sending spikes at its actuation inputs (Fig. 43, 4 white circles on the left of the grid). The spikes received on the actuating inputs move the agent on the grid (left, right, up, down). The neurons are split into actuation groups so that the neuron from a group is responsible for a particular actuator (for example, 4 neurons on Fig. 43 are connected one-to-one to 4 actuators).

The modulating output pins from the environments are shown on the bottom of the grid denoted with "+", "-" and "s" (Fig. 43). The agent receives a reward spike if it consumes the resource and a punishment spike if it bumped into the boundaries of the grid. Additionally, it receives a punishment spike if it received an ambiguous control signal (left+right or up+down). The reinforcement outputs "+" and "-" are connected to the corresponding inputs of all neurons.

If the energy is depleted the "stimulation" output "s" starts to generate spikes with a rate proportional to the energy depletion. This signal indicates that the controller needs to take some new actions because the current control policy does not work well. The stimulation spikes also arrive at stimulation inputs ("s") of all neurons.

### 4.4.1. Training process of a virtual grid agent

At the beginning of the experiment the agent has a certain amount of energy and does not experience hunger. The modulation outputs of the environment are silent. The global position sensors are active and generate spikes depending on the position of the agent and the resource. All input weights of all neurons are initialized with zeroes, and the neurons do not generate output spikes except for rare stochastic spikes. Soon the energy of the agent decreases and it starts to experience hunger. The stimulating output starts to generate spikes with a slowly increasing rate. The larger the stimulation spike rate, the more frequently the neurons generate the output spikes at random time moments. These random spikes arrive at the agent's actuators and make the agent move on the grid. During such random movement the agent might accidentally occupy a cell with the resource and consume it. The resource consumption will lead to a single reward spike arriving at all reward inputs "+" of all neurons. This triggers learning by rules (59) where $r(t) = \delta(t^r)$. This increases the probability of generating the same movement action in the same sensory context (the same positions of the agent and the resource). If some neurons generated bad control signals that led to bumping into the walls or to an ambiguous control command, the neurons will receive a negative reinforcement spike on their "-" output. In this case the weights of the neurons are also changed according to the rules (59) however now $r(t) = -\delta(t^p)$. This will decrease the probability of the similar behavior in similar sensory context. During initial random exploration, the neurons will receive various reinforcement signals at various times. However, on average the network will learn to generate the most beneficial actions for the agent: the agent will go directly towards the resource without bumping into the walls and without receiving ambiguous control signals.

### 4.4.2. Various control network architectures

The result of training depends on the network architecture. The simplest architecture is shown in Fig. 43. This network consists only of four neurons. Each neuron is responsible for a particular movement ("up", "down", "left", "right"). This network successfully solves the task of consuming resources on a 3x3 grid, provided that the firing rate of the position sensors is high enough so that periods between spikes are smaller than the alpha functions length. In this case every motor neuron is always aware of the positions of the agent and the resource when the reinforcement signals arrive. The video of the agent's behavior after the training is available here: http://www.youtube.com/watch?v=yl2rcSKHfLU. The training process and the results would be similar to the non-temporal binary neural networks because the neurons use current spatial locations of the agent and the resource and do not use any temporal information. For example, the "up" neuron learns to generate a spike when the agent is located below the resource. Also this neuron is inhibited if the agent is at the top row of the grid to avoid bumping into the wall.

The network of 8 neurons that is shown in Fig. 44 also uses only spatial information. It learns faster than the 4 neuron network because now neurons can start separating responsibilities for actuators. However, there is no substantial improvement in the quality of the resulting policy: http://www.youtube.com/watch?v=5_8oLnpDeW4.

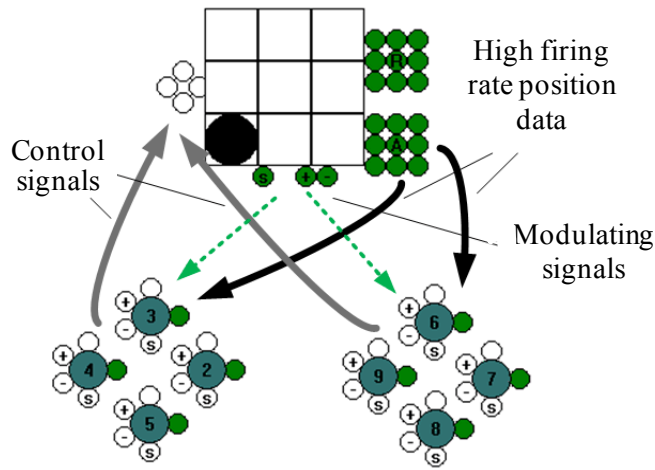

**Fig. 44.** Spiking network consisting of 8 neurons controls the virtual agent.

If the firing rate of the position sensors is low, then neurons in the network (Fig. 44) are not able to learn a good control policy. In this case neurons often receive the reinforcement spikes after the sensory context has already been forgotten because of the limited length of the alpha functions. The failure to learn a policy leads to an increase in the stimulation signal rate and in the amount of the random exploration. This in turn leads to an excessive amount of the negative reinforcement due to bumps with the boundaries. This leads to forgetting of any positive associations. In order to keep the information in the network, we added recurrent connections between the neurons and another 4 neurons as the second layer (Fig. 45). In this network the neurons have the information about the activity of other neurons. Experiments (http://www.youtube.com/watch?v=SfUHtyLDv80) show that the neurons learn to generate sequences of activity that are triggered by the position sensors or the stimulation input. Control commands are generated more frequently than incoming position sensor spikes so that the agent can make several movements between the sensory updates. The network uses the recurrent connections and the second layer to learn spatiotemporal associations between the positions and the commands. Also notice that the activity does not explode or dies out which is a common problem in recurrent networks. A small activity is kept up by the stimulation signal and an excessive activity is penalized by the negative reinforcement from the wall bumps.

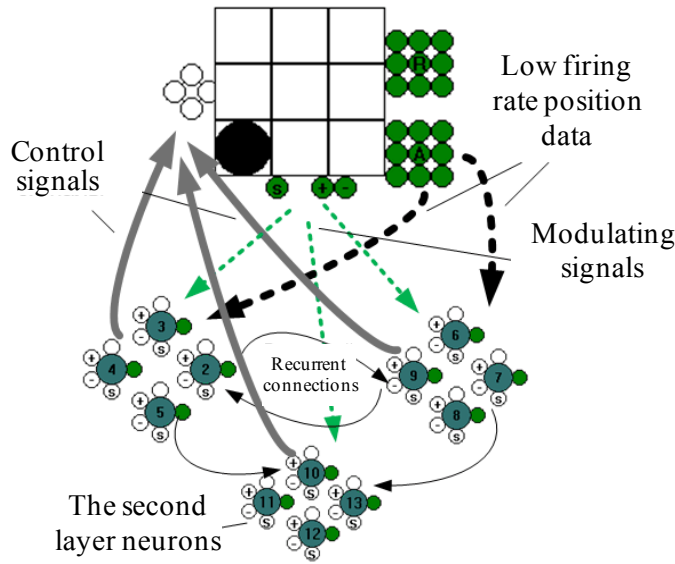

**Fig. 45.** Two-layer 12 neurons spiking network controller with recurrent connections controls the virtual agent.

### 4.4.3. Controlling the agent in a larger virtual grid environment

The exploration-exploitation dilemma [54] is especcially prominent in larger sensory-action spaces. Already in the 5x5 grid the number of possible positions of the agent and a resource is 600. The exploration level is controlled by the stochasticity of the neurons. If the stochasticity is small the network can not explore the environment and therefore develops a suboptimal policy – to always stay in place. In this case the agent never receives a negative reinforcement from the bumps or from the ambigious controls. If the stochasticity is high enough the network finds a reasonable policy of collecting resources: http://www.youtube.com/watch?v=HQy-I3lTp4M.

If the grid's dimensions are $10 \times 10$ the number of possible sensory configurations is 9900. The network needs much more time to explore all possibilities. During the exploration the agent frequently bumps into the walls and always ends up choosing a policy of standing in place. In order to train the network we used different learning coefficients $\gamma$ for the positive and negative reinforcement in (59). In particular, we used $\gamma^+ = 0.005$ for a positive reinforcement spike and a very small negative reinforcement coefficient $\gamma^- = 0.01\gamma^+$.

The experimental setup with 10x10 grid is shown in Fig. 46. The network consists of 32 neurons. Each neuron receives input spikes from 10*10*2=200 position sensors. The neurons are split on 4 groups for each direction of movement. Plots of the received reward and punishment during the training are shown in Fig. 46 (right). Initially the agent's hunger increases the exploratory behavior and the network receives a lot of negative reinforcement that can be seen on the punishment curve in grey as a large peak. After the initial exploration the network starts to improve the policy. During that the amount of the positive reinforcement increases (black curve) and the amount of punishment drops.

The main source of punishment comes from ambiguous control signals. The behavior of the agent after the training can be seen here: http://www.youtube.com/watch?v=aEoVvnr7OYk.

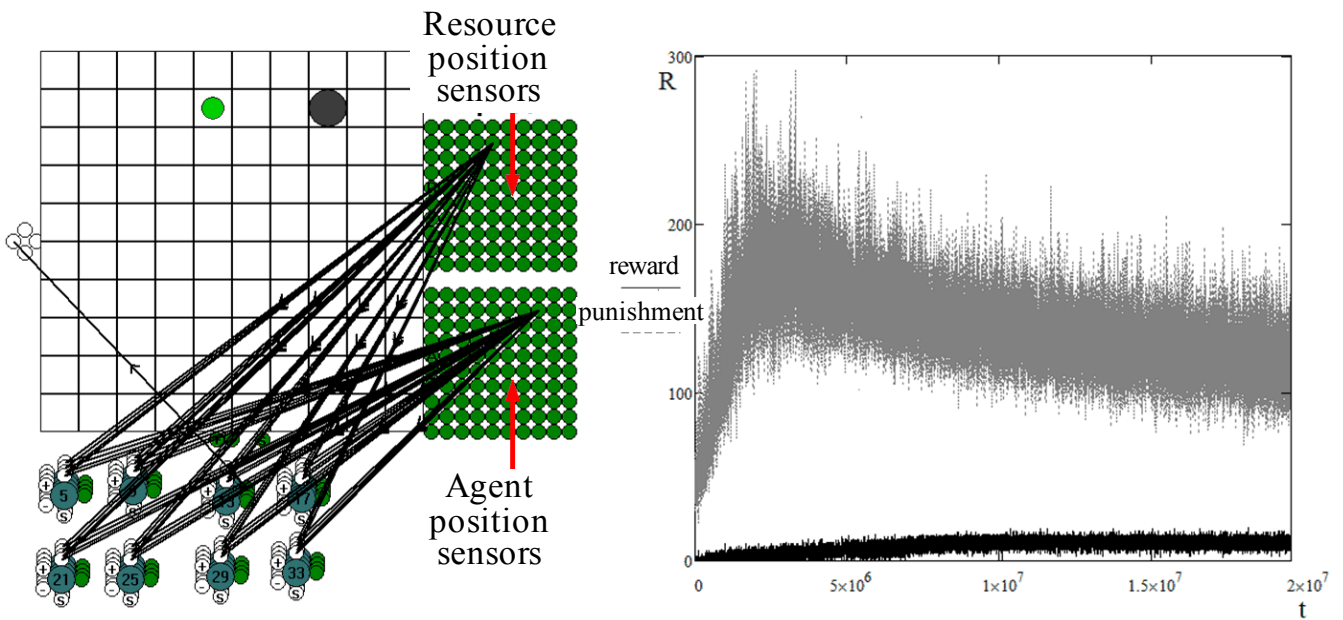

**Fig. 46.** The spiking neural network controls the agent in the virtual 10x10 grid is shown on the left. The accumulated reward and punishment values during the course of training are shown on the right.

### 4.5. A soccer robot model controlled by a spiking neural network

To test further the reinforcement learning algorithm for spiking networks we implemented a dynamical model of a robotic soccer player. This agent is more difficult to control since its state is described by several continuous values: a position, an angle, linear and angular speed of the robot and the ball.

The model of the robot was developed during the control system design of real soccer player robots from FIRA, Mirosot league of robosoccer (http://www.fira.net). Robots in Mirosot league are 7.5 cm by 7.5 cm in size (Fig. 47). The detailed description of the robots of the Russian team participating in FIRA RoboWorld Cup 2006 can be found in [91]. The usage of the classic neural networks with backpropagation and genetic algorithms to implement a low level control of the robots is described in [92].

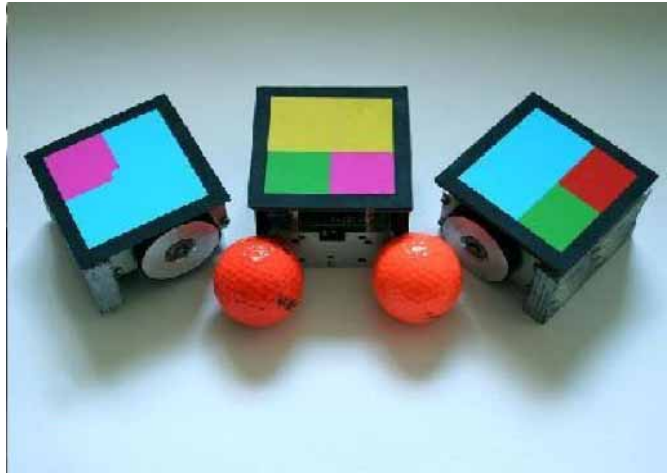

**Fig. 47.** The Mirosot robots of the Russian team "Moscow Pioneers".

A robot model in a small virtual field controlled by a spiking neural network is shown in Fig. 48. The overall scheme of interactions between the network and the environment is similar to the one shown in Fig. 40. The virtual environment has sensory and modulatory outputs and control inputs. Robot's sensors send out spiking streams with information about the current wheel velocities, the robot's angle, the direction to the ball and the proximity to the walls. The continuous values of angles and velocities are discretized and each discretization interval is assigned to a separate neuron. For example, the robot's angle $2\pi$ is split onto 16 intervals and at each time step there is only one active neuron (a neuron responsible for $[-\pi/16; \pi/16)$ interval is active if the angle lies in this interval etc.). Wheels' velocities are constrained to 1m/s and are discretized with a 0.2 m/s step. The wall detectors become active if the robot touches a corresponding wall.

The robot is controlled by four actuators: +L, -L, +R, -R. A spike received on an actuator slightly increases or decreases a corresponding value of a desired wheel velocity $V_L^{need}$ or $V_R^{need}$.

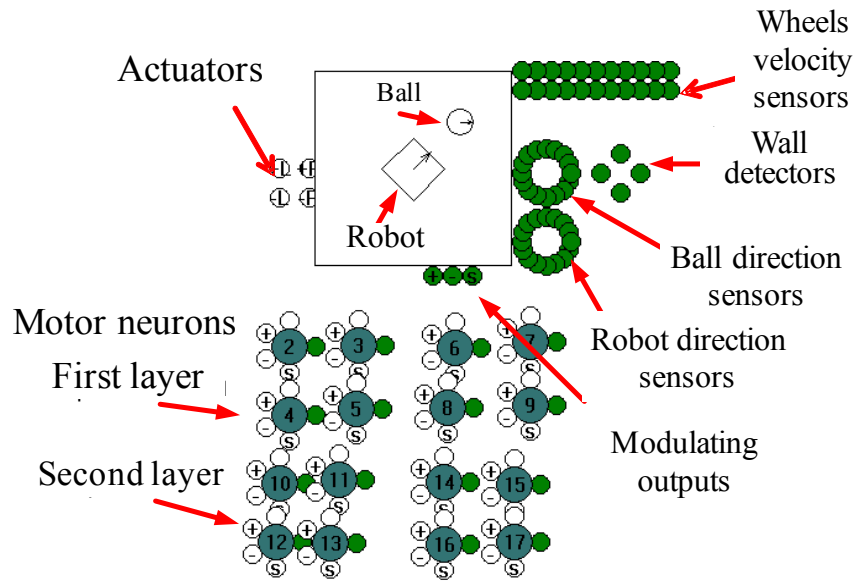

**Fig. 48.** A soccer robot model controlled by a spiking neural network consisting of 16 neurons.

The task of the network is to "kick" the stationary ball with the robot's body. The training of the network was split into iterations. At the beginning of each iteration the robot's and ball's positions were randomly initialized. After that we ran the physical simulation. During the simulation the robot received control signals from the network. If the robot touched the ball, the network received a positive reinforcement spike. If the iteration timed out without success, the network received a negative reinforcement spike. The intensity of the stimulating signal was raised closer to the end of each training iteration forcing the robot to kick the ball as fast as possible.

We tested out several network architectures of the spiking controller: one-layer, two-layer, three-layer where only the last layer neurons were connected to the robot controls. The accumulated sum of the total reinforcement received so far during the training is shown in Fig. 49 for each control architecture. We also plotted the accumulated sum of reinforcement for the network with disabled learning. This network makes the robot perform only exploratory motions and its accumulated reinforcement value serve as a good baseline. In this experiment we failed to see benefits of the multilayer spiking networks. The best results were achieved with a single-layer network. We speculate that it might be related to many local minima in the multilayer networks due to nonlinearities leading to suboptimal policies. The video with the resulting control strategy is shown here: http://www.youtube.com/watch?v=JxINB6n4Rbw.

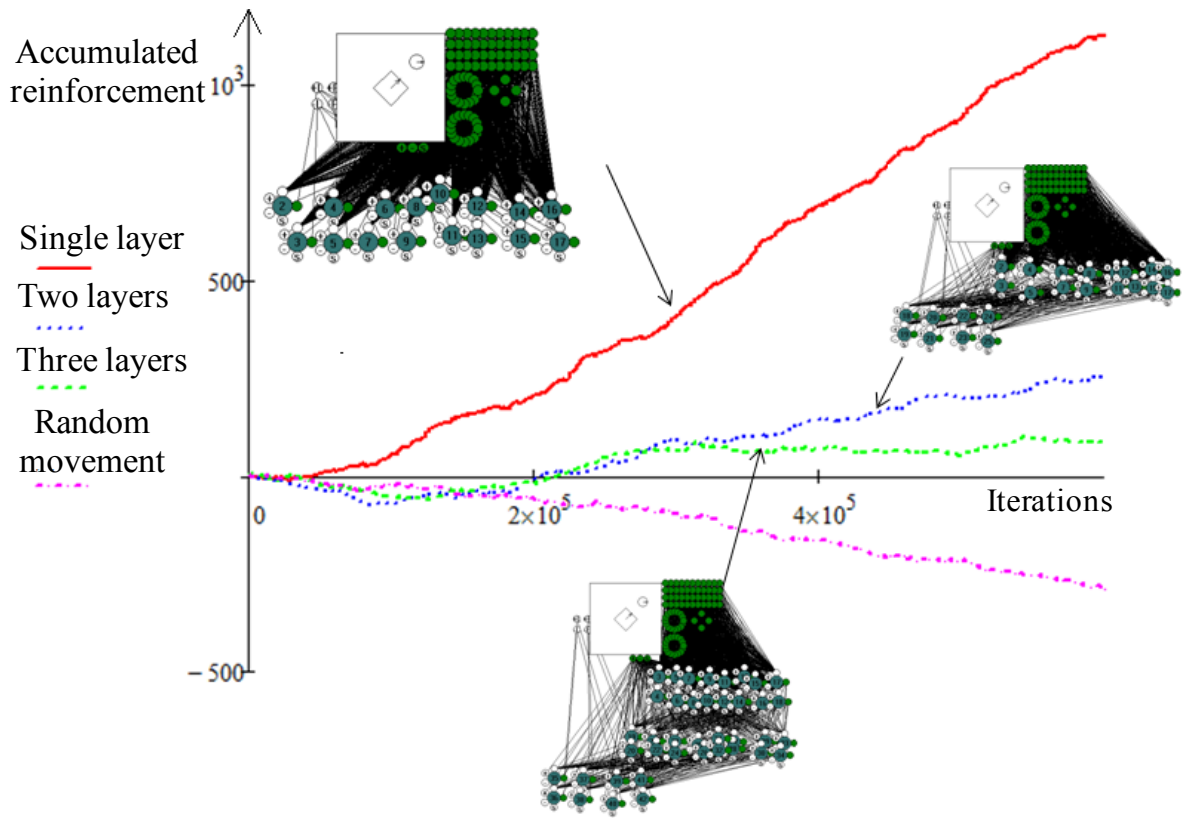

**Fig. 49.** Accumulated reinforcement values during the training are shown for different control architectures and for the network with disabled learning (random movement behavior).

### 4.6. The network adaptation to the changes in the environment

Notice that neural networks that learned to control the agent in the virtual grid and the robot model are practically the same. They have the same number of neurons and the same number of outputs (four). However, the dynamics of the environment and the spatiotemporal structure of sensory signals was quite different. Also the actuation was implemented differently: the agent received direct displacement commands while the robot received velocity increments. The ability of the network to control such different objects is due to the generality of reinforcement learning algorithms that do not use any a priori information about the controlled system.

The goal of this chapter is to show that the network is able to relearn to control a different object online without changing the network's metaparameters such as neurons constants, the learning coefficient or the network architecture. Changes can happen in the external environment or in the properties of the controlled object itself, for example, after a break down of sensors or actuators. The really radical external change is to change the whole external environment and the object to another one (Fig. 50).

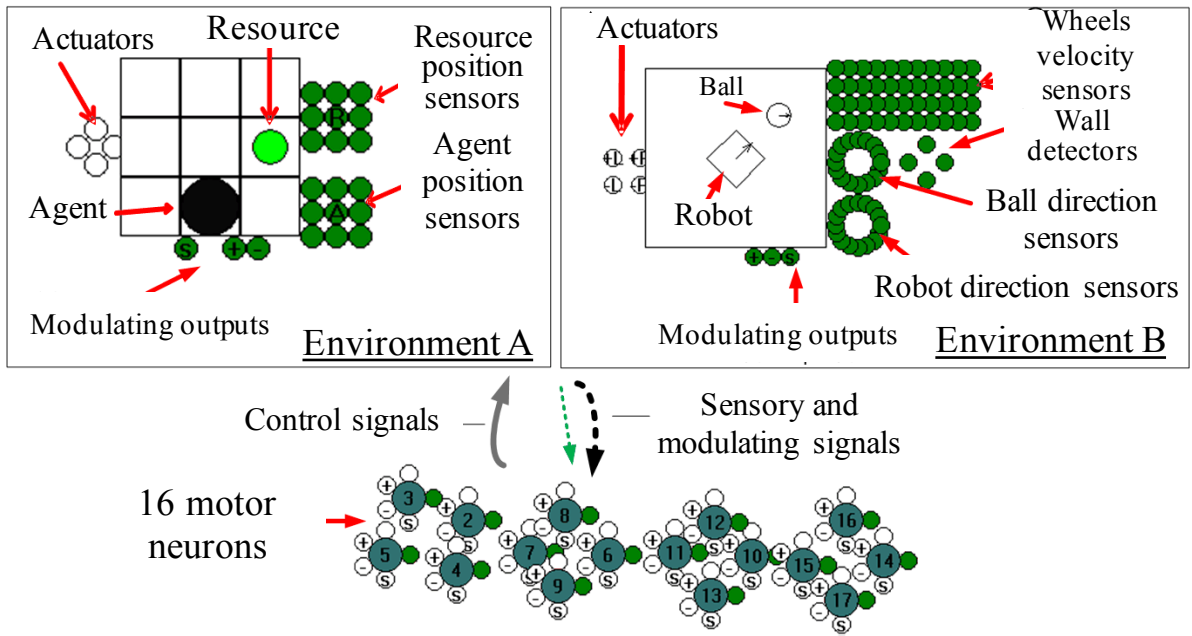

**Fig. 50.** A network consisting of 16 neurons is able to control the agent on the discrete grid and the robot model after the training without changing the metaparameters and architecture of the network.

We conducted an experiment where the network first learns to control the agent on the discrete grid. The network consisted of 16 neurons. The plots of the reward and punishment during the training are shown in Fig. 51 (left). The accumulated reinforcement is shown in Fig. 51 (right). At iterations shown with arrows "A" and "B" we emulate break downs in the agent's actuators. At time "A" an actuator that is responsible for the "left" movement started failing 40% of the time without actually moving the agent to the left. Also the agent received the punishment spike when the failure happened (similar to effects of an injured limb). The average amount of the negative reinforcement grew but the network learned a new policy minimizing the "left" movement. At time "B" an actuator "up" started to move the agent two cells up instead of one. Such change increased the amount of negative reinforcement because the agent started to bump into the top wall a lot. The amount of positive reinforcement also dropped since the old policy did not always work well. For example, if the resource is one cell up from the agent, the network had to go down first and then jump two cells up instead of just moving up. However, the network soon learned a new policy that took into account the effects of the break downs "A" and "B" (http://www.youtube.com/watch?v=SxWBFufStsU). At the moment "C" we changed the virtual agent on the grid to the robot soccer player model by reconnecting the network to the another set of sensors and actuators. The soccer robot environment has quite different properties. The whole environment is slower since the robot needs time to reach the ball and control signals do not change the robot state instantaneously because of inertia. The dimensionality of inputs grew 5 times. We left the learned weights of synapses and initialized new synapses with zeroes. However, we can see from the reinforcement plots in Fig. 51 that network relearned to control the new object.

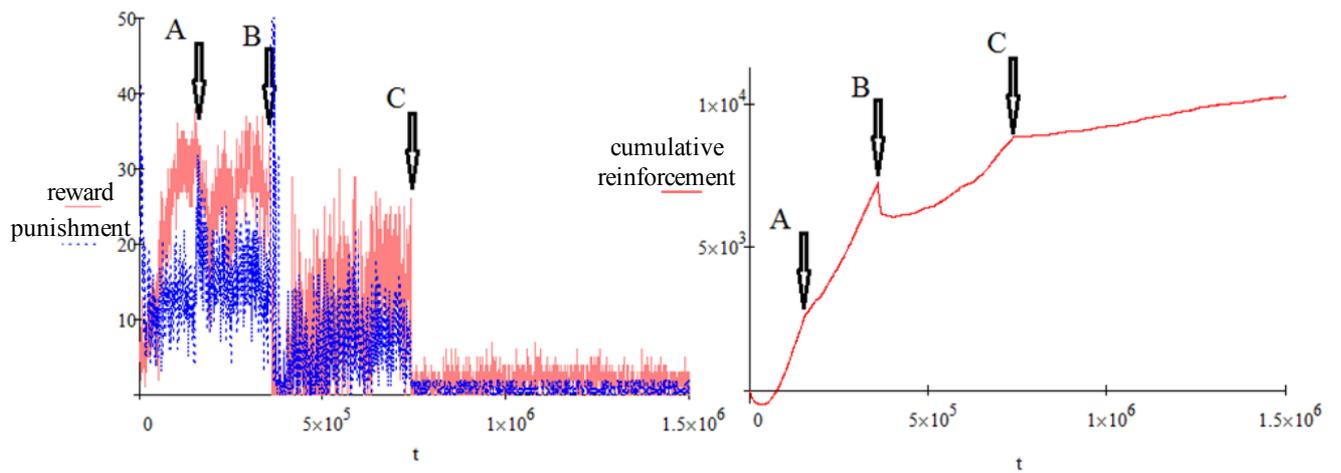

**Fig. 51.** Reward and punishment (left) and cumulative reinforcement (right) during the training in changing external environment. The network consists of 16 neurons (Fig. 50). First, network learned to control the agent in the discrete grid. Arrows show the changes in the environment: A and B indicate the break downs of agent's actuators, C indicates the change of the controlled object from the virtual agent to the soccer player robot model.

# Conclusion

The main purpose of this thesis was to develop a general mathematical framework for spiking neurons learning. We developed a generalized description of the stochastic spiking neuron based on the analysis of several existing spiking neuron models. It was proposed to formalize the spiking neuron learning as a set of tasks of information-theoretic cost functions optimization. We considered several supervised, unsupervised and reinforcement learning tasks and derived learning rules for the generic spiking neuron model for each task.

We developed a new spiking neuron model - Spike Multi Response Model (SMRM) with the stochastic threshold that has a set of alpha-functions on each input channel. The set of alpha functions allows the neuron to flexibly adapt its response to the spatiotemporal structure of a multidimensional spike input. We derived learning rules for this model and tested it on a set of simple tasks. All experiments were performed using a custom software developed by us.

Supervised learning for the generic spiking neuron was formalized as a task of the surprisal minimization. The derived learning rules were tested on a set of simple spatiotemporal patterns detection tasks. After training the neuron was able to detect a particular input pattern in a noisy stream of input spikes by using the input pattern's temporal structure. We developed a method of choosing alpha functions parameters that maximize the speed and robustness of supervised learning.

We developed a spiking autoassociative memory network in order to show another application of the derived supervised learning rules. The memory network is able to store and recall long multidimensional event patterns. We tested the network on the task of learning the drawing process on the $8 \times 15$ virtual canvas. The trained network was able to predict and recall movements of the virtual pen based on few initial pixels.

We described a particular example of an unsupervised learning task – a task of increasing the robustness of the pattern generation by a generalized stochastic neuron. The task was formalized as a minimization of the entropy of the neuron's output. We developed a new unsupervised online learning algorithm that allows the neuron to decrease the entropy and increase the robustness of its output. The experiments conducted with the SMRM neuron showed that after the training the most likely output pattern is generated robustly and less likely patterns are not generated anymore. We developed a composite learning method that combines supervised pretraining with additional unsupervised learning. We showed on a simple task that this algorithm decreases the number of iterations needed to learn a pattern.

We formulated reinforcement learning of a spiking neuron in the information-theoretic framework. We showed that the parameters' gradient can be represented as a convolution of the gradient of the surprisal with the exponential kernel modulated by received reinforcement. Positive reinforcement leads to the minimization of the surprisal of recently generated output patterns while the negative reinforcement maximizes it.

We derived reinforcement learning rules for the SMRM neuron and tested several control spiking networks on a set of simple control tasks. We trained the same spiking network to control a virtual agent in a discrete grid and a soccer player robot model. We investigated several network architectures and showed that a simple recurrent spiking network is able to control the agent in the lack of spatial information by using its recurrent activity. We showed that a simple network is able to relearn to control different objects and cope with other changes in the controlled object such as the actuators break downs.

# Appendix1. Custom software for spiking neurons modelling

Currently there exist multiple software packages to simulate neural networks. In order to model classic float-output neural networks, several extensions to the popular mathematical frameworks such as Matlab [93], Maple [94], Mathcad [95] can be used. These extensions are not applicable to spiking neuron networks due to a higher complexity of spiking network dynamics and the usage of binary events instead of float values for internal communication. In float-output networks every element generates a signal on every step and signal propagation is quite simple. Signal propagation in spiking networks can be complex since neuron's activity is usually sparse in time and it is suboptimal to transmit information about the lack of activity at every step as opposed to transmitting the spikes only. Also, spiking networks frequently use signal propagation delays that are not typically used in float-output networks. It is also a difficult task to implement the spike-timing dependent plasticity because one needs to store all times of events for certain duration. Such properties make it hard to reuse vectorized computation abilities of the traditional packages. There is an extension for Matlab CSIM [96] that does allow to model spiking networks. However, all the computations are done in a separate C++ based library which makes it hard to customize the model.

There are software packages that have addressed most of the issues of modelling biologically plausible neural networks. The most popular are Neuron [97], Genesis [98] and Nest [99]. These systems have a high degree of detalization of neuron models allowing developers to simulate dynamics of ion channels, spatial arrangements of dendrites and axons etc. However, these systems are too complex for the purpose of this thesis since we use a very simple neuron models without much concern for their biological plausibility. The main focus of this thesis is to investigate into various complex learning rules. Usually the single available learning rule is Hebb's rule and various forms of STDP. These learning rules are bidirectional since a particular synapse needs to access only pre- and post-synaptic neuron in order to change its weight. Certain optimizations can be done to implement such rules more efficiently. For example a python package Brian [100] implements STDP using $O(n+m)$ computations instead of $O(n \cdot m)$ where $n$, $m$ are dimensions of the connectivity matrix between two groups of neurons. However, in this thesis we mainly interested in three-directional learning rules that depend on the state of pre-, post- and some third signal such as teacher input or reinforcement. In this case the Brian optimization is not applicable and makes it hard to experiment with learning rules. Because of the mentioned issues we developed a custom software for the spiking neurons modelling written in C++.

A component of a spiking neural network is a black box that has arbitrary number of inputs and outputs. In the developed software we do not restrict a number of factors that can participate in learning. To achieve that we introduced a concept of a type of the input channel. Some inputs are sensory, while others can trigger learning or modulate the neuron's behavior (e.g. in the reinforcement learning) (Fig. 52). Also there are no restrictions on the implementation of synapses: the SMRM model uses 3 float valued weights per synapse and a teaching input does not have any weight. Here are some examples of spiking network components: neurons, layers of neurons, delay elements, spiking data sources, virtual environments. A developer can customize and create new components, while the core of the system makes sure that spikes are propagated as fast as possible.

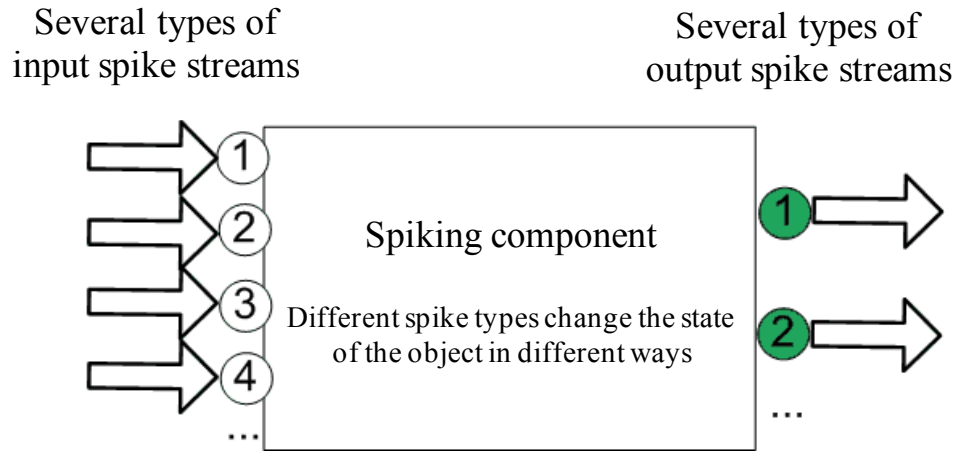

**Fig. 52.** A spiking component can process several types of spike streams. Different spike streams can have different meaning such as sensory inputs or reinforcement signal. In general, outputs also can have different meaning and types.

Spiking neurons are dynamic entities and it is quite useful to see what the network components are doing in real time. Visualization is especially important during the experiments with virtual environments and in the task of learning the patterns by the spiking memory network. Additionally, we would like to have an interactive user interface for tuning neurons parameters. We developed a custom graphical user interface for all components of the network (Fig. 53, Fig. 55). An interactive user interface is a rare feature in the spiking neurons software packages (mainly they allow only to build plots and spike rasters). Using this interface, a developer is able to perform all the mentioned above activities. This includes dynamic connection and editing of the components using the "drag-and-drop" interface and plotting in real time component's state values.

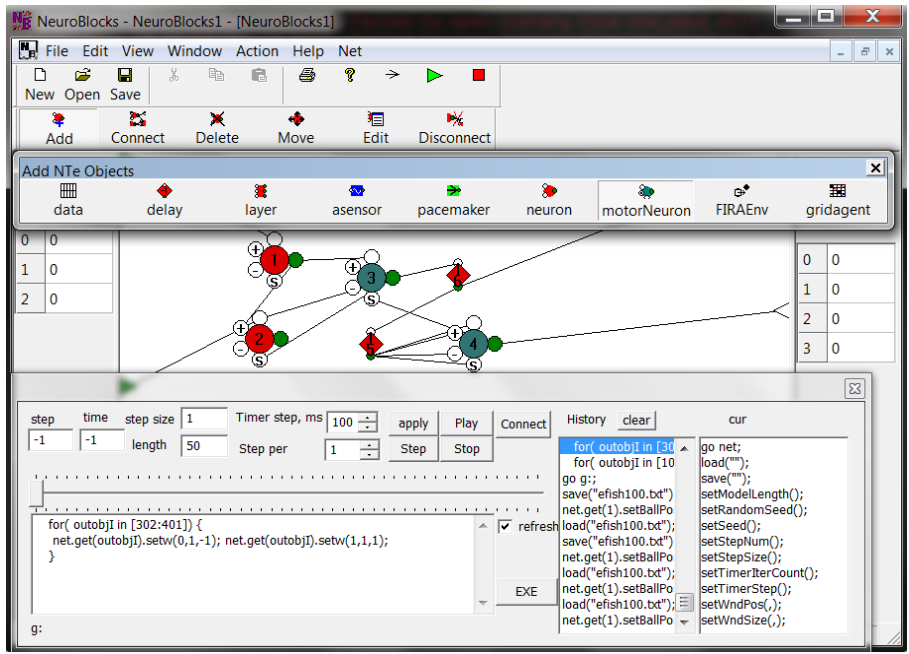

**Fig. 53.** Graphical user interface for the spiking modelling software. Top: a toolbar with basic operations: create different components, destroy, connect etc. Center: a network with 4 neurons and 2 delay elements. Bottom: a dialog for working with scripting language. An example of a script for setting the weights is shown in the edit window.

Sometimes it is hard to experiment with network architectures using only GUI. The network might have a repetitive structure, also some connectivity patterns are better described algorithmically. For such purposes we extended the software with a custom scripting language and read-eval-print loop (REPL) environment based on Python and C++. Notice that almost every spiking software package uses a custom description language (Neuron, Genesis, Nest) or a generic scripting language (CSIM, Brian). The network components are built using object-oriented design patterns which was reflected in the created language. In order to simplify tuning parameters in hierarchical sets of objects, we introduced a command "*go*". This command changes the current scope of the execution allowing the subsequent commands to be interpreted in the internal scope of a specific object. For example, initially the commands are executed in the global scope like in Python REPL. The scripting environment is already initialized with a common global objects. In particular, a "*net*" object represents a currently loaded spiking network. A command "*go net;*" will change the scope of the interpreter and REPL to the internal scope of this object so that the user can easily access its parameters. For example, instead of calling "*net.method()*" the user can just call "*method()*". Subcomponents of the component are accessible with the function "*get(i);*", where *i* is the subcomponent's index. For example, if the network consists of several neurons "*get(1).setSynapseCapacity(3);*" will set the number of alpha-functions of the first neuron to be equal to 3. A command "*go get(1);*" will change the scope to the scope of the first neuron so that the user can directly call "*setSynapseCapacity*(3);". The user might go further down the hierarchy of objects (neurons input streams, synapses, its visual representation etc.)

Every object of the network can be described using 3 independent components:
1) Main spike processing logic
2) Interactive GUI components
3) Scripting language representation components.

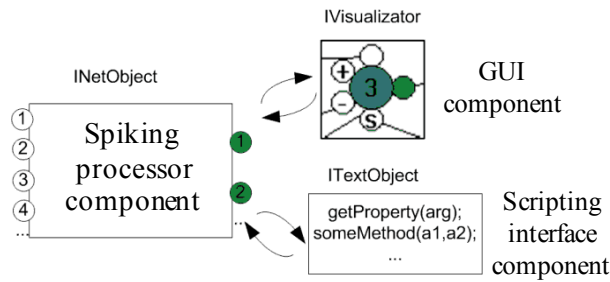

**Fig. 54.** Every component of the network is built out of 3 components: spiking processing logic, GUI and scripting components.

During the design of the software we took inspiration from the COM object model. Every component is placed into a dynamic library which is loaded during the startup. The component has a set of supported interfaces that can be queried. The components communicate exclusively via the registered public interfaces. In order to add a component, a developer needs to create a dynamic library and to expose the required interfaces. It is possible to create only the main logic component and develop GUI and scripting parts when needed. There is a set of C++ base classes and library functions already available for the developer.

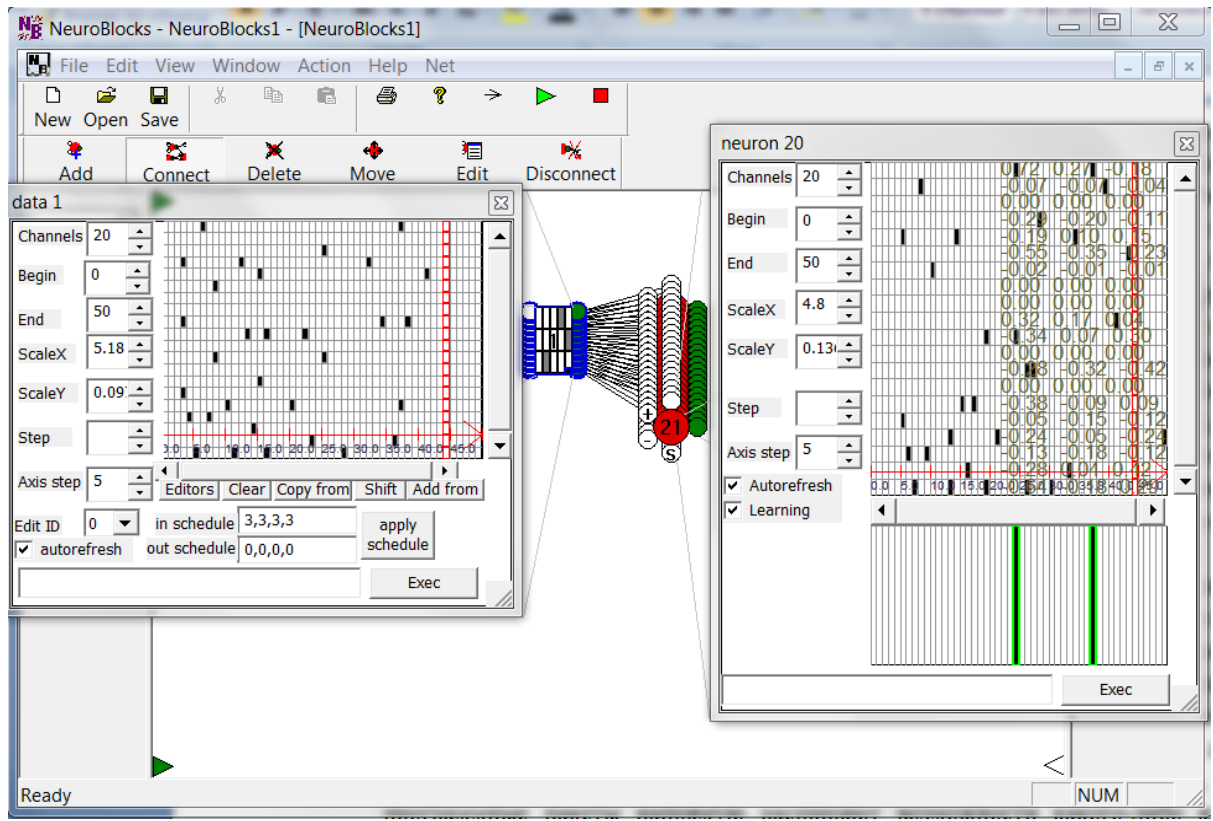

**Fig. 55.** An example of the developer environment during the pattern detection training. There is a loaded input spiking pattern on the left that can be dynamically edited. The input spikes from the pattern arrive to the set of neurons in the center. A dialog that displays a history of activity of a particular neuron is shown on the right.

**References:**

1. **Nicholls J.G., Martin A. R., Wallace B.G., Kuffler S.W.** From Neuron to Brain. Sinauer Associates, 1992 - Medical - 807 pages.

2. **RiekeF. , WarlandD. , RuyterR. vanSteveninck, BialekW.**Spikes: ExploringtheNeuralCode //ComputationalNeurosciencesseries – MIT Press, 1997 – 416 p.

3. **McCulloch W. S., Pitts W.** A logical calculus of ideas immanent in nervous activity //Bulletin of Mathematical Biophysics -1943. –Vol. 5 - P. 115-133.

4. **Rosenblatt F.** The Perceptron: A Probabilistic Model for Information Storage and Organization in the Brain //Psychological Review.-1958. -Vol. 65 - No 6 -P. 386-408.

5. **Adrian E. D.** The impulses produced by sensory nerve endings //J. Physiology - Lond. 1926. -Vol. 61 -P.49-72.

6. **Kandel E. C.; Schwartz J. H.**Principles of Neural Science -New York : Elsevier, 3rd edition, 1991 -1182 p.

7. **Hebb D.O.**The Organization of Behavior - New York : John Wiley & Sons - 1949. – 378 p.

8. **Hopfield J. J.** Neural networks and physical systems with emergent collective computational abilitie //Proceedings of the National Academy of Sciences of the USA,1982 - Vol. 79 - No 8 -P. 2554-2558.

9. **Osowski S.** Нейронные сети для обработки информации [Neural networks for data processing] - М. : Финансы и статистика, 2002. -344 с.

10. **Rumelhart D.E., Hinton G.E., Williams R.J.** Learning Internal Representations by Error Propagation //Parallel Distributed Processing. Parallel Distributed Processing -Cambridge, MA, MIT Press, 1986.- Vol. 1. -P. 318-362.

11. **Chauvin Y.; Rumelhart D.E.**Backpropagation: Theory, Architectures, and Applications - New Jersey Hove, UK: Hillsdale, 1995. -576 p.

12. **Neaupane K. ; Achet S.** Some applications of a backpropagation neural network in geo-engineering//Environmental Geology - 2003. -Vol. 45 - No 4 -P. 567-575.

13. **Astion M.L., Wilding P.** The application of backpropagation neural networks to problems in pathology and laboratory medicine //Arch. Pathol.Lab.Med. -1992. -Vol. 116, 10, P. 995-1001.

14. **Baestaens D.-E., Van Den Bergh W.M.** Neural Network Solutions for Trading in Financial Markets . - Financial Times, 1994 - Business & Economics - 245 p.

15. **Mozer M. C., Hillsdale N. J.** A focused backpropagation algorithm for temporal pattern recognition. //Backpropagation, Lawrence Erlbaum Associates -1995. - P. 137-169.

16. **O'Reilly R. C., MunakataY..**Computational Explorations in Cognitive Neuroscience: Understanding the Mind by Simulating the Brain -Cambridge, MA: MIT Press, 2000 -512 p.

17. **Carr C. E., Konishi M.** A circuit for detection of interaural time differences in the brain stem of the barn owl//J. Neuroscience –1990. -Vol.70 – No 10 -P. 3227-3246.

18. **Bell C.C. , Han V. , Sugawara Y., Grant K.** Synaptic plasticity in a cerebellum-like structure depends on temporal order//Nature -1997. -Vol. 387 -P. 278 - 281.

19. **Thorpe S., Fize D., Marlot C.** Speed of processing in the human visual system//Nature –1996. -Vol.381 – No 6582. -P. 520 – 522.

20. **Tovee M. J., Rolls E. T.** Information encoding in short firing rate epochs by single neurons in the primate temporal visual cortex//Visual Cognition. –1995. -Vol. 2. – No 1. -P. 35 – 58.

21. **Johansson R.S., Birznieks I.** First spikes in ensembles of human tactile afferents code complex spatial fingertip events//Nature Neuroscience. – 2004. -Vol. 7. -P. 170 – 177.

22. **Mehta M. R., Lee A. K.,Wilson M. A.** Role of experience and oscillations in transforming a rate code into a temporal code // Nature. –2002. -Vol. 417. -P. 741-746.

23. **Bi G.Q., Poo M.M.** Synaptic modifications in cultured hippocampal neurons: dependence on spike timing, synaptic strength, and postsynaptic cell type //Journal of Neuroscience. –1998. -Vol.18. – No 24. -P. 10464-10472.

24. **Maas W.** Networks of spiking neurons: the third generation of neural network models //Transactions of the Society for Computer Simulation International - Special issue: simulation methodology in transportation systems archive. –1997. -Vol. 14. – No. 4. -P. 1659-1671.

25. **Stein R. B.** Some models of neuronal variability //Biophys. J. – 1967. -Vol. 7. No 1. -P. 37–68.

26. **Feng J.** Is the integrate-and-fire model good enough - a review //Neural Networks. –2001. -Vol.14. – No 6. -P. 955-975.

27. **Izhikevich E. M.** Dynamical Systems in Neuroscience: The Geometry of Excitability and Bursting -The MIT Press, 2007 - 457 p.

28. SpikeNet Technology[Electronic resource] - 2006. – Mode access: http://www.spikenet-technology.com.

29. **Hopfield J. J.** Pattern recognition computation using action potential timing for stimulus representation //Nature -1995. -Vol. 376. -P. 33 – 36.

30. **O'Keefe J.** Hippocampus, theta, and spatial memory //Curr. Opin. Neurobiol –1993. -Vol. 3. -P.917-924.

31. **Gerstner W., Kistler W.M.** Spiking Neuron Models: Single Neurons, Populations, Plasticity - Cambridge University Press, 2002 - 480 p.

32. **Melamed O., Gerstner W., Maass W., Tsodyks M., Markram H.** Coding and learning of behavioral sequences //Trends in Neurosciences –2004.-Vol. 27. – No 1. -P. 11-14.

33. **Saggie K., Keinan A., Ruppin E.** Solving a delayed response task with spiking and McCulloch-Pitts agents //Advances in Artificial Life: 7th European Conference, ECAL 2003 - Dortmund, Germany. 2003. -P. 199-208.

34. **Komartsova L.G., Maksimov, A.B.,** Neurocomputers: A Study Guide for University Students (2nd ed.) [In Russian], Moscow: The Bauman MSTU Press, 2004, p. 400.

35. **Galushkin A.I.** Neural Networks Theory. Springer-Verlag Berlin Heidelberg, 2007. -396 c.

36. **Paquot Y., Duport F. Dambre J., Schrauwen B., Haelterman M., Massar S.** Artificial intelligence at light speed : toward optoelectronic reservoir computing //Belgian Physical Society Magazine. –2001. -Vol. 3. -P. 15-22.

37. **Thorpe S. , Delorme A., Rullen R.** Spike-based strategies for rapid processing //Neural Networks. - 2001. – Vol. 14. -P.715-725.

38. **Ponulak F.**ReSuMe - New supervised learning method for Spiking Neural Networks[Electronic resource]. -Poznan University of Technology, Institute of Control and Information Engineering , 2005. –Mode access: http://d1.cie.put.poznan.pl/~fp/.

39. **Legenstein R., Markram H., Maass W.** Input prediction and autonomous movement analysis in recurrent circuits of spiking neurons //Rev Neurosci. -2003. -Vol. 14. - No 1-2. - P.5-19.

40. **Paolo E. Di.** Spike-Timing Dependent Plasticity for Evolved Robots //Adaptive Behavior. –2002. -Vol. 10. - No 3. -P.73-95.

41. **Damper R. I.,  French R. L. B.,  Scutt T. W.** ARBIB: an Autonomous Robot Based on Inspirations from Biology //Robotics and Autonomous Systems -1998. -Vol. 31. – No. 4. -P.247-274.

42. **Wiles J., Ball D., Heath S., Nolan C., Stratton P.** Spike-time robotics: a rapid response circuit for a robot that seeks temporally varying stimuli //Australian Journal of Intelligent Information Processing Systems. -2010. -P.1-10.

43. **Floreano D., Zufferey J.-C., Mattiussi C.** Evolving Spiking Neurons from Wheels to Wings //Dynamic Systems Approach for Embodiment and Sociality. -2003. -Vol. 6. -P. 65-70.

44. **Nolfi S., Floreano D.** Synthesis of Autonomous Robots Through Evolution //Trends in Cognitive Sciences. -2002. -Vol. 6. - No 1. -P.31-37.

45. **Florian R. V.** Spiking Neural Controllers for Pushing Objects Around //Proceedings of the Ninth International Conference on the Simulation of Adaptive Behavior (SAB'06) -2006. -Vol. 4095 -P.570-581.

46. **Castillo P.A., Rivas V., Merelo J.J., Gonzalez J., Prieto A., Romero G.** G-Prop-II: Global Optimization of Multilayer Perceptrons using GAs //CEC 99. Proceedings of the 1999 Congress on Evolutionary Computation -1999. – P. 149-163.

47. **Komartsova L.G.**Модифицированный алгоритм обучения многослойного персептрона [Modified algorithm for multilayer perceptron training] // "Математика. Компьютер. Образование". Сб. трудов XII международной конференции. Под общей редакцией Г.Ю. Ризниченко [Proceedings of the 12-th conference «Mathematics. Computer. Education»]- Ижевск: Научно-издательский центр "Регулярная и хаотическая динамика", 2005. - Т. 2. -С. 427-432.

48. **Bohte S.M., Kok J.N., Poutré J.A.L.**SpikeProp: backpropagationfornetworksofspikingneurons // ProceedingsofESANN. -2000. -P.419-424.

49. **Pfister J.P., Toyoizumi T., Barber D., Gerstner W.** Optimal Spike-Timing Dependent Plasticity for Precise Action Potential Firing in Supervised Learning //Neural computation -2006. -Vol. 18. - No 6 -P.1318 – 1348.

50. **Bohte S.M., Mozer M.C.** A computational theory of spike-timing dependent plasticity: achieving robust neural responses via conditional entropy minimization //SEN-E0505. –2005. -P.1-25.

51. **Toyoizumi T., Pfister J.-P. , Aihara K. , Gerstner W.** Optimality Model of Unsupervised Spike-Timing Dependent Plasticity: Synaptic Memory and Weight Distribution //Neural Computation. -2007. -Vol. 19. - No 3. -P. 639-671.

52. **Markowitz D.A., Collman F., Brody C.D., Hopfield J.J., Tank D.W.** Rate-specific synchrony: using noisy oscillations to detect equally active neurons //Proc. Natl. Acad. Sci. -2008. – Vol.105. – No 24. -P.8422-8427.

53. **Kingman J.F.** Poisson Processes. – Clarendon Press, Dec 17, 1992 - Mathematics - 112 p

54. **Sutton R.S., Barto A.G.** Reinforcement Learning: An Introduction. - Cambridge: MIT Press, 1998. - 432 p.

55. **Webster R. A.**Neurotransmitters, Drugs and Brain Function. -  John Wiley and Sons, 2002. - 534 p.

56. **Deutch A.Y., BeanA. J.** Colocalization in Dopamine Neurons //Psychopharmacology: The Fourth Generation of Progress. - New York, Raven Press, 1995.- P. 205-214

57. **Holmes P. V., Crawely J. Q. N.**Coexisting Neurotransmitters in Central Noradrenergic Neurons. // Psychopharmacology: The Fourth Generation of Progress. - New York, Raven Press; 1995. - P. 347-353

58. **Porr B., Worgotter F.** Isotropic sequence order learning // Neural Computation. - 2003. -Vol. 15. – No. 4. -P. 831-864.

59. **Wermter S., Christo P.** Temporal Sequence Detection with Spiking Neurons: Towards Recognizing Robot Language Instructions // Connection Science. -2006. -Vol. 18. - No 1. -P.1-22.

60. **Perkel D. H., Feldman M. W.** Neurotransmitter release statistics: Moment estimates for inhomogeneous Bernoulli trials. Berlin //J. Math. Biol.–1979.– Vol. 7. – No 1. – P. 31-40.

61. **Dunin-Barkovskii V. L., Osovets N. B.** Neural network with formed dynamics of activity //Radiophysics and Quantum Electronics. - 1994. -Vol. 37. - No 9. -P. 687-693.

62. **Szatmáry B., Izhikevich E. M.** Spike-Timing Theory of Working Memory // PLoS Comput. Biol. – 2010 - Vol. 6 - No 8.

63. **Wills S. A.** Computation with Spiking Neurons [Electronic resource]. PhD Disertation. – 2004. – Access mode: http://ecs.victoria.ac.nz/twiki/pub/Courses/COMP421_2010T1/Readings/SebWillsPhD-chapter3.pdf.

64. **Hopfield J. J. , Brody C. D.** Sequence reproduction, single trial learning, and mimicry based on a mammalian-like distributed code for time. [Electronic resource]. – 2010. – Access mode: http://arxiv.org/abs/0910.2660.

65. **Baudry M., Davis J. L., Thompson R. F.** Advances in Synaptic Plasticity - N.Y.: MIT Press, 1999. - 335 p.

66. **Bi G., M. Poo.** Synaptic modification of correlated activity: Hebb's postulate revisited //Ann. Rev. Neuroscience –2001. - Vol. 24. -P.139-166.

67. **Stratonovich R.L.** Теория информации [Information theory] - М.: Сов. Радио, 1975 - 424 с.

68. **Antonelo E. A., Schrauwen B., Stroobandt D.** Mobile Robot Control in the Road Sign Problem using Reservoir Computing Networks //IEEE Int. Conf. on Robotics and Automation (ICRA) –2008.- P.911-916.

69. **Queiroz M. S., Braga A., Berredo R. C.** Reinforcement Learning of a Simple Control Task Using the Spike Response Model //Neurocomputing. –2006. -Vol. 70. – No. 1-3. -P. 14-20.

70. **Lee K., Kwon D.-S.** Synaptic plasticity model of a spiking neural network for reinforcement learning //Neurocomputing. -2008. -Vol. 17. – No 13-15. -P. 3037-3043.

71. **Florian R. V.** A reinforcement learning algorithm for spiking neural networks //SYNASC '05 Proceedings of the Seventh International Symposium on Symbolic and Numeric Algorithms for Scientific Computing. –2005. - P. 299-306.

72. **Burgsteiner H.** Training networks of biological realistic spiking neurons for real-time robot control //Proceedings of the 9th International Conference on Engineering Applications of Neural Networks, Lile, France. –2005. -P. 129-136.

73. **Alnajjar F., Murase K. .** A Simple Aplysia-Like Spiking Neural Network to Generate Adaptive Behavior in Autonomous Robots //Adaptive Behavior. –2008. -Vol. 16. – No 5. -P. 306–324.

74. **Soula H., Alwan A., Beslon G.** Learning at the edge of chaos: Temporal Coupling of Spiking Neurons Controller for Autonomous Robotic //Proceedings of American Association for Artificial Intelligence (AAAI) Spring Symposia on Developmental Robotic, Stanford, USA.  - 2005.

75. **Joshi P., Maass W.** Movement Generation with Circuits of Spiking Neurons  //Neural Computation.  - 2005. -Vol. 17. - No 8. -P. 1715-1738.

76. **Carrillo R., Ros E., Boucheny C., Coenen O. J.-M.D.** A real-time spiking cerebellum model for learning robot control //Biosystems. –2008. -Vol. 94. - No 1-2. -P. 18-27.

77. **Boucheny C., Carrillo R., Ros E., Coenen O. J.-M.D.** Real-Time Spiking Neural Network: An Adaptive Cerebellar Model  //Computational Intelligence and Bioinspired Systems: Lecture Notes in Computer Science. –2005. -Vol. 3512. -P. 136-144.

78. **Manoonpong P., Woegoetter F., Pasemann F.** Biological Inspiration for Mechanical Design and Control of Autonomous Walking Robots: Towards Life-like Robots //The International Journal of Applied Biomedical Engineering (IJABME). –2010. -Vol. 3. -No 1. -P. 1-12

79. **Maass W., Natschlager T., Markram H. .** Real-time computing without stable states: a new framework for neural computation based on perturbations //Neural Computations. – 2002. - Vol. 14. – No 11. -P. 2531-2560.

80. **Baxter J., Weaver L., Bartlett P. L.** Direct gradient-based reinforcement learning: II. Gradient ascent algorithms and experiments // Technical report, Australian National University, Research School of Information Sciences and Engineering. -1999.

81. **Bellman R.** A Markovian Decision Process //Journal of Mathematics and Mechanics. – 1957. - Vol. 6.

82. **Farries M. A., Fairhall A. L.** Reinforcement Learning With Modulated Spike Timing–Dependent Synaptic Plasticity //Neurophysiol. – 2007. -Vol. 98. - No 6. -P. 3648-3665.

83. **Baras D., Meir R.** Reinforcement Learning, Spike Time Dependent Plasticity and the BCM Rule //Neural Computation. – 2007. -Vol. 19. - No 8. -P. 2245-2279.

84. **Levine M.W., Shefner, J.M.** Fundamentals of sensation and perception. - Pacific Grove, CA: Brooks/Cole, 1991. -512 p.

85. **Rejeb L., Guessoum Z., M'Hallah R.** An Adaptive Approach for the Exploration-Exploitation Dilemma forLearning Agents //Multi-Agent Systems and Applications IV. – 2005. -Vol. 3690. - P.316-325.

86. **Bartlett P. L., Baxter, J.** A biologically plausible and locally optimal learning algorithm for spiking neurons[Electronic resource]. –2000. Access Mode: http://arp.anu.edu.au/ftp/papers/jon/brains.pdf.gz.

87. **Legenstein R., Pecevski D., Maass W.** A Learning Theory for Reward-Modulated Spike-Timing-Dependent Plasticity with Application to Biofeedback //PLoS Comput Biol. – 2008. -Vol. 4. - No 10. - e1000180. doi:10.1371/journal.pcbi.1000180

88. **Izhikevich E. M.** Solving the Distal Reward Problem through Linkage of STDP and Dopamine Signaling //Cerebral Cortex. – 2007. -Vol. 17. -P. 2443 - 2452.

89. **Fremaux N., Sprekeler H., Gerstner W.** Functional Requirements for Reward-Modulated Spike-Timing-Dependent Plasticity //The Journal of Neuroscience. – 2010. -Vol. 30. – No. 40. -P.13326-1333.

90. **Redko W.G.** Эволюция, нейронные сети, интеллект: Модели и концепции эволюци-онной кибернетики [Evolution, neural networks, intelligence: Models and concepts of evolutionary cybernetics] -М.: УРСС, 2005 - 224 с.

91. **Pakhomov V., Yelkin E.** Introducing an Another One Mirosot Robot Soccer System // Proceeding of FIRA Robot World Congress 2006. -Dortmund University, 2006. -P. 137-145.

92. **Sinyavskiy O.Y., Kobrin A.I.** Research opportunities of management by movement models of the mobile robot football player with the help of neural net algorithms //Proceeding of FIRA Robot World Congress 2006. -Dortmund University, 2006. -P. 231-240.

93. Matlab 7.11 overview. [Electronic resource]. -Access Mode: http://www.mathworks.com/help/pdf_doc/matlab/getstart.pdf.

94. Maple 14 overview. [Electronic resource]. -Access Mode: http://www.maplesoft.com/view.aspx?SF=53244/0/Maple14UserManua.pdf.

95. Mathcad 15.0 overview. [Electronic resource]. -Access Mode: http://www.ptc.com/WCMS/files/121836/en/6011_Mathcad_15_DS.pdf.

96. CSIM : A Neural Circuit SIMulator. [Electronic resource]. -Access Mode: http://www.lsm.tugraz.at/download/csim-1.1-usermanual.pdf.

97. **Hines M.L., Carnevale N.T.** The NEURON simulation environment //The Handbook of Brain Theory and Neural Networks, 2nd ed., edited by M.A. Arbib. - Cambridge, MA: MIT Press, 2003. -P. 769-773.

98. **Bower J. M., Beeman D., Hucka M.** The GENESIS Simulation System //The Handbook of Brain Theory and Neural Networks, 2nd ed., edited by M.A. Arbib. - Cambridge, MA: MIT Press, 2003. -P. 475-478

99. **Gewaltig M.O., Diesmann M.** NEST //Scholarpedia. – 2007. – Vol. 2. – No. 4.

100. **Goodman D.F.M., Brette R.** The Brian simulator //Frontiers in Neuroscience. -2009. -Vol. 3. - No 2. -P. 192- 197.

**Publications of the thesis materials:**

105. **Kobrin A.I., Sinyavskiy O.Y.** Сочетание процессов обучения и самообучения спайкового нейрона, обеспечивающее минимизацию информационной энтропии [Combination of supervised and unsupervised learning in spiking neurons using the informational entropy minimization] //Вестник МЭИ [Vestnik MPEI]. - 2011.– №. 2 -Р.

106. **Sinyavskiy O.Y., Kobrin A.I.** Generalized stochastic spiking neuron model and extended Spike Response Model in spatial-temporal impulse pattern detection task // Optical Memory & Neural Networks (Information Optics). - Allerton press, 2010. –Vol. 19. - No 4. -P. 300-309

107. **Sinyavskiy O.Y., Kobrin A.I.** Обучение спайкового нейрона с учителем в задаче детектирования пространственно-временного импульсного паттерна [Supervised learning in spiking neurons in the spatial-temporal impulse pattern detection task] // Нейрокомпьютеры: разработка и применение [Neurocomputers].: - М. Радиотехника, 2010. - №8. - С. 69-76.

108. **Sinyavskiy O.Y.** Autoassociative spatial-temporal pattern memory based on stochastic spiking neurons // Annals of DAAAM International Symposium. – 2010. -P. 121-122
109. **Sinyavskiy O.Y., Kobrin A.I.** Construction of adaptive robot control system and robot sensor information processing using spiking neural networks //Proceedings Taiwan-Russian Bilateral Symposium on Problems in Advanced Mechanics. -Moscow State University, 2010. -P. 218-227.
110. **Sinyavskiy O.Y., Kobrin A.I.** Использование метода обучения с подкреплением в спайковых нейронных сетях в системе управления роботом [Using reinforcement learning in spiking neural networks for a robotic control system] // Труды 7-ой научно-технической конференции «Мехатроника, Автоматизация, Управление» (МАУ-2010) [Proceedings of the 7-th conference «Mechatronics, Automation, Control»]. - Спб. 2010. -С. 361-364
111. **Sinyavskiy O.Y., Kobrin A.I**. Понижение неопределенности времен генерации спайков с помощью минимизации полной условной энтропии нейрона [Decreasing the undeterminancy of the spike generation process by minimizing conditional entropy of the neuron] // XII Всероссийская Научно-Техническая Конференция "Нейроинформатика-2010". Сборник Научных Трудов. [Proceedings of the Russian scientific conference "Neuroinformatics-2010"] - М., 2010. -С. 276-285.
112. **Sinyavskiy O.Y., Kobrin A.I.** Использование информационных характеристик потока импульсных сигналов для обучения спайковых нейронных сетей [Using information theoretic quantities of the impulse signal streams in spiking neural networks training]// Интегрированные модели и мягкие вычисления в искусственном интеллекте (2009 г.) Сборник научных трудов. [Proceedings of the conference "Integrated models and fuzzy computations in artificial intelligence"] – 2009. -Т.2. -С. 678-687.
113. **Sinyavskiy O.Y., Kobrin A.I.** Обучение спайковых нейронных сетей работе с нестационарными импульсными последовательностями [Training of spiking networks to process nonstationary impulse sequences]// XI Всероссийская Научно-Техническая Конференция "Нейроинформатика-2009". Сборник Научных Трудов. [Proceedings of the 11-th Russian scientific conference "Neuroinformatics-2009] МИФИ, М., 2009. -С. 139-149.
114. **Sinyavskiy O.Y., Kobrin A.I.** Обучение динамических нейронных сетей работе с нестационарными импульсными последовательностями [Training of dynamic neural networks to process nonstationary impulse sequences] // Российская ассоциация искусственного интеллекта КИИ-2008 Одиннадцатая национальная конференция по искусственному интеллекту с международным участием. Труды конференции. [Proceedings of the 11-th national artificial intelligence conference "KII-2008"] - М., "Ленанд", 2008. -Т. 1. -С. 251-259

115. **Sinyavskiy O.Y.** Принципы построения универсальной программы для работы с искусственными нейронными сетями [Methods of constructing the software to for artificial neural networks development] //Труды международной конференции "Современные проблемы математики, механики и информатики" [Proceedings of the conference "Modern problems in mathematics, mechanics and informatics"]. -ТулГУ, 2005.-С. 121-127